\documentclass[lettersize,journal]{IEEEtran}
\usepackage{amsmath,amsfonts}
\usepackage{algorithmic}
\usepackage{array}
\usepackage[caption=false,font=normalsize,labelfont=sf,textfont=sf]{subfig}
\usepackage{textcomp}
\usepackage{stfloats}
\usepackage{url}
\usepackage{verbatim}
\usepackage{graphicx}
\usepackage{algorithmic}                        
\usepackage{algorithm}                          
\usepackage{cite}                               
\usepackage{balance}
\usepackage{amssymb}                            
\usepackage{mathtools}                          
\usepackage{medmath}                            
\usepackage{bm}                                 
\usepackage{eucal}                              
\usepackage{float}                              
\usepackage{afterpage}                          
\usepackage{booktabs}                           
\usepackage{comment}                            
 \usepackage{multirow}                          
\usepackage{tikz}                               
\usetikzlibrary{arrows.meta, positioning}       
\usepackage{xcolor}                             
\usepackage[colorlinks=true, 
            citecolor=blue,     
            linkcolor=blue,     
            urlcolor=blue,      
            bookmarks=true, 
            bookmarksopen=true,
            pdftex]{hyperref}

\let\oldcite\cite               
\renewcommand{\cite}[1]{{\color{blue}\oldcite{#1}}}

\let\oldeqref\eqref             
\renewcommand{\eqref}[1]{{\color{blue}\oldeqref{#1}}}

\let\oldref\ref                 
\renewcommand{\ref}[1]{{\color{blue}\oldref{#1}}}

\makeatletter
\pretocmd{\@seccntformat}{\color{black}}{}{}
\makeatother

\def\BibTeX{{\rm B\kern-.05em{\sc i\kern-.025em b}\kern-.08em
    T\kern-.1667em\lower.7ex\hbox{E}\kern-.125emX}}
\begin{document}
\title{Dual-Level Models for Physics-Informed Multi-Step Time Series Forecasting}

\author{
  \IEEEauthorblockN{Mahdi Nasiri\textsuperscript{1}, Johanna Kortelainen\textsuperscript{2}, and Simo Särkkä\textsuperscript{1,3}}\\
  \IEEEauthorblockA{\textsuperscript{1}Department of Electrical Engineering and Automation, Aalto University, Espoo, Finland}\\
  \IEEEauthorblockA{\textsuperscript{2}Metso Oyj, Tampere, Finland}\\
  \IEEEauthorblockA{\textsuperscript{3}ELLIS Institute Finland, Aalto University, Espoo, Finland}\\
}

\maketitle

\begin{abstract}
This paper develops an approach for multi-step forecasting of dynamical systems by integrating probabilistic input forecasting with physics-informed output prediction. Accurate multi-step forecasting of time series systems is important for the automatic control and optimization of physical processes, enabling more precise decision-making. While mechanistic-based and data-driven machine learning (ML) approaches have been employed for time series forecasting, they face significant limitations. Incomplete knowledge of process mathematical models limits mechanistic-based direct employment, while purely data-driven ML models struggle with dynamic environments, leading to poor generalization. To address these limitations, this paper proposes a dual-level strategy for physics-informed forecasting of dynamical systems. On the first level, input variables are forecast using a hybrid method that integrates a long short-term memory (LSTM) network into probabilistic state transition models (STMs). On the second level, these stochastically predicted inputs are sequentially fed into a physics-informed neural network (PINN) to generate multi-step output predictions. The experimental results of the paper demonstrate that the hybrid input forecasting models achieve a higher log-likelihood and lower mean squared errors (MSE) compared to conventional STMs. Furthermore, the PINNs driven by the input forecasting models outperform their purely data-driven counterparts in terms of MSE and log-likelihood, exhibiting stronger generalization and forecasting performance across multiple test cases. 
\end{abstract}

\begin{IEEEkeywords}
Machine learning, physics-informed neural networks, time series forecasting, Gaussian process, Kalman filter, predictive modeling.
\end{IEEEkeywords}

\section{Introduction}
\label{sec:intro}
\IEEEPARstart{T}{ime} series forecasting refers to the task of using historical data to predict the future behavior of a time series~\cite{Kong2025, Benidis2022, Hendikawati2020, Hoang2015, Antti2007}, and it has applications across various domains such as traffic flow prediction~\cite{li2018diffusion, Lv2015}, weather forecasting~\cite{Angryk2020}, and energy management~\cite{Zhang2025forecasting, HONG2016, SAXENA2019, SMYL2019}. It is a particularly important task in optimization and automated decision-making in industrial processes~\cite{Athiyarath2020, Kong2025, Faloutsos2019}. Forecasting can be performed one step ahead or multiple steps into the future. Multi-step forecasting, however, poses a greater challenge due to additional complications such as error accumulation, accuracy reduction, and the difficulty of uncertainty propagation over the prediction horizon~\cite{Hoang2015, Bontempi2013}.

Various approaches have been employed to address forecasting problems, including classical signal processing methods such as exponential smoothing, multiple linear regression, Kalman filtering (KF), and stochastic time series analysis~\cite{Kong2025, Sharma2020}; machine learning approaches such as deep neural networks~\cite{FISCHER2018, zhou2021informer}; and physics-based methods such as deterministic physical models extended with statistical approaches~\cite{Zhangphysicsbased2018, Langenbruchphysicsbased2018}. While classical signal processing methods provide interpretability and uncertainty quantification, they lack the flexibility to model nonlinearity and long-term temporal dependencies. Furthermore, methods like the Kalman filter~\cite{kalman1960} require the state and observation functions to be known a priori. Deep learning approaches~\cite{Goodfellow-et-al-2016}, particularly recurrent architectures like long short-term memory (LSTM) networks, are effective at discovering complex temporal structures and modeling nonlinear relationships. However, they face significant limitations, including a strong dependence on large volumes of training data, challenges in handling dynamic and noisy real-world processes, and a lack of physical interpretability, which limit their ability to extrapolate and generalize beyond the training data~\cite{karniadakis2021physics, willard2022integrating, rajulapati2022integration}. Physics-based models, by contrast, are generally robust and well-established; however, they often rely on simplifying assumptions due to system complexity and incomplete knowledge of the underlying processes, which can lead to biased predictions~\cite{Jovanovic2015, willard2022integrating}. 

To address the aforementioned challenges, methodologies that combine data-driven learning with established theoretical models have emerged. One such approach integrates neural networks with the Kalman filter, leveraging data-driven pattern recognition while preserving structural interpretability and allowing uncertainty quantification~\cite{Sharma2020, Zhou2024}. Another paradigm is physics-informed neural networks (PINNs), which embed physical laws directly into the learning process to produce solutions that are physically interpretable and often lead to improved generalization, even with sparse or noisy data~\cite{raissi2019physics, willard2022integrating, Kapoor2024}. In this paper, the aim is to combine Kalman filtering and PINNs into a dual-level methodology that leverages the strengths of both approaches.

The contribution of this paper is to develop a novel physics-informed approach for multi-step forecasting of dynamical systems with built-in uncertainty quantification. The approach consists of two levels. The first level uses a stochastic model for forecasting the system inputs. This first-level model can be either a pure stochastic differential equation or a hybrid model that integrates a long short-term memory (LSTM) network into the state-transition model. At the second level, the forecast inputs are propagated through a physics-informed neural network, constrained by the governing equations of the system, to predict the system's future outputs. The experimental results demonstrate that the LSTM-integrated state-space models (SSMs) achieve superior input-forecasting accuracy compared to classical state-space models. Furthermore, the results show that PINNs coupled with forecast inputs exhibit better long-horizon stability and generalization than purely data-driven models, and provide more precise uncertainty quantification.

\section{Preliminaries}
\label{sec:preliminary}
This section discusses the fundamental concepts and methodologies that form the basis of the proposed approach. It begins with a review of temporal Gaussian process state-space models, followed by parameter estimation techniques for linear state-space models. The section then presents the discrete-time physics-informed neural network and concludes with an overview of multi-step time series forecasting strategies, including recursive forecasting.

\subsection{Temporal Gaussian Process State-Space Models}
\label{subsec:gp_ssm} 
Gaussian process regression is a statistical machine learning approach that employs Gaussian processes (GPs) as prior distributions over regression functions, which are then conditioned on observed data~\cite{williams2006gaussian}. In system identification, GPs can model input-output mappings from observations; when explicit inputs are absent, they can be used to construct models directly for time series outputs~\cite{Särkkä2019}. Temporal GPs enable direct time-domain regression by treating time series as functions of time, conditioned on discrete observations~\cite{hartikainen2010kalman}. 

A significant limitation of Gaussian process regression is its computational complexity, which becomes infeasible for long time series due to cubic scaling with the number of observations~\cite{williams2006gaussian, Särkkä2019book}. However, for stationary processes, a temporal GP regression problem corresponds to state estimation in a partially observed stochastic differential equation (SDE) model that can be represented as a state-space model of the form~\cite{hartikainen2010kalman, Särkkä2019book}
\begin{equation}
\begin{aligned}
\frac{d\mathbf{x}(t)}{dt} &= \mathbf{F} \mathbf{x}(t) + \mathbf{L} \mathbf{w}(t), \\
y_k &= \mathbf{H} \mathbf{x}(t_k) + r_k,
\end{aligned}
\label{eq:sde_continuous}
\end{equation}
where $\mathbf{x}(t) = (x_1(t), x_2(t), \dots, x_m(t))$ represents a vector of $m$ stochastic processes, \(\mathbf{w}(t) \in \mathbb{R}^s\) denotes a white noise process with spectral density matrix \(\mathbf{\Sigma}_w \in \mathbb{R}^{s \times s}\), and the measurement noise is $r_k \sim \CMcal{N}(0, R)$. The dynamics are governed by the transition matrix \(\mathbf{F} \in \mathbb{R}^{m \times m}\), noise influence matrix \(\mathbf{L} \in \mathbb{R}^{m \times s}\), and initial state covariance matrix \(\mathbf{P}_0\) of the initial state with $\mathbb{E}[\mathbf{x}(0)] = \mathbf{0}$.

While the continuous time SDE formulation in Equation~\eqref{eq:sde_continuous} is convenient for theoretical modeling, numerical implementation requires a discrete time version. To employ standard filtering and smoothing methods to solve the GP regression problem using methods from signal processing~\cite{hartikainen2010kalman, sarkka2023bayesian}, the SDE must be discretized at time points corresponding to observation and prediction instants. For discrete time states \(\mathbf{x}_k = \mathbf{x}(t_k)\), the state-space model becomes
\begin{equation}
\begin{aligned}
\mathbf{x}_k &= \mathbf{A}_{k-1} \mathbf{x}_{k-1} + \mathbf{q}_{k-1}, 
&\quad \text{with} &\quad \mathbf{q}_{k-1} \sim \CMcal{N}(\mathbf{0}, \mathbf{Q}_{k-1}), \\
y_k &= \mathbf{H} \mathbf{x}_k + r_k, 
&\quad \text{with} &\quad r_k \sim \CMcal{N}(0, R_k).
\end{aligned}
\label{eq:state_space_model}
\end{equation}
Here, \(\mathbf{A}_k\) represents the discrete time transition matrix between time steps \(t_k\) and \(t_{k+1}\), and \(\mathbf{Q}_k\) denotes the corresponding process noise covariance. The recursion is initialized with the prior \(\mathbf{x}_0 \sim \CMcal{N}(\mathbf{0}, \mathbf{P}_0)\), and the discrete time matrices are computed as
%
\begin{align}
\mathbf{A}_k &= \Phi(\Delta t_k), \label{eq:A_k}\\
\mathbf{Q}_k &= \int_0^{\Delta t_k} \Phi(\Delta t_k - \tau) \mathbf{L} \mathbf{\Sigma}_w \mathbf{L}^{\top} \Phi(\Delta t_k - \tau)^{\top}  d\tau,
\label{eq:Q_k}
\end{align}
where \(\Phi(\tau) = \exp(\mathbf{F} \tau)\) and \(\Delta t_k = t_{k+1} - t_k\). The integral in Equation~\eqref{eq:Q_k} can be computed using matrix fraction decomposition (see, \cite{Särkkä2019book} for details).

When the model and its associated covariance function are stationary, the stochastic differential equation has a steady-state \(\mathbf{x}_{\infty} \sim \CMcal{N}(\mathbf{0}, \mathbf{P}_{\infty})\). This steady-state reflects the behavior the system converges to as \(t \to \infty\), characterized by the stationary covariance matrix \(\mathbf{P}_{\infty}\) satisfying
\begin{equation}
    \frac{d\mathbf{P}_{\infty}}{dt} = \mathbf{F} \mathbf{P}_{\infty} + \mathbf{P}_{\infty} \mathbf{F}^{\top} + \mathbf{L} \mathbf{\Sigma}_w  \mathbf{L}^{\top} = \mathbf{0},
\label{eq:P_inf}
\end{equation}
which is a specific case of the continuous time Riccati equation, known as the Lyapunov equation. The discrete time equivalent of this equation is given by
\begin{equation}
    \mathbf{P}_{\infty} = \mathbf{A}_k \mathbf{P}_{\infty} \mathbf{A}_k^\top + \mathbf{Q}_k.
\label{eq:P_inf_discrete}
\end{equation}

For stationary models, the steady-state distribution is the stochastic state where the process stays once it has reached it. Consequently, if we select the initial state covariance to be the stationary covariance matrix, \(\mathbf{P}_0 = \mathbf{P}_{\infty}\), the process will be at the steady state for all $t \geq 0$. When \(\mathbf{P}_{\infty}\) exists and is known, the process noise covariance \(\mathbf{Q}_k\) can be efficiently computed by rearranging the discrete Lyapunov equation as
\begin{equation}
\mathbf{Q}_k = \mathbf{P}_{\infty} - \mathbf{A}_k \mathbf{P}_{\infty} \mathbf{A}_k^\top,
\label{eq:Q_k_simple}
\end{equation}
which provides a computationally lightweight alternative to solving the integral in Equation~\eqref{eq:Q_k}. 

Given the linear discrete time state-space model in Equation~\eqref{eq:state_space_model}, the GP regression problem can be solved using the Kalman filter and Rauch--Tung--Striebel (RTS) smoother~\cite{hartikainen2010kalman}.

\subsection{Parameter Estimation in Linear State Space Models}
\label{subsec:param_est}
The previous section considered the estimation of the latent state, where the state-space model parameters are assumed to be known. However, in real-world scenarios, there are model parameters, \(\mathbf{\theta}\), that typically are unknown and must be estimated. This section discusses maximum likelihood parameter estimation~\cite{sarkka2023bayesian} via direct optimization for linear Gaussian state-space models.

Consider a discrete linear Gaussian state-space model where the unknown parameters are denoted by \(\mathbf{\theta}\):
\begin{equation}
\begin{aligned}
\mathbf{x}_{k} &= \mathbf{A}(\mathbf{\theta}) \,\mathbf{x}_{k-1} + \mathbf{q}_{k-1}, \\
\mathbf{y}_{k} &= \mathbf{H}(\mathbf{\theta}) \,\mathbf{x}_{k} + r_{k},
\end{aligned}
\label{eq:parametric_state_space}
\end{equation}
with process noise \(\mathbf{q}_{k-1} \sim \CMcal{N}(\mathbf{0}, \mathbf{Q}(\mathbf{\theta}))\), measurement noise \(r_{k} \sim \CMcal{N}(0, R(\mathbf{\theta}))\), and initial state \(\mathbf{x}_{0} \sim \CMcal{N}(\mathbf{m}_0(\mathbf{\theta}), \mathbf{P}_0(\mathbf{\theta}))\). For notational simplicity, the model matrices in Equation~\eqref{eq:parametric_state_space} are assumed to be time-invariant.

The estimation of the parameters \(\mathbf{\theta}\) can be performed by minimizing the unnormalized negative log-posterior or energy function
\begin{equation}
    \varphi_{k}(\mathbf{\theta}) = \varphi_{k-1}(\mathbf{\theta}) + \frac{1}{2} \log |2\pi\,\mathbf{S}_{k}(\mathbf{\theta})| + \frac{1}{2} \mathbf{v}_{k}^{\mathsf{T}}(\mathbf{\theta})\, \mathbf{S}_{k}^{-1}(\mathbf{\theta})\, \mathbf{v}_{k}(\mathbf{\theta}),
\label{eq:energy_func}
\end{equation}
where \(\mathbf{S}_{k}(\mathbf{\theta})\) and \(\mathbf{v}_{k}(\mathbf{\theta})\) are computed using the Kalman filter with parameters fixed to \(\mathbf{\theta}\). With a non-informative prior, corresponding to $\varphi_{0}(\mathbf{\theta}) = 0$, this is equivalent to maximum likelihood estimation of the parameters~\cite{sarkka2023bayesian}.

The Kalman filter recursions consist of the following prediction and update steps:
\begin{itemize}
    \item \textbf{Prediction:}
    \begin{align}
        \mathbf{m}_{k}^{-}(\mathbf{\theta}) &= \mathbf{A}(\mathbf{\theta})\,\mathbf{m}_{k-1}(\mathbf{\theta}), \nonumber \\
        \mathbf{P}_{k}^{-}(\mathbf{\theta}) &= \mathbf{A}(\mathbf{\theta})\,\mathbf{P}_{k-1}(\mathbf{\theta})\,\mathbf{A}^{\mathsf{T}}(\mathbf{\theta}) + \mathbf{Q}(\mathbf{\theta}).
        \label{eq:kalman_pred}
    \end{align}
    
    \item \textbf{Update:}
    \begin{align}
        \mathbf{v}_{k}(\mathbf{\theta}) &= \mathbf{y}_{k} - \mathbf{H}(\mathbf{\theta})\,\mathbf{m}_{k}^{-}(\mathbf{\theta}), \nonumber \\
        \mathbf{S}_{k}(\mathbf{\theta}) &= \mathbf{H}(\mathbf{\theta})\,\mathbf{P}_{k}^{-}(\mathbf{\theta})\,\mathbf{H}^{\mathsf{T}}(\mathbf{\theta}) + R(\mathbf{\theta}), \nonumber \\
        \mathbf{K}_{k}(\mathbf{\theta}) &= \mathbf{P}_{k}^{-}(\mathbf{\theta})\,\mathbf{H}^{\mathsf{T}}(\mathbf{\theta})\,\mathbf{S}_{k}^{-1}(\mathbf{\theta}), \nonumber \\
        \mathbf{m}_{k}(\mathbf{\theta}) &= \mathbf{m}_{k}^{-}(\mathbf{\theta}) + \mathbf{K}_{k}(\mathbf{\theta})\,\mathbf{v}_{k}(\mathbf{\theta}), \nonumber \\
        \mathbf{P}_{k}(\mathbf{\theta}) &= \mathbf{P}_{k}^{-}(\mathbf{\theta}) - \mathbf{K}_{k}(\mathbf{\theta})\,\mathbf{S}_{k}(\mathbf{\theta})\,\mathbf{K}_{k}^{\mathsf{T}}(\mathbf{\theta}).
        \label{eq:kalman_update}
    \end{align}
\end{itemize}

Thus, by running the Kalman filter with parameters fixed to \(\mathbf{\theta}\), the energy function, which is the unnormalized negative log-posterior of the parameters given the data up to time $k$, can be computed recursively. The minimization of \(\varphi_k(\mathbf{\theta})\) can be performed using standard optimization techniques, including both gradient-based and gradient-free methods~\cite{sarkka2023bayesian}.

\subsection{Physics-Informed Neural Networks}
\label{subsec:discrete_pinn}
Physics-informed neural networks (PINNs)~\cite{raissi2017physics} integrate physical laws with observed data to model physical systems. These systems are typically described using differential equations, including linear or nonlinear partial differential equations (PDEs) and ordinary differential equations (ODEs), which define constraints that solutions must satisfy over a specified domain. PINNs employ neural networks to approximate the solutions of these differential equations, effectively reformulating the solving of differential equations as an optimization task~\cite{raissi2019physics, karniadakis2021physics, Cuomo2022}. 

Consider a physical system governed by a parameterized nonlinear partial differential equation,  
\begin{equation}  
u_t + \CMcal{N}[u; \mathbf{\lambda}] = 0, \quad x \in \Omega, \quad t \in [0, T],  
\label{eq:nonlinear_operator_param}  
\end{equation}  
where the latent state is denoted by \( u(t, x) \), \( u_t \) represents its partial derivative with respect to time, and \( \CMcal{N}[\cdot; \mathbf{\lambda}] \) is a nonlinear differential operator parameterized by \( \mathbf{\lambda} \). The spatial domain \( \Omega \subseteq \mathbb{R}^D \) has dimensionality \( D \). The differential equation defined in Equation~\eqref{eq:nonlinear_operator_param} encompasses a wide range of physical systems, including those described by conservation laws and kinetic equations.

Given sparse and noisy measurement data of the physical system, PINNs are used to address two main tasks~\cite{raissi2017physics, raissi2019physics}. The first, referred to as the \emph{forward problem}, involves solving the differential equation to determine the latent state \( u(t, x) \) when the parameters \( \mathbf{\lambda} \) are known. The second task, known as the \emph{inverse problem}, focuses on identifying the unknown parameters of the equations that best describe the system from observations of the state, thereby discovering the underlying differential equations~\cite{Rudy2016DatadrivenDO, RaissiHidden, raissi2017physicspart2, raissi2019physics}. 

In the PINNs approach, the differential equation can be employed in either continuous or discrete form~\cite{raissi2019physics}. Following a similar approach, this paper focuses on solving the forward problem using the discrete form of the governing equations of physical systems. For that purpose, let us consider the differential equation, where we have dropped the dependence on the parameters for notational simplicity:
\begin{equation}  
u_t + \CMcal{N}[u] = 0, \quad x \in \Omega, \quad t \in [0, T].  
\label{eq:nonlinear_operator}  
\end{equation}  

Let us discretize Equation~\eqref{eq:nonlinear_operator} in time using a general $q$-stage Runge-Kutta method~\cite{iserles2009}, which yields
\begin{equation}
\begin{aligned}
u_{n+c_i} &= u_n - \Delta t \sum_{j=1}^{q} a_{ij} \CMcal{N}[u_{n+c_j}], \quad i = 1, \ldots, q, \\
u_{n+1} &= u_n - \Delta t \sum_{j=1}^{q} b_j \CMcal{N}[u_{n+c_j}],
\end{aligned}
\label{eq:rk_discretization}
\end{equation}
where $u_{n+c_j}(x) = u(t_n + c_j \Delta t, x)$ denotes the state at the $j$-th stage, $\Delta t$ represents the step size, and the coefficients $\{a_{ij}, b_j, c_j\}$ determine whether the time-stepping scheme is explicit or implicit. 

Equations~\eqref{eq:rk_discretization} can also be written in an equivalent form as
\begin{equation}
\begin{aligned}
u_n &= u_n^i, \quad i = 1, \ldots, q, \\
u_n &= u_n^{q+1},
\end{aligned}
\label{eq:rk_residual_form}
\end{equation}
with 
\begin{equation}
\begin{aligned}
u_n^i &:= u_{n+c_i} + \Delta t \sum_{j=1}^{q} a_{ij} \CMcal{N}[u_{n+c_j}], \quad i = 1, \ldots, q, \\
u_n^{q+1} &:= u_{n+1} + \Delta t \sum_{j=1}^{q} b_j \CMcal{N}[u_{n+c_j}].
\end{aligned}
\label{eq:residual_definitions}
\end{equation}
A multi-output neural network can now be employed to approximate the solution of all Runge-Kutta stages and the next time step
\begin{equation}
\left[u_{n+c_1}(x), \ldots, u_{n+c_q}(x), u_{n+1}(x)\right].
\label{eq:network_outputs}
\end{equation}
This neural network approximation, along with the constraints in Equations~\eqref{eq:residual_definitions}, forms a physics-informed neural network that takes spatial coordinates $x$ as input and produces
\begin{equation}
\left[u^1_n(x), \ldots, u^q_n(x), u^{q+1}_n(x)\right].
\label{eq:residual_outputs}
\end{equation}

The parameters of the physics-informed neural network are optimized by minimizing a mean squared error (MSE) loss function~\cite{raissi2017physics}, defined as
\begin{equation} 
    \mathrm{MSE} = \frac{1}{N_n} \sum_{j=1}^{q+1}\sum_{i=1}^{N_{n}}|u^{j}_{n}(x_{n,i})-u_{n,i}|^{2},
    \label{eq:PINN_MSE}
\end{equation}
where the set \( \{x_{n,i}, u_{n,i}\}_{i=1}^{N_{n}} \) represents the training data at time $t_n$. The loss function can be extended to include additional terms, as discussed in the context of the specific PINN models considered in the following sections.   

The performance of the discrete-time PINN approach is controlled by the step size $\Delta t$ and the number of Runge-Kutta stages $q$. A primary advantage of the discrete-time PINN approach is that it allows the use of a high number of Runge-Kutta stages, which enables taking large time steps without sacrificing predictive accuracy~\cite{raissi2019physics}. This capability is particularly valuable for efficiently resolving solutions over long time horizons. Furthermore, by employing A-stable implicit Runge-Kutta schemes, the method maintains numerical stability even for stiff problems, regardless of the chosen order $q$. The theoretical temporal error for such schemes scales as $\mathcal{O}(\Delta t^{2q})$~\cite{iserles2009}, which can drive the discretization error far below machine precision for sufficiently high $q$.

\subsection{Multi-step Time Series Forecasting}
\label{subsec:forecast}
Time series forecasting~\cite{Kong2025, Benidis2022, Hendikawati2020, Hoang2015, Antti2007} involves predicting future values from historical observations of sequentially ordered data. In such problems, past and present measurements are used as inputs to a predictive model, while the desired output is the estimated future value. Forecasting can be employed for short-term or long-term horizons. Short-term prediction refers to single-step-ahead forecasting, whereas long-term prediction involves forecasting several steps ahead and is generally more challenging due to error accumulation and increasing uncertainty over time.

Five established strategies address multi-step forecasting: recursive or iterative, direct, direct-recursive (DirREC), multiple-input multiple-output (MIMO), and direct multiple-output (DirMO) (see,~\cite{Hoang2015} for details). Forecasting models may be univariate or multivariate. Univariate forecasting focuses on predicting the future of a single time-dependent variable, whereas multivariate forecasting simultaneously models multiple interdependent variables, accounting for their correlations during prediction~\cite{Torres2021, Kong2025}. This paper focuses on the recursive forecasting approach and tackles both univariate and multivariate problems.

\subsubsection{Recursive Forecasting}
\label{subsubsec:recursive_forecast}
The recursive or iterative strategy represents one of the earliest and most intuitive approaches to multistep forecasting~\cite{Kline2004, Antti2007, HAMZACEBI20093839}. In this approach, a single model, denoted as $f$, is optimized to predict the target one step into the future. For predictions extending $H$ steps, with $H>1$ indicating the \textit{forecast horizon}, the procedure begins with the model making a prediction for the initial step. This predicted value is then integrated as an input variable for the next step prediction, employing the same model designed for one-step predictions. This process repeats sequentially through each step until predictions cover the entire forecast horizon.

Consider a forecasting model $f$ employed to predict the next time step using a history of $L$ lags
\begin{equation}
    \mathbf{y}_{t+1} = f\left( \mathbf{y}_t, \dots, \mathbf{y}_{t-L+1}, \mathbf{x}_t, \dots, \mathbf{x}_{t-L+1} \right),
    \label{eq:forecast_model}
\end{equation}
with $\mathbf{y}_{t} \in \mathbb{R}^{d_y}$ representing target variables, $\mathbf{x}_{t} \in \mathbb{R}^{d_x}$ denoting corresponding exogenous input variables at time $t$, and $t \in \{L, \dots, N-1\}$. Let the trained model for one-step predictions be denoted as $\hat{f}$. The predictions across a forecast horizon of length $H$ can then be recursively expressed as
\begin{equation}
    \hat{\mathbf{y}}_{N+h} = 
    \begin{cases}
        \hat{f}(\mathbf{y}_N, \dots, \mathbf{y}_{N-L+1}, & \\ 
        \quad \mathbf{x}_N, \dots, \mathbf{x}_{N-L+1}) & \text{if } h = 1 \\
        \hat{f}(\hat{\mathbf{y}}_{N+h-1}, \dots, \hat{\mathbf{y}}_{N+1}, & \\
        \quad \mathbf{y}_N, \dots, \mathbf{y}_{N+h-L}, & \\
        \quad \mathbf{x}_{N+h-1}, \dots, \mathbf{x}_{N+h-L}) & \text{if } h \in \{2, \dots, L\} \\
        \hat{f}(\hat{\mathbf{y}}_{N+h-1}, \dots, \hat{\mathbf{y}}_{N+h-L}, & \\
        \quad \mathbf{x}_{N+h-1}, \dots, \mathbf{x}_{N+h-L}) & \text{if } h \in \{L+1, \dots, H\}
    \end{cases}
    \label{eq:recursive_forecast}
\end{equation}
where the exogenous variables $\mathbf{x}_{N+h}$ must be available at prediction time, either through known future values or independent forecasts.

\begin{figure}[h]
  \centering
  \includegraphics[width=\linewidth]{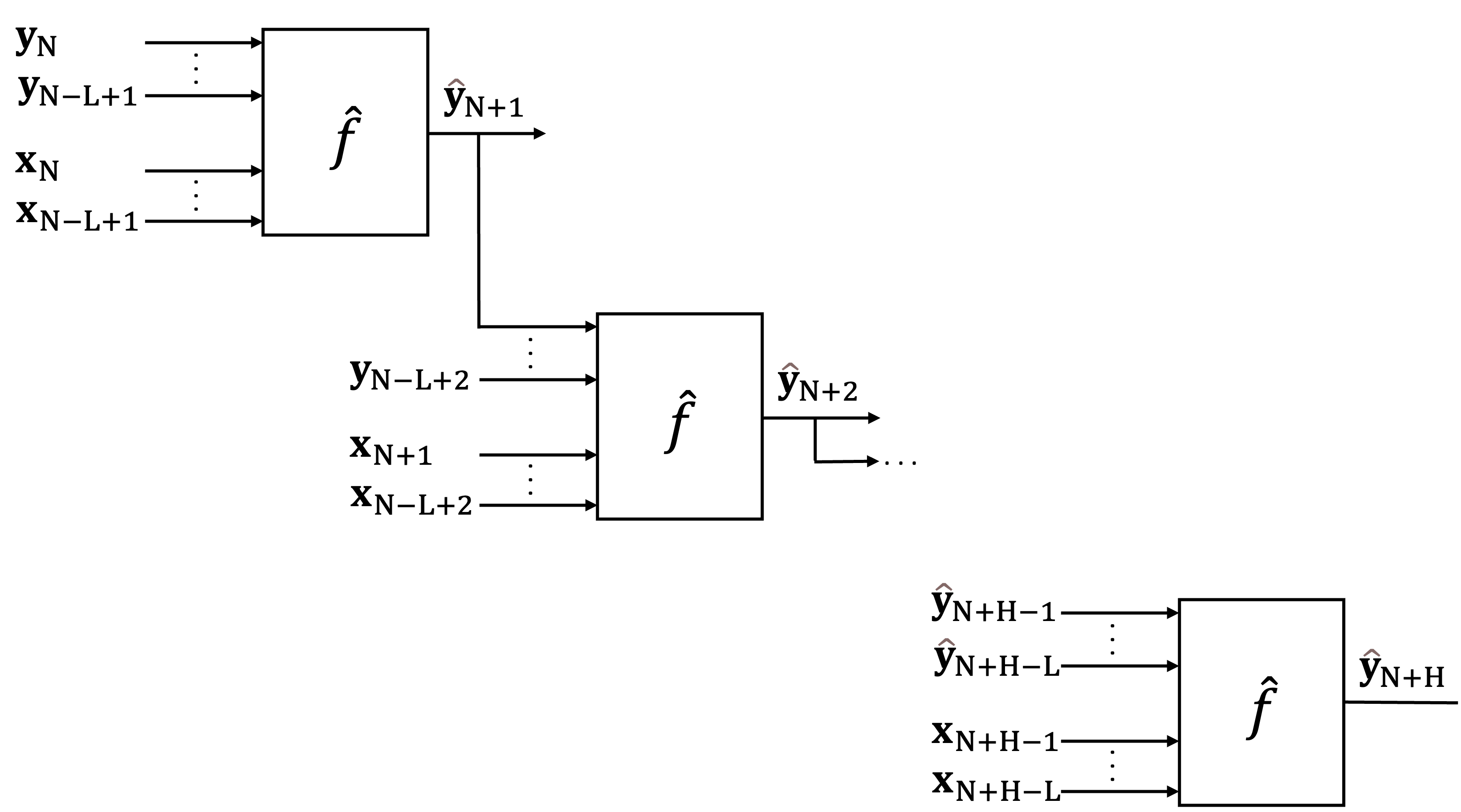}
  \caption{Schematic of recursive multistep forecasting. The model $\hat{f}$ is applied iteratively, with predictions fed back as inputs for subsequent steps. The composition of the feature window evolves as the forecast horizon increases, transitioning from windows containing only observed target values (top) to windows containing predicted target values (bottom). Exogenous variables $\mathbf{x}_t$ appear at each time step, representing either known inputs or independent forecasts.}
  \label{fig:multi_step_forecast}
\end{figure}

Figure~\ref{fig:multi_step_forecast} illustrates the recursive approach employed for multistep forecasting. Initially, the trained model receives a window of historical values of the target variable $\mathbf{y}$ and exogenous variable $\mathbf{x}$ to make a prediction one step ahead. Subsequently, the forecasted value $\hat{\mathbf{y}}$ is used as part of the input window for the following step's prediction, using the same model. This procedure is repeated for each step until the predictions reach the end of the forecast horizon.

\section{Materials and Methods}
\label{sec:methods}
This section presents the methodological approach that we developed for multi-step forecasting of dynamical systems. The proposed dual-level strategy integrates probabilistic state-space modeling for input forecasting with physics-informed neural networks for output prediction. First, two conventional Gaussian process state-space models are discussed, along with their hybrid counterparts that integrate long short-term memory networks into the state-transition function, which are used to forecast the inputs of the dynamical system. Then, three different dynamical systems used to construct discrete-time PINNs for output prediction are presented. Finally, the multi-step forecasting strategy for the complete system is detailed, describing how the forecast inputs are sequentially propagated through the physics-informed models.

\subsection{Augmented State-Space Models}
\label{subsec:augmented_models}
This work develops augmented state-space models based on the state-space representations of two widely used kernels, namely the Matérn and exponential covariance functions. The standard forms of these models describe zero-mean processes. To enhance their flexibility for applications where the underlying process exhibits a non-zero mean or a trend, the models are augmented by introducing additional state variables. The following sections describe these augmented models in detail.

\subsubsection{Mat\'ern State-Space Model}
\label{subsubsec:matern32_model}
Building upon the state-space formulation of Gaussian processes, the first augmented model is constructed using the Matérn covariance function with smoothness parameter $\nu = 3 / 2$. This kernel models processes that are once differentiable, making it suitable for capturing smooth temporal dynamics~\cite{Särkkä2019}. The state-space representation for this kernel is given by
\begin{equation}
\frac{d}{dt} \begin{bmatrix} \tilde{x}_1(t) \\ \tilde{x}_2(t) \end{bmatrix}
=
\begin{bmatrix} 0 & 1 \\ -\lambda^2 & -2\lambda \end{bmatrix}
\begin{bmatrix} \tilde{x}_1(t) \\ \tilde{x}_2(t) \end{bmatrix}
+
\begin{bmatrix} 0 \\ 1 \end{bmatrix} w_1(t),
\label{eq:matern_base}
\end{equation}
where $w_1(t)$ is a white noise process with spectral density $q_{w1}$, and the state components $\tilde{x}_1(t)$ and $\tilde{x}_2(t)$ represent the process value and its derivative, respectively.

The Matérn model in Equation~\eqref{eq:matern_base} describes a zero-mean process, fluctuating around zero. To generalize this model for applications where the underlying process may have a non-zero mean, the model is augmented. This is achieved by incorporating a third state variable, $\tilde{x}_3(t)$, which is modeled as a Wiener process
\begin{equation}
\frac{d\tilde{x}_3(t)}{dt} = w_2(t),
\end{equation}
where $w_2(t)$ is an independent white noise process with spectral density $q_{w2}$. A new state vector $\mathbf{x}(t) = [x_1(t), x_2(t), x_3(t)]^\top$ is then defined as
\begin{align*}
x_1(t) &= \tilde{x}_1(t) + \tilde{x}_3(t), \\
x_2(t) &= \tilde{x}_2(t), \\
x_3(t) &= \tilde{x}_3(t).
\end{align*}
Finally, the augmented system can be expressed in canonical linear state-space form as
\begin{equation}
\begin{aligned}
\frac{d\mathbf{x}(t)}{dt} &=
\begin{bmatrix} 0 & 1 & 0 \\ -\lambda^2 & -2\lambda & \lambda^2 \\ 0 & 0 & 0 \end{bmatrix}
\mathbf{x}(t) +
\begin{bmatrix} 0 & 1 \\ 1 & 0 \\ 0 & 1 \end{bmatrix}
\begin{bmatrix} w_1(t) \\ w_2(t) \end{bmatrix}, \\
y_k &= \begin{bmatrix} 1 & 0 & 0 \end{bmatrix}\, \mathbf{x}(t_k) + r_k, \quad r_k \sim \CMcal{N}(0, R).
\end{aligned}
\label{eq:augmented_cont}
\end{equation}

For practical implementation, this continuous time model is discretized, following the procedure in Section~\ref{subsec:gp_ssm}. The discrete time transition matrix $\mathbf{A}_k$ and process noise covariance matrix $\mathbf{Q}_k$ are computed using Equations~\eqref{eq:A_k} and~\eqref{eq:Q_k}, yielding
\begin{equation}
\mathbf{A}_k = \medmath{\begin{bmatrix}
(1 + \lambda \Delta t)e^{-\lambda \Delta t} & \Delta t e^{-\lambda \Delta t} & 1 - (1 + \lambda \Delta t)e^{-\lambda \Delta t} \\
-\lambda^2 \Delta t e^{-\lambda \Delta t} & (1 - \lambda \Delta t)e^{-\lambda \Delta t} & \lambda^2 \Delta t e^{-\lambda \Delta t} \\
0 & 0 & 1
\end{bmatrix}},
\label{eq:Discrete_matern_A}
\end{equation}
\begin{equation}
\mathbf{Q}_k = \begin{bmatrix}
Q_{11} & Q_{12} & Q_{13} \\
Q_{21} & Q_{22} & 0 \\
Q_{31} & 0 & Q_{33}
\end{bmatrix},
\label{eq:Discrete_matern_Q}
\end{equation}
where
\begin{align*}
Q_{11} &= \frac{-q_{w1} e^{-2\lambda \Delta t} - q_{w1} - 4\Delta t \lambda^3 q_{w2}}{4\lambda^3} \\
       &\quad + \frac{2\Delta t \lambda q_{w1} e^{-2\lambda \Delta t} + 2\Delta t^2 \lambda^2 q_{w1} e^{-2\lambda \Delta t}}{4\lambda^3}, \\
Q_{12} &= Q_{21} = \frac{\Delta t^2 q_{w1} e^{-2\lambda \Delta t}}{2}, \\
Q_{13} &= Q_{31} = \Delta t q_{w2}, \\
Q_{22} &= \frac{q_{w1} e^{-2\lambda \Delta t} \left(e^{2\lambda \Delta t} + 2\Delta t \lambda - 2\Delta t^2 \lambda^2 - 1\right)}{4\lambda}, \\
Q_{33} &= \Delta t q_{w2}.
\end{align*}

The unknown parameters of this model, $\mathbf{\theta} = \{\lambda, q_{w1}, q_{w2}, R\}$, are estimated from observed data by minimizing the energy function, as described in Section~\ref{subsec:param_est}. In this work, the optimization is performed using the Adam optimizer~\cite{Kingma2014AdamAM}, a gradient descent-based algorithm.

\subsubsection{Exponential State-Space Model}
\label{subsubsec:ou_model}
The second model developed in this work is derived from the Matérn covariance family with smoothness parameter $\nu 1/2$, corresponding to the exponential kernel, and represents the Ornstein-Uhlenbeck process. This process is continuous but non-differentiable, making it appropriate for modeling systems with rough temporal characteristics. The state-space representation of this process consists of a first-order stochastic differential equation defined by
%
\begin{equation}
\frac{d\tilde{x}_1(t)}{dt} = -\lambda \tilde{x}_1(t) + w_1(t),
\label{eq:ou_base}
\end{equation}
where $w_1(t)$ is a white noise process with spectral density $q_{w1}$.

Following a similar augmentation approach to Section~\ref{subsubsec:matern32_model}, this zero-mean process is extended to capture non-zero mean behavior by introducing a second state variable $\tilde{x}_2(t)$ governed by a Wiener process
\begin{equation}
\frac{d\tilde{x}_2(t)}{dt} = w_2(t),
\end{equation}
where $w_2(t)$ denotes an independent white noise process with spectral density $q_{w2}$. A transformed state vector $\mathbf{x}(t) = [x_1(t), x_2(t)]^\top$ is then defined as
\begin{align*}
x_1(t) &= \tilde{x}_1(t) + \tilde{x}_2(t), \\
x_2(t) &= \tilde{x}_2(t).
\end{align*}
The augmented continuous-time state-space model is therefore given by
\begin{equation}
\begin{aligned}
\frac{d\mathbf{x}(t)}{dt} &=
\begin{bmatrix} -\lambda & \lambda \\ 0 & 0 \end{bmatrix}
\mathbf{x}(t)
+
\begin{bmatrix} 1 & 1 \\ 0 & 1 \end{bmatrix}
\begin{bmatrix} w_1(t) \\ w_2(t) \end{bmatrix}, \\
y_k &= \begin{bmatrix} 1 & 0 \end{bmatrix}\, \mathbf{x}(t_k) + r_k, \quad r_k \sim \CMcal{N}(0, R).
\end{aligned}
\label{eq:ou_augmented}
\end{equation}

The discrete-time transition matrix and process noise covariance matrix are given by
\begin{equation}
\mathbf{A}_k = \begin{bmatrix} e^{-\lambda \Delta t} & 1 - e^{-\lambda \Delta t} \\ 0 & 1 \end{bmatrix},
\label{eq:Discrete_OU_A}
\end{equation}

\begin{equation}
\mathbf{Q}_k = \begin{bmatrix} 
\dfrac{q_{w1}(1 - e^{-2 \lambda \Delta t}) + 2 \lambda \Delta t q_{w2}}{2 \lambda} & \Delta t q_{w2} \\ 
\Delta t q_{w2} & \Delta t q_{w2} 
\end{bmatrix},
\label{eq:Discrete_OU_Q}
\end{equation}
and the unknown parameters for this model are $\mathbf{\theta} = \{\lambda, q_{w1}, q_{w2}, R\}$.

\subsection{Hybrid State-Space Models}
\label{subsec:hybrid_ssm}
To enhance the modeling capacity of the augmented linear state-space models presented in Section~\ref{subsec:augmented_models} while preserving their probabilistic structure, we integrate a long short-term memory network into the state-transition models. In this hybrid approach, the LSTM leverages historical measurements to capture long-term temporal dependencies. Specifically, we employ the augmented state-space models derived from the Matérn and exponential covariance functions, replacing the linear prediction of the first state component with an LSTM-based prediction. The Kalman filter recursion is used for maximum likelihood parameter estimation during training and to obtain the initial state distribution for forecasting. The following subsections detail the hybrid state transition models and the corresponding modified filtering algorithm for both covariance functions.

\subsubsection{Hybrid Mat\'ern State-Space Model}
\label{subsubsec:hybrid_matern}
The first hybrid model builds upon the augmented state-space model obtained from the Mat\'ern covariance function. Recall the discrete-time state-space model
\begin{equation}
\begin{aligned}
\mathbf{x}_{k} &= \mathbf{A}(\mathbf{\theta}) \,\mathbf{x}_{k-1} + \mathbf{q}_{k-1}, &\quad &\mathbf{q}_{k-1} \sim \CMcal{N}(\mathbf{0}, \mathbf{Q}(\mathbf{\theta})),\\
\mathbf{y}_{k} &= \begin{bmatrix} 1 & 0 & 0 \end{bmatrix} \,\mathbf{x}_{k} + r_k, &\quad &r_k \sim \CMcal{N}(0, R(\mathbf{\theta})),
\end{aligned}
\label{eq:discrete_SS_matern}
\end{equation}
where $\mathbf{A}(\mathbf{\theta})$ and $\mathbf{Q}(\mathbf{\theta})$ are defined by Equations~\eqref{eq:Discrete_matern_A} and~\eqref{eq:Discrete_matern_Q}, respectively, and $\mathbf{\theta} = \{\lambda, q_{w1}, q_{w2}, R\}$.

In the hybrid formulation, the LSTM network is integrated into the state transition process to predict the first component of the state. Specifically, the LSTM is trained to predict the observable component of the state using a window of $n$ past measurements $\{\mathbf{y}_{k-1}, \mathbf{y}_{k-2}, \ldots, \mathbf{y}_{k-n}\}$, thereby learning temporal patterns that the linear state transition model cannot capture. The new state transition model is defined as
\begin{equation}
    \mathbf{x}_k = \begin{bmatrix} 
        \operatorname{LSTM}(\mathbf{y}_{k-1}, \mathbf{y}_{k-2}, \ldots, \mathbf{y}_{k-n}) \\
        \mathbf{a}_2^\top \mathbf{x}_{k-1} \\
        \mathbf{a}_3^\top \mathbf{x}_{k-1}
    \end{bmatrix} + \mathbf{q}_{k-1},
    \label{eq:hybrid_forecast_matern}
\end{equation}
where $\mathbf{a}_2$ and $\mathbf{a}_3$ denote the second and third rows of the state transition matrix $\mathbf{A}$, respectively, and $\mathbf{q}_{k-1} \sim \CMcal{N}(\mathbf{0}, \mathbf{Q})$. In the equation above, the dependency on the unknown parameters $\mathbf{\theta}$ is omitted for simplicity. 

The LSTM parameters are optimized jointly with the unknown parameters of the state-space model, $\mathbf{\theta}$, by minimizing the negative log-likelihood function defined in Equation~\eqref{eq:energy_func}. The required terms,  $\mathbf{S}_{k}$ and $\mathbf{v}_{k}$, in this function are computed using a modified Kalman filter algorithm, as follows:
\begin{itemize}
    \item \textbf{Prediction:}
\end{itemize}
\begin{align}
    \begin{split}
        \mathbf{m}_{k}^{-} &= \begin{bmatrix} 
            \operatorname{LSTM}(\mathbf{y}_{k-1}, \mathbf{y}_{k-2}, \ldots, \mathbf{y}_{k-n}) \\
            \mathbf{a}_2^\top \,\mathbf{m}_{k-1} \\
            \mathbf{a}_3^\top \,\mathbf{m}_{k-1}
        \end{bmatrix}, \\
        \mathbf{P}_{k}^{-} &= \mathbf{A} \,\mathbf{P}_{k-1} \,\mathbf{A}^{\mathsf{T}} + \mathbf{Q}.
    \end{split}
    \\
    \intertext{\begin{itemize}
        \item[\textbullet] \textbf{Update:}
    \end{itemize}}
    \begin{split}
        \mathbf{v}_{k} &= \mathbf{y}_{k} - \mathbf{H} \,\mathbf{m}_{k}^{-}, \\
        \mathbf{S}_{k} &= \mathbf{H} \,\mathbf{P}_{k}^{-} \,\mathbf{H}^{\mathsf{T}} + R, \\
        \mathbf{K}_{k} &= \mathbf{P}_{k}^{-} \,\mathbf{H}^{\mathsf{T}} \,\mathbf{S}_{k}^{-1}, \\
        \mathbf{m}_{k} &= \mathbf{m}_{k}^{-} + \mathbf{K}_{k} \,\mathbf{v}_{k}, \\
        \mathbf{P}_{k} &= \mathbf{P}_{k}^{-} - \mathbf{K}_{k} \,\mathbf{S}_{k} \,\mathbf{K}_{k}^{\mathsf{T}}.
    \end{split}
\end{align}
Note that $\mathbf{H}\mathbf{m}_{k}^{-}$ is equivalent to the LSTM output, that is $\mathbf{H}\mathbf{m}_{k}^{-} := \operatorname{LSTM}(\mathbf{y}_{k-1}, \mathbf{y}_{k-2}, \ldots, \mathbf{y}_{k-n})$. This approach leverages the interpretability of the Matérn state transition while modeling long-term dependencies in measurement dynamics via the LSTM.

%
%
%
%
%
\subsubsection{Hybrid Exponential State-Space Model}
\label{subsubsec:hybrid_ou}
The second hybrid model extends the approach to the augmented state-space model derived from the exponential covariance function in Section~\ref{subsubsec:ou_model}. Recall the discrete-time state-space representation
\begin{equation}
\begin{aligned}
\mathbf{x}_{k} &= \begin{bmatrix} e^{-\lambda \Delta t} & 1 - e^{-\lambda \Delta t} \\ 0 & 1 \end{bmatrix} \,\mathbf{x}_{k-1} + \mathbf{q}_{k-1},\\
\mathbf{y}_{k} &= \begin{bmatrix} 1 & 0 \end{bmatrix} \,\mathbf{x}_{k} + r_k,
\end{aligned}
\label{eq:discrete_SS_OU}
\end{equation}
where $\mathbf{q}_{k-1} \sim \CMcal{N}(\mathbf{0}, \mathbf{Q}(\mathbf{\theta}))$ and $r_k \sim \CMcal{N}(0, R(\mathbf{\theta}))$. Here, $\mathbf{Q}(\mathbf{\theta})$ is given by Equation~\eqref{eq:Discrete_OU_Q} and $\mathbf{\theta} = \{\lambda, q_{w1}, q_{w2}, R\}$.

Following the same methodology as Section~\ref{subsubsec:hybrid_matern}, an LSTM network is integrated into the state transition to predict the first state component. The resulting hybrid state transition model is defined as
\begin{equation}
    \mathbf{x}_k = \begin{bmatrix} 
        \operatorname{LSTM}(\mathbf{y}_{k-1}, \mathbf{y}_{k-2}, \ldots, \mathbf{y}_{k-n}) \\
        \mathbf{a}_2^\top \mathbf{x}_{k-1}
    \end{bmatrix} + \mathbf{q}_{k-1},
    \label{eq:hybrid_forecast_ou}
\end{equation}
where $\mathbf{q}_{k-1} \sim \CMcal{N}(\mathbf{0}, \mathbf{Q})$ and $\mathbf{a}_2$ denotes the second row of the state transition matrix $\mathbf{A}$. As before, the dependency on parameters $\mathbf{\theta}$ is omitted for simplicity.

The LSTM parameters are optimized jointly with $\mathbf{\theta}$ by minimizing the negative log-likelihood function, and within the corresponding modified Kalman filter recursion, the prediction of the state mean is given by
\begin{equation}
\mathbf{m}_{k}^{-} = \begin{bmatrix} 
    \operatorname{LSTM}(\mathbf{y}_{k-1}, \mathbf{y}_{k-2}, \ldots, \mathbf{y}_{k-n}) \\
    \mathbf{a}_2^\top \,\mathbf{m}_{k-1}
\end{bmatrix}.
\end{equation}
%

%
%
%
%
%
\subsection{Physics-Informed Neural Network Models}
\label{subsec:pinn}
This section discusses the three dynamical systems used to develop the discrete time physics-informed neural networks. For each system, the governing equations are first presented, followed by the formulation of the corresponding PINN model.

\subsubsection{Continuous Stirred Tank Reactor System}
\label{subsubsec:cstr_model}
A continuous stirred tank reactor (CSTR) is one of the most common systems in chemical process modeling, characterized by perfect mixing~\cite{SHARMA2017}. For this study, a CSTR in which a single component undergoes an irreversible second-order reaction, such as \(A + A \rightarrow B\), is considered. The reaction rate per unit volume is expressed as \(r = k C^2\), where \(k\) is the second-order reaction rate constant and \(C\) is the concentration of the reacting component within the reactor.

Assuming a constant fluid density and a constant reactor volume \(V\), the dynamic material balance for the reacting component is derived from~\cite{bequette1998process}
\begin{equation}
    \frac{dC}{dt} = \frac{F}{V} C_{in} - \frac{F}{V} C - k C^2,
    \label{eq:cstr_dynamic}
\end{equation}
where \(F\) is the constant volumetric feed flow rate, and \(C_{in}\) is the concentration of the component in the feed stream that is potentially time-varying. 

The discrete time nonlinear operator for the CSTR system is given by
\begin{equation}
    \CMcal{N}[C_{n+c_j}] = -\frac{F}{V} C_{in} + \frac{F}{V} C_{n+c_j} + k \left(C_{n+c_j}\right)^2,
    \label{eq:cstr_nonlinear}
\end{equation}
with \(C_{n+c_j}\) indicating the concentration evaluated at the intermediate Runge-Kutta time points \(t_n + c_j \Delta t\).

The physics-informed neural network parameters are optimized by minimizing the mean squared error loss
\begin{equation}
\mathrm{MSE} = \mathrm{MSE}_f + \mathrm{MSE}_u,
\label{eq:cstr_mse}
\end{equation}
where 
\begin{align}
\mathrm{MSE}_f = \frac{1}{N_n} \sum_{j=1}^{q+1} \sum_{i=1}^{N_n} \left|u_n^j(x_{n,i}) - u_{n,i}\right|^2,
\label{eq:cstr_mse_f}\\
\mathrm{MSE}_u = \frac{1}{N_n} \sum_{i=1}^{N_n} \left|u_{n+1}(x_{n,i}) - u_{n+1,i}\right|^2.
\label{eq:cstr_mse_u}
\end{align}
In the equations above, \( \{x_{n,i}, u_{n,i}, u_{n+1,i}\}_{i=1}^{N_{n}} \) denote the training data, $u = C$, and the input variable is $x = C_{in}$. The quantities \(u_n^j(x_{n,i})\) are computed using the constraints defined in Equations~\eqref{eq:residual_definitions} and \(u_{n+1}(x_{n,i})\) is the last output of the neural network. The loss component \(\mathrm{MSE}_f\) enforces the discrete physics constraints, whereas \(\mathrm{MSE}_u\) penalizes deviations between the predicted and observed concentrations at time $t_{n+1}$.

\subsubsection{Axial Dispersion Plug Flow Reactor System}
\label{subsubsec:PFR_model}
The plug flow reactor (PFR) represents a system in chemical reaction engineering where fluid molecules flow through a tubular configuration. The inclusion of axial dispersion is a recognized approach for representing real flow behavior in an ideal PFR, which assumes that fluid molecules move with a constant velocity at any point along the reactor axis~\cite{DIGIULIANO2019}. This study considers an isothermal PFR with axial dispersion, where a single component undergoes an irreversible second-order reaction.

Under the assumptions of isothermal operation, constant fluid properties, uniform cross-sectional area, constant axial dispersion coefficient \(D\), and negligible radial gradients in velocity and concentration, the dynamic behavior of the reactant concentration \(C\) along the reactor length is described by the microscopic material balance of the form~\cite{WilliamsModelling_2021}
\begin{equation}
    \frac{\partial C}{\partial t} = D \frac{\partial^2 C}{\partial z^2} - v \frac{\partial C}{\partial z} - k C^2,
    \label{eq:pfr_dynamic}
\end{equation}
with \(v\) representing the time-varying superficial velocity, \(k\) is the second-order reaction rate constant, \(z\) denotes the reactor axial coordinate, and \(t\) represents time.

The system satisfies the Danckwerts boundary conditions given by
\begin{align}
    \text{At } z = 0: &\quad v(C_{in} - C) = -D \frac{\partial C}{\partial z}, \label{eq:danckwerts_inlet} \\
    \text{At } z = L: &\quad \frac{\partial C}{\partial z} = 0, \label{eq:danckwerts_outlet}
\end{align}
where \(C_{in}\) denotes the time-varying inlet concentration of the component and \(L\) is the total reactor length. The initial condition is expressed by
\begin{equation}
    \text{At } t = 0: \quad C(z,0) = C_{0}(z),
    \label{eq:pfr_initial}
\end{equation}
with \(C_{0}(z)\) describing the initial concentration profile along the reactor. 

The nonlinear operator in Equations~\eqref{eq:residual_definitions} takes the form
\begin{equation}
    \CMcal{N}[C_{n+c_j}] = -D \frac{\partial^2 C_{n+c_j}}{\partial z^2} + v \frac{\partial C_{n+c_j}}{\partial z} + k \left(C_{n+c_j}\right)^2.
    \label{eq:pfr_nonlinear}
\end{equation}
The parameters optimization of the physics-informed neural network is obtained by minimizing the mean squared error loss
\begin{equation}
\mathrm{MSE} = \mathrm{MSE}_f + \mathrm{MSE}_u + \mathrm{MSE}_b,
    \label{eq:pfr_mse}
\end{equation}
where $\mathrm{MSE}_f$ and $\mathrm{MSE}_u$ are defined in Equations~\eqref{eq:cstr_mse_f} and~\eqref{eq:cstr_mse_u} respectively, and $\mathrm{MSE}_b$ enforces the boundary conditions, given by
\begin{equation}
\begin{aligned}
\mathrm{MSE}_b &= \frac{1}{N_n} \sum_{j=1}^{q} \sum_{i=1}^{N_n} \Big| v\big(C_{in} - u_{n+c_j}(z=0)\big) \\
&\hphantom{{}= \frac{1}{N_n} \sum_{j=1}^{q} \sum_{i=1}^{N_n} \Big|} + D \frac{\partial u_{n+c_j}}{\partial z}(z=0) \Big|^2 \\
&\quad + \frac{1}{N_n} \sum_{j=1}^{q} \sum_{i=1}^{N_n} \Big| \frac{\partial u_{n+c_j}}{\partial z}(z=L) \Big|^2.
\end{aligned}
\label{eq:pfr_mse_b}
\end{equation}

In this model, the inputs of the neural network are $\mathbf{x} = \{C_{in}, v, z\}$ and $u = C$. The required derivatives in Equation~\eqref{eq:pfr_mse_b} are obtained using automatic differentiation~\cite{Baydin2018Vancouver}.

\subsubsection{Froth Flotation System}
\label{subsubsec:flotation_model} 
Reference \cite{Cubillos1997FlotationControl} proposed a kinetic model for modeling froth flotation dynamics that considers unidirectional material transfer from the pulp phase to the froth phase. As presented in~\cite{nasiri2025}, with the assumption that the volumes of the froth (\( V_f \)) and pulp (\( V_p \)) phases and the densities of the feed (\( \rho_{\textrm{feed}} \)), froth (\( \rho_f \)), and pulp (\( \rho_p \)) phases remain constant throughout the flotation process, the dynamical model in terms of concentrations can be expressed by 
\begin{align}
    \frac{dC_p}{dt} &= C_\textrm{feed} \frac{\rho_\textrm{feed}}{\rho_p} \frac{Q_\textrm{feed}}{V_p} - \frac{Q_t}{V_p} C_p - \frac{R}{\rho_p V_p}, \label{eq:flotation_c_p_dynamic} \\
    \frac{dC_f}{dt} &= - \frac{Q_c}{V_f} C_f + \frac{R}{\rho_f V_f}, \label{eq:flotation_c_f_dynamic}
\end{align}
where \( R \) is the average flotation rate.

For this flotation system, the discrete-time nonlinear operators are given by
\begin{align}
    \CMcal{N}_p[C_{p,n+c_j}] &= -C_\textrm{feed} \frac{\rho_\textrm{feed}}{\rho_p} \frac{Q_\textrm{feed}}{V_p} + \frac{Q_t}{V_p} C_{p,n+c_j} + \frac{R}{\rho_p V_p}, \label{eq:nonlinear_pulp} \\
    \CMcal{N}_f[C_{f,n+c_j}] &= \frac{Q_c}{V_f} C_{f,n+c_j} - \frac{R}{\rho_f V_f}, \label{eq:nonlinear_froth}
\end{align}
where \( C_{p,n+c_j} = C_p(t_n + c_j \Delta t) \) and \( C_{f,n+c_j} = C_f(t_n + c_j \Delta t) \) represent the concentrations at the Runge-Kutta stages.

Two multi-output networks, \( u_p(\mathbf{x}) \) and \( u_f(\mathbf{x}) \), are then used to approximate the solutions of all Runge-Kutta stages and the next time steps
\begin{equation}
\left[u_{p,n+c_1}(\mathbf{x}), \ldots, u_{p,n+c_q}(\mathbf{x}), u_{p,n+1}(\mathbf{x})\right],
\label{eq:network_outputs_C_p}
\end{equation}
and
\begin{equation}
\left[u_{f,n+c_1}(\mathbf{x}), \ldots, u_{f,n+c_q}(\mathbf{x}), u_{f,n+1}(\mathbf{x})\right],
\label{eq:network_outputs_C_f}
\end{equation}
respectively, where $u_p = C_p$, $u_f = C_f$, and $\mathbf{x} = \{C_\textrm{feed}, Q_\textrm{feed}, Q_t, Q_c\}$ represents the input variables. The flotation rate \( R \), which is unknown and potentially nonlinear, is estimated using a separate shallow neural network \( R(u_{p,n+1}(\mathbf{x}), u_{f,n+1}(\mathbf{x})) \). The shallow network \( R(u_{p,n+1}(\mathbf{x}), u_{f,n+1}(\mathbf{x})) \) is constructed not to produce a direct mapping from inputs to a predefined target; instead, it takes the last predicted outputs from \( u_{p}(\mathbf{x}) \) and \( u_{f}(\mathbf{x}) \) to infer the average flotation rate \( R \).

The parameters of the networks \( u_p(\mathbf{x}) \), \( u_f(\mathbf{x}) \), and \( R(\cdot) \), which form the physics-informed neural network, are optimized by minimizing the mean squared error loss specified in Equation~\eqref{eq:cstr_mse}, where
\begin{align}
     \mathrm{MSE}_f &= \frac{1}{N_n} \sum_{j=1}^{q+1} \sum_{i=1}^{N_n} \Big( \big| u_{p,n}^j(\mathbf{x}_{n,i}) - u_{p,n,i}\big|^2 \nonumber \\
     &\hphantom{{}= \frac{1}{N_n} \sum_{j=1}^{q+1} \sum_{i=1}^{N_n} \Big(} + \big|u_{f,n}^j(\mathbf{x}_{n,i}) - u_{f,n,i} \big|^2 \Big), 
     \label{eq:flotation_mse_f}\\
     \mathrm{MSE}_u &= \frac{1}{N_n} \sum_{i=1}^{N_n} \Big( \big|u_{p,n+1}(\mathbf{x}_{n,i}) - u_{p,n+1,i}\big|^2 \nonumber \\
     &\hphantom{{}= \frac{1}{N_n} \sum_{i=1}^{N_n} \Big(} + \big|u_{f,n+1}(\mathbf{x}_{n,i}) - u_{f,n+1,i}\big|^2 \Big).
     \label{eq:flotation_mse_u}
\end{align}

\subsection{Complete System Forecasting Strategy}
\label{subsec:complete_forecasting_strategy}
The multi-step-ahead forecasting approach of this paper for a physical system integrates sequential input forecasting with physics-constrained output prediction. This strategy employs two distinct but interconnected phases: probabilistic forecasting of input variables followed by physics-informed neural networks for multi-step prediction of the system outputs.

Input variable forecasting is performed using four state-space models, categorized into two approaches. For the conventional state-space models discussed in Sections~\ref{subsubsec:matern32_model} and~\ref{subsubsec:ou_model}, the forecasting procedure begins by applying the Kalman filter to process a minimal sequence of observations to obtain a reliable initial state estimate for forecasting. Specifically, the Kalman filter is applied for three steps to obtain the final state estimate $\mathbf{m}$ and covariance $\mathbf{P}$. The three-step sequence is chosen because the three-dimensional augmented Matérn model requires at least three observations for its state to be fully observable. To ensure a fair comparison across all models, the same three-step initialization is also used for the two-dimensional augmented exponential model. These estimated values are then used as the distribution of the initial state for recursive forecasting using the state transition model
\begin{equation}
    \mathbf{x}_k = \mathbf{A}_{k-1} \mathbf{x}_{k-1} + \mathbf{q}_{k-1}, \quad \text{with} \quad \mathbf{q}_{k-1} \sim \CMcal{N}(\mathbf{0}, \mathbf{Q}_{k-1}),
\end{equation}
where the matrices $\mathbf{A}_k$ and $\mathbf{Q}_k$ are defined in the corresponding sections. Multiple trajectories are generated by sampling initial states from $\CMcal{N}(\mathbf{m}, \mathbf{P})$ and propagating them through the state transition model with process noise. The variable of interest for forecasting is the first component of the state vector $\mathbf{x}^{(1)}_k$, representing the predicted input variable.

For the hybrid models discussed in Section~\ref{subsec:hybrid_ssm}, the forecasting procedure employs a different approach that integrates LSTM networks into the state transitions. After applying the hybrid Kalman filter for three steps to obtain the final state estimate $\mathbf{m}$ and covariance $\mathbf{P}$, recursive forecasting proceeds using the state transition model defined in Section~\ref{subsubsec:hybrid_matern},
\begin{equation}
    \mathbf{x}_k = \begin{bmatrix} 
        \operatorname{LSTM}(\mathbf{y}_{k-1}, \mathbf{y}_{k-2}, \mathbf{y}_{k-3}) \\
        \mathbf{a}_2^\top \mathbf{x}_{k-1} \\
        \mathbf{a}_3^\top \mathbf{x}_{k-1}
    \end{bmatrix} + \mathbf{q}_{k-1},
    \label{eq:hybrid_forecast_matern_1}
\end{equation}
and the model defined in Section~\ref{subsubsec:hybrid_ou},
\begin{equation}
    \mathbf{x}_k = \begin{bmatrix} 
        \operatorname{LSTM}(\mathbf{y}_{k-1}, \mathbf{y}_{k-2}, \mathbf{y}_{k-3}) \\
        \mathbf{a}_2^\top \mathbf{x}_{k-1}
    \end{bmatrix} + \mathbf{q}_{k-1},
    \label{eq:hybrid_forecast_OU_1}
\end{equation}
where $\mathbf{a}_2$ and $\mathbf{a}_3$ denote the second and third rows of the state transition matrix $\mathbf{A}_{k-1}$ respectively, $\mathbf{q}_{k-1} \sim \CMcal{N}(\mathbf{0}, \mathbf{Q}_{k-1})$, and the matrices $\mathbf{A}_k$ and $\mathbf{Q}_k$ are defined in the corresponding sections. Consistent with the filtering phase, which requires three observations to estimate $\mathbf{m}$ and $\mathbf{P}$, the LSTM is correspondingly provided with three lagged observations as its input. The $\operatorname{LSTM}(\cdot)$ implements the function $\hat{f}$ in the recursive forecasting scheme of Equation~\eqref{eq:recursive_forecast}. For the first forecast step, its inputs are the three historical measurements used during the Kalman filtering phase. For subsequent steps, the inputs are updated recursively according to Equation~\eqref{eq:recursive_forecast}, with $L=3$ and no exogenous inputs. Consequently, after three forecast steps, the LSTM operates purely on its own previous predictions. As with the conventional state transition models, probabilistic forecasts are generated by sampling initial states from $\CMcal{N}(\mathbf{m}, \mathbf{P})$ and propagating them through the hybrid state transition model, including process noise. The forecasted input variable corresponds to the first state component $\mathbf{x}^{(1)}_k$.

In both approaches, the recursive forecasting generates multiple trajectories that capture the uncertainty in input variable predictions through Monte Carlo simulation. These forecasted input trajectories are then sequentially fed as exogenous inputs to the trained physics-informed neural network models described in Sections~\ref{subsubsec:cstr_model},~\ref{subsubsec:PFR_model}, and~\ref{subsubsec:flotation_model}. For each input trajectory realization, the PINN recursively computes the system output at each time step using the predicted inputs from previous time steps, enabling multi-step prediction of system outputs. 

Algorithm~\ref{alg:complete_forecast} summarizes the complete two-phase forecasting procedure, detailing the recursive input forecasting using state-space models and the subsequent physics-informed output prediction.

\begin{algorithm}[H]
\caption{Complete System Forecasting Procedure}
\label{alg:complete_forecast}
\begin{algorithmic}[1]
\REQUIRE Trained SSMs $\{\mathcal{M}_i\}_{i=1}^{n_{\text{inputs}}}$ (conventional or hybrid), trained PINN, historical measurements $\{y_N^{(i)}, y_{N-1}^{(i)}, y_{N-2}^{(i)}\}_{i=1}^{n_{\text{inputs}}}$, forecast horizon $H$, number of samples $M$
\ENSURE Forecast distributions: inputs $\{\mathcal{X}^{(i)}\}_{i=1}^{n_{\text{inputs}}}$, outputs $\mathcal{U}$

\STATE \textbf{Phase 1: Input Forecasting via State-Space Models}
\FOR{each input variable $i = 1$ \textbf{to} $n_{\text{inputs}}$}
    \STATE Run KF for 3 steps using $\mathcal{M}_i$ $\rightarrow$ obtain $\mathbf{m}^{(i)}, \mathbf{P}^{(i)}$
    \FOR{$j = 1$ \textbf{to} $M$}
        \STATE Sample $\mathbf{x}^{(i,j)}_0 \sim \mathcal{N}(\mathbf{m}^{(i)}, \mathbf{P}^{(i)})$
        \FOR{$h = 1$ \textbf{to} $H$}
            \IF{$\mathcal{M}_i$ is conventional}
                \STATE $\mathbf{x}^{(i,j)}_h = \mathbf{A}_{h-1} \mathbf{x}^{(i,j)}_{h-1} + \mathbf{q}^{(i)}_{h-1}$, 
                \STATE with $\mathbf{q}^{(i)}_{h-1} \sim \mathcal{N}(\mathbf{0}, \mathbf{Q}^{(i)}_{h-1})$
            \ELSE 
                \IF{$h = 1$} \STATE $\mathbf{z} = [y^{(i)}_{N}, y^{(i)}_{N-1}, y^{(i)}_{N-2}]$
                \ELSIF{$h = 2$} \STATE $\mathbf{z} = [\hat{X}^{(i,j)}_{1}, y^{(i)}_{N}, y^{(i)}_{N-1}]$
                \ELSIF{$h = 3$} \STATE $\mathbf{z} = [\hat{X}^{(i,j)}_{2}, \hat{X}^{(i,j)}_{1}, y^{(i)}_{N}]$
                \ELSE \STATE $\mathbf{z} = [\hat{X}^{(i,j)}_{h-1}, \hat{X}^{(i,j)}_{h-2}, \hat{X}^{(i,j)}_{h-3}]$
                \ENDIF
                \STATE $\hat{X}^{(i,j)}_h = \operatorname{LSTM}^{(i)}(\mathbf{z})$
                \STATE $\mathbf{x}^{(i,j)}_h = [\hat{X}^{(i,j)}_h; \mathbf{a}_2^\top \mathbf{x}^{(i,j)}_{h-1}; (\mathbf{a}_3^\top \mathbf{x}^{(i,j)}_{h-1})] + \mathbf{q}^{(i)}_{h-1}$
            \ENDIF
            \STATE $\hat{X}^{(i,j)}_h = (\mathbf{x}^{(i,j)}_h)^{(1)}$
        \ENDFOR
        \STATE $\mathcal{X}^{(i,j)} = \{\hat{X}^{(i,j)}_1, \dots, \hat{X}^{(i,j)}_H\}$
    \ENDFOR
    \STATE $\mathcal{X}^{(i)} = \{\mathcal{X}^{(i,1)}, \dots, \mathcal{X}^{(i,M)}\}$
\ENDFOR

\STATE \textbf{Phase 2: Output Prediction via PINN}
\FOR{$j = 1$ \textbf{to} $M$}
    \FOR{$h = 1$ \textbf{to} $H$}
        \STATE $\mathbf{X}^{(j)}_h = [\hat{X}^{(1,j)}_h, \dots, \hat{X}^{(n_{\text{inputs}},j)}_h]^\top$
        \STATE $\hat{u}^{(j)}_h = \mathrm{PINN}(\mathbf{X}^{(j)}_h)$ 
    \ENDFOR
    \STATE $\mathcal{U}^{(j)} = \{\hat{u}^{(j)}_1, \dots, \hat{u}^{(j)}_H\}$
\ENDFOR

\STATE \textbf{Return} $\{\mathcal{X}^{(i)}\}_{i=1}^{n_{\text{inputs}}}$, $\mathcal{U}$
\end{algorithmic}
\end{algorithm}

\section{Results}
\label{sec:result}
In this section, the performance of the proposed forecasting method is evaluated on three dynamical systems. For each system, the continuous stirred tank reactor (CSTR), axial‑dispersion plug‑flow reactor (ADPFR), and froth flotation, the results are organized to evaluate forecasting performance in three areas. First, the performance of the physics-informed neural networks (PINNs) is assessed against their data-driven feed-forward neural network (FFNN) counterparts for output prediction. Second, the performance of the input forecasting models is evaluated by comparing the hybrid state-transition models with conventional ones. Finally, the complete system forecasting strategy is evaluated, in which the forecasted inputs from all state-transition models are fed sequentially to both the PINNs and their data-driven counterparts to compare their multi-step-ahead forecasting capabilities for the system outputs.

\subsection{Experimental Setup}
\label{subsec:experimental_setup}
This section details the experimental setup and training procedures common to all case studies.

The input for each dynamical system is forecast using the four state-transition models described in Section~\ref{subsec:complete_forecasting_strategy}. The LSTM network used in the hybrid models consists of five layers with 128 neurons per layer and uses the hyperbolic tangent activation function. All input prediction models are optimized using the Adam algorithm~\cite{Kingma2014AdamAM} with a learning rate of $10^{-5}$ to minimize the negative log-likelihood discussed in Section~\ref{subsec:param_est}. An early-stopping criterion with a patience of 1000 iterations and a tolerance of $10^{-5}$ is applied during the training and validation phases, and the model instance achieving the lowest validation negative log-likelihood is preserved for final evaluation on the test dataset.

For output prediction, both physics-informed neural networks and purely data-driven feed-forward neural networks are employed. The models for each system share an identical core architecture for their main networks, consisting of three hidden layers with 256, 512, and 256 neurons, respectively. However, each network in the PINN models outputs \(q+1\) quantities, corresponding to \(q\) Runge-Kutta stages and the solution at the next time step, whereas each network in the data-driven counterparts outputs only the solution at the next time step. Specifically, the PINN for the CSTR system employs a 10-stage scheme (\(q=10\)), thus outputting 11 quantities, while the PINNs for the ADPFR and froth flotation systems employ a 50-stage scheme (\(q=50\)). Note that, for the flotation system, this scheme is implemented by two separate multi-output networks, each outputting 51 quantities. An additional component unique to the flotation PINN is the shallow neural network \( R(u_{p,n+1}(\mathbf{x}), u_{f,n+1}(\mathbf{x})) \) used to approximate the average flotation rate \( R \), which has a single hidden layer containing 100 neurons.

All these networks use the hyperbolic tangent activation function and are optimized using the Adam algorithm with a fixed learning rate of $10^{-5}$. The model parameters are optimized by minimizing the mean squared error, as described in Section~\ref{subsec:pinn} for the PINNs, while the data-driven models are optimized using only the \( \mathrm{MSE}_u \) loss component. The networks employ relatively simple multilayer perceptron architectures without explicit regularization techniques such as dropout or L1/L2 penalty terms. The models are trained on unnormalized data to effectively capture solutions of the physical principles~\cite{nasiri2025}. To avoid excessively long training durations while ensuring optimal model performance, an early stopping procedure with a patience of 30,000 epochs and a tolerance of $10^{-5}$ is used. During the training and validation processes, the model instance achieving the lowest validation mean squared error is preserved, and the final evaluation is conducted on the test dataset.

All models are implemented in Python~\cite{van1995python}, leveraging the PyTorch framework~\cite{paszke2019pytorch} for neural network development. 

The complete system prediction integrates the input forecasting models with the output prediction networks to perform 10-step ahead forecasting. Following the methodology in Section~\ref{subsec:complete_forecasting_strategy}, predicted input trajectories from each state transition model are sequentially fed to the trained neural networks to generate the final multi-step output predictions.

\subsection{Case Study Specifications}
\label{subsec:case_study_spec}
The proposed forecasting approach is evaluated on three distinct dynamical systems, including a continuous stirred tank reactor (CSTR), an axial dispersion plug flow reactor (ADPFR), and an industrial froth flotation process. These systems differ in their governing equations, spatial dimensionality, and data characteristics, providing an extensive evaluation of the forecasting method. The specific configurations, simulation parameters, and dataset characteristics for each case study are detailed below.

\subsubsection{CSTR Dynamical System}
To evaluate the performance of the forecasting models for the CSTR system, the dynamic material balance in Equation~\eqref{eq:cstr_dynamic} is solved numerically using MATLAB's \texttt{ode45} solver, starting from an initial state $C(0) = 1.05$ kmol·m$^{-3}$ and simulated for 10,000 minutes with a multi-frequency signal for the time-varying inlet concentration defined as
\begin{equation*}
    C_{in}(t) = 2.0 + \sum_{n=1}^{6} \left[ a_n \sin\left(\frac{2n\pi t}{T_n}\right) + b_n \cos\left(\frac{2n\pi t}{T_n}\right) \right].
\end{equation*}
The model parameters correspond to a residence time $V/F = 5$ minutes and the second-order reaction rate constant $k = 0.32$ m$^3$·kmol$^{-1}$·min$^{-1}$. From the resulting dataset, 10,000 data points with a sampling interval of $\Delta t = 1.0$ minute are extracted. The dataset is divided in temporal order, the first 7,000 points are used for training, the next 1,500 for validation, and the final 1,500 for testing. For training the output prediction neural networks, PINN and FFNN, only 15 data points are randomly selected from the training set, whereas for training the input forecasting models, all 7,000 training points are used.

\subsubsection{ADPFR Dynamical System}
For the ADPFR system, the inlet concentration $C_{in}(t)$ and superficial velocity $v(t)$ are defined as multi-frequency signals composed of multiple superimposed sinusoidal components with varying periods. The partial differential equation in Equation~\eqref{eq:pfr_dynamic} is solved numerically using MATLAB's \texttt{pdepe} solver. The reactor geometry and kinetic parameters include reactor length $L = 1.0$ m, axial dispersion coefficient $D = 0.01$ m$^2$\text{s}$^{-1}$, and second-order reaction rate constant $k = 0.2$ m$^3$\text{mol}$^{-1}$\text{s}$^{-1}$. The system, with initial condition $C(x,0) = 0$ mol·m$^{-3}$ and Danckwerts boundary conditions defined in Equations~\eqref{eq:danckwerts_inlet} and~\eqref{eq:danckwerts_outlet}, is integrated up to a final time $t = 2,750$ s using a spatial discretization of 1000 nodes.

From the simulation results, 5,500 temporal snapshots with a sampling interval of $\Delta t = 0.5$ s are extracted, corresponding to $1000$ spatial points at each time instance. The dataset is divided in temporal order; the first 4,000 time steps are used for training, the next 900 for validation, and the remaining 600 for testing. Similar to the case of the CSTR, only 150 training data points are randomly selected for training the output prediction networks, while all training points are used for training the input forecasting models.

\subsubsection{Flotation Dynamical System}
The predictive performance of the developed models for forecasting flotation dynamics is evaluated using an industrial dataset supplied by Metso Oyj, corresponding to the first rougher flotation cell of a subcircuit within a gold processing circuit. For this analysis, a dataset generated from a digital twin of the flotation circuit is used. This digital twin dynamically integrates real process data with physics-based models calibrated using historical plant data~\cite{Metso2024Geminex}.

The underlying simulation model, implemented in the HSC‑Sim software environment, is calibrated at the rougher bank level using historical plant data for the concentrate stream. This ensures that the simulated bank-level performance aligns with observed plant behavior. However, the model of the individual rougher cell is based on first principles, and the cell‑specific outputs, concentrate and tailings grades, should be regarded as estimates rather than directly validated measurements.

The input data driving the simulation for this study corresponds to a five‑day operational period from September 2023. It consists of real plant measurements, including feed gold grade and particle size (P80), sampled at approximately 15‑minute intervals via on‑stream analyzers. These real feed characteristics, along with continuously sampled solids flow rate and density, and the recorded time series of control actions, are fed into the digital twin. The simulation, executed with an internal time step of $\Delta t = 5$ s, generates the corresponding dynamic responses for the internal states and output streams of the rougher cell.

From the full simulation, a dataset comprising 30,000 data points is extracted. The constant parameters of the dynamical model in Equations~\eqref{eq:flotation_c_p_dynamic} and~\eqref{eq:flotation_c_f_dynamic} are set to their mean values from this dataset, that is $V_p = 24.86$ m$^{3}$, $V_f = 5.0$ m$^{3}$, $\rho_{\textrm{feed}} = 1.003$ t·m$^{-3}$, $\rho_p = 1.002$ t·m$^{-3}$, and $\rho_f = 1.20$ t·m$^{-3}$. The dataset is partitioned in temporal order, using the first 10,000 data points for training, the next 10,000 data points for validation, and the last 10,000 for testing.

\subsection{System Output Prediction}
\label{subsec:output_forecast}
\begin{figure*}[htbp]
    \centering
    
    \subfloat[CSTR: PINN\label{fig:pinn_cstr}]{
        \includegraphics[width=0.48\linewidth]{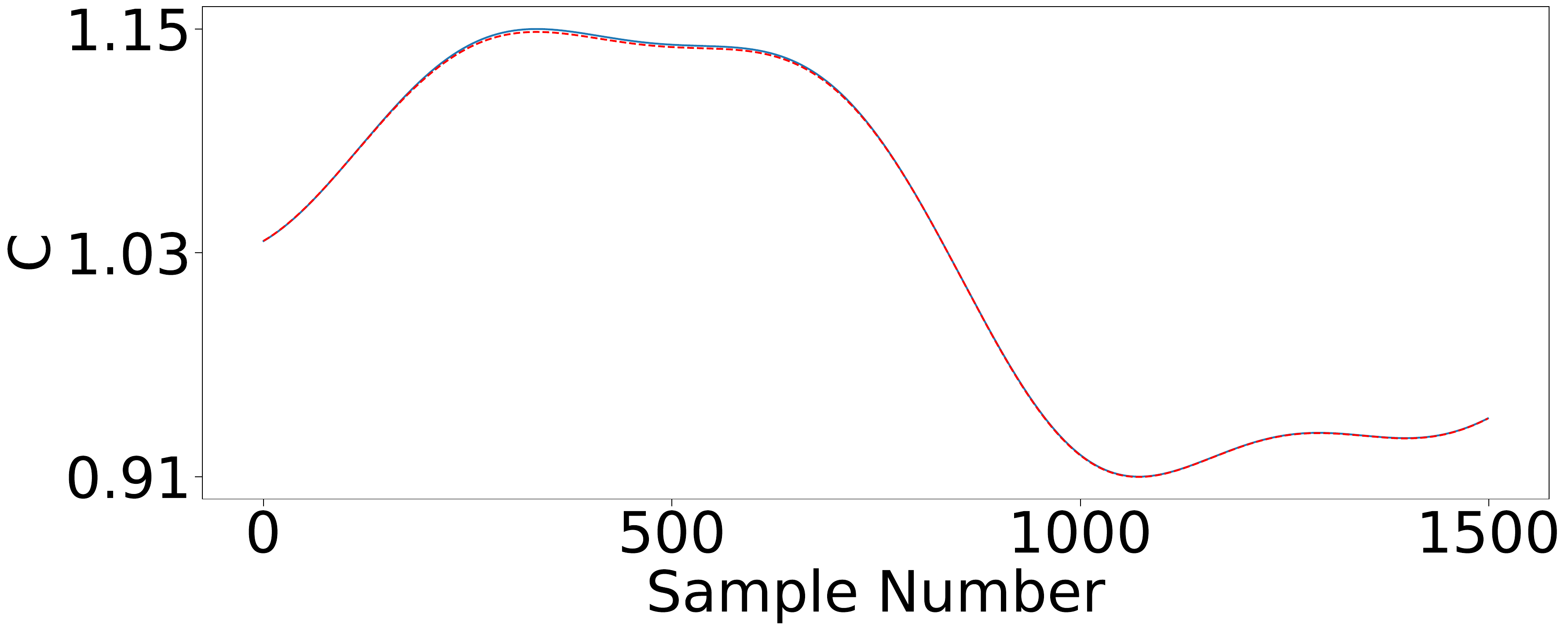}}
    \hfill
    \subfloat[CSTR: FFNN\label{fig:mlp_cstr}]{
        \includegraphics[width=0.48\linewidth]{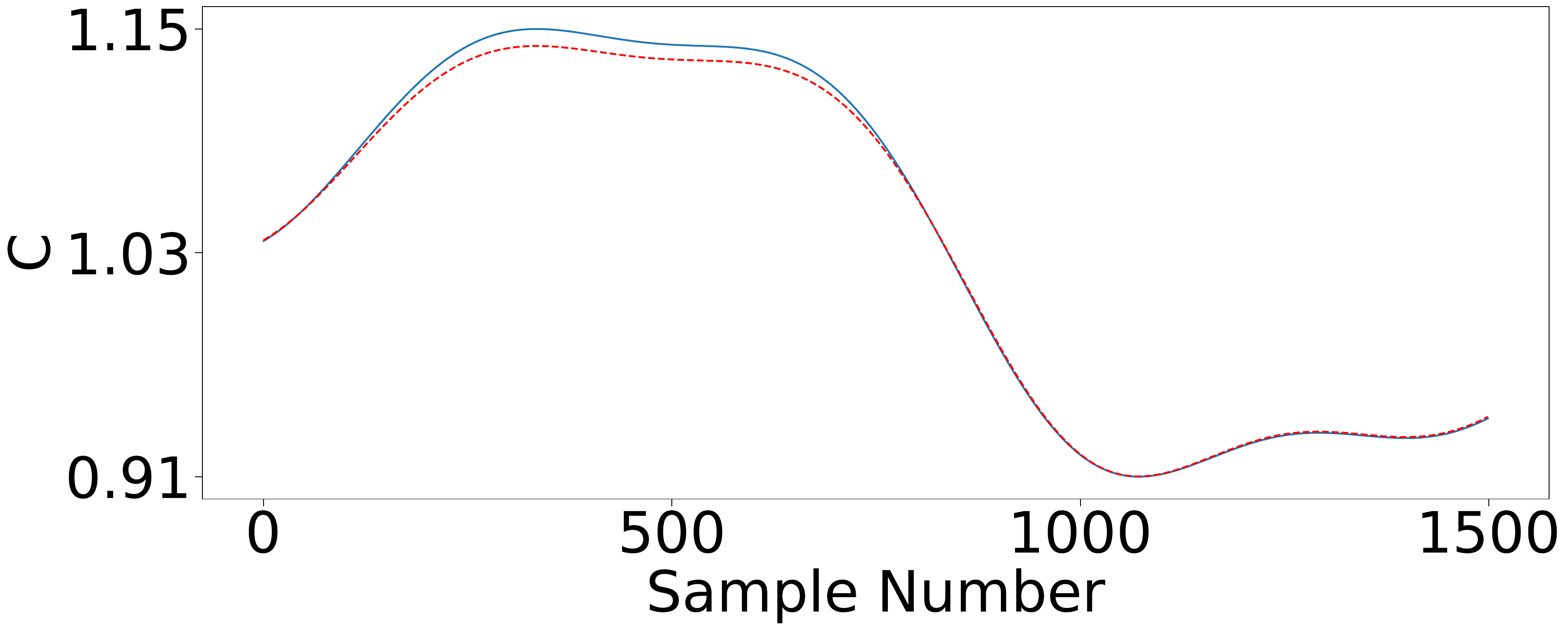}}
        
    \subfloat[ADPFR: PINN\label{fig:pinn_pfr}]{
        \includegraphics[width=0.48\linewidth]{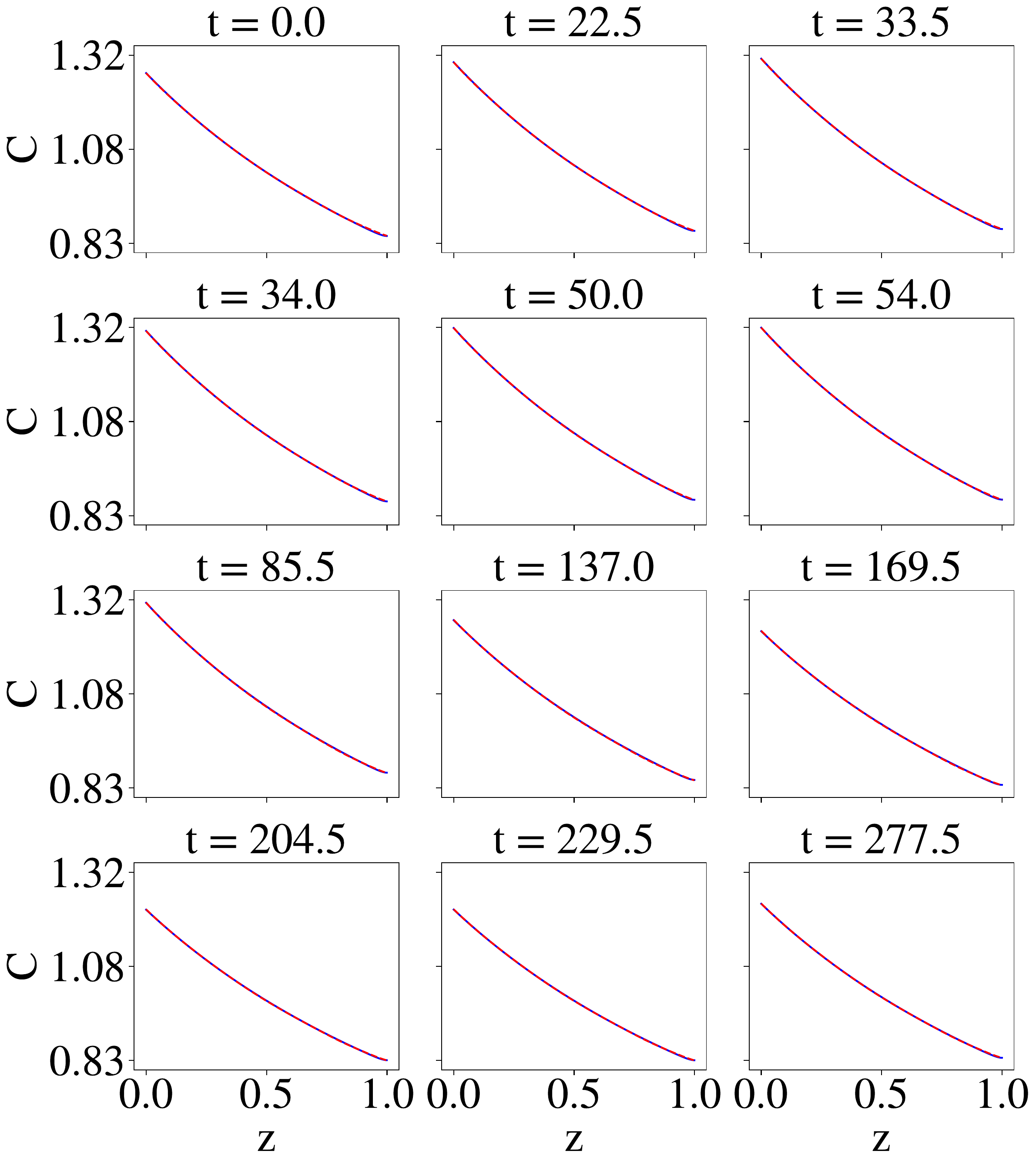}}
    \hfill
    \subfloat[ADPFR: FFNN\label{fig:mlp_pfr}]{
        \includegraphics[width=0.48\linewidth]{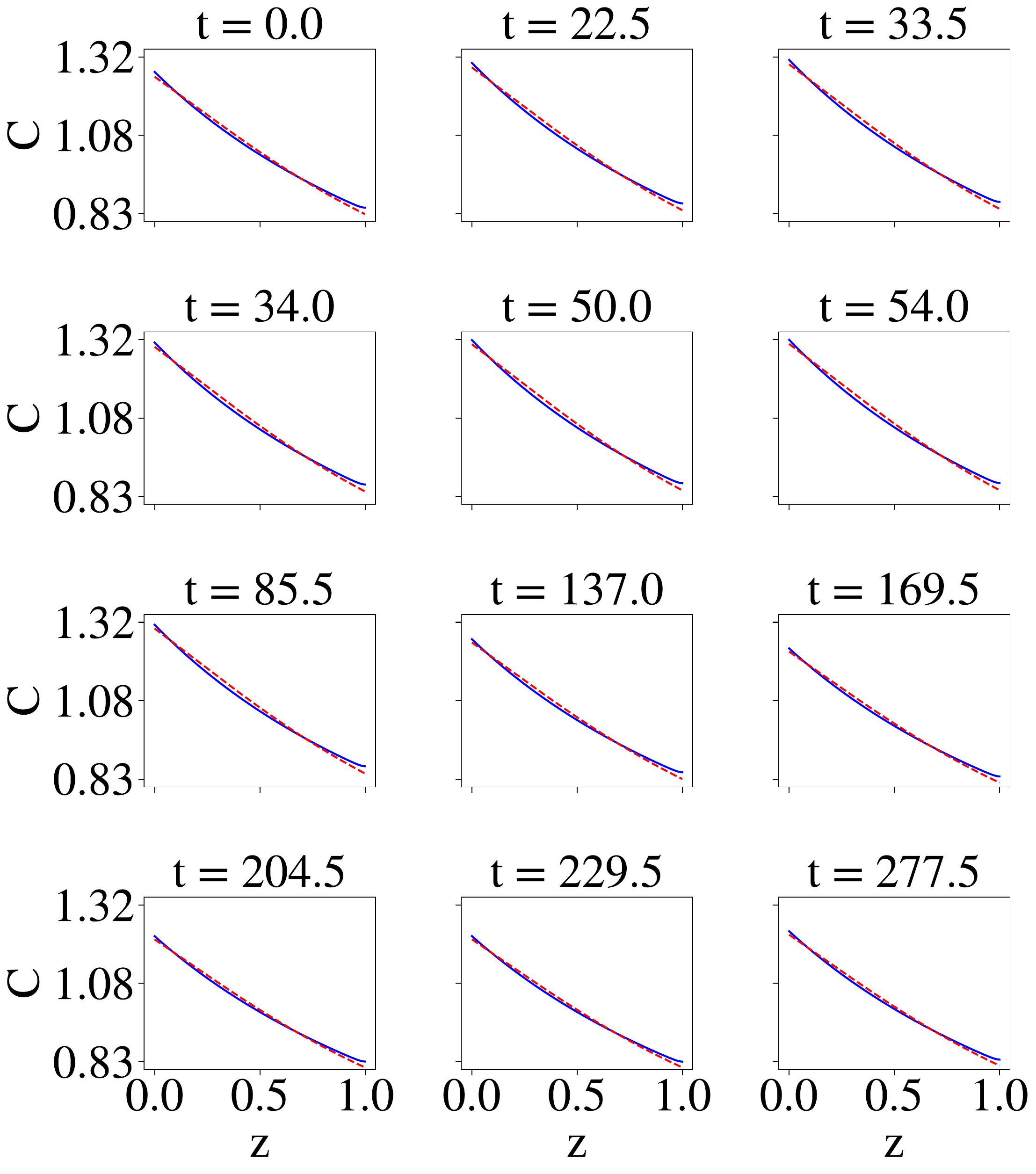}}
        
    \subfloat[Flotation: PINN\label{fig:pinn_flotation}]{
        \includegraphics[width=0.48\linewidth]{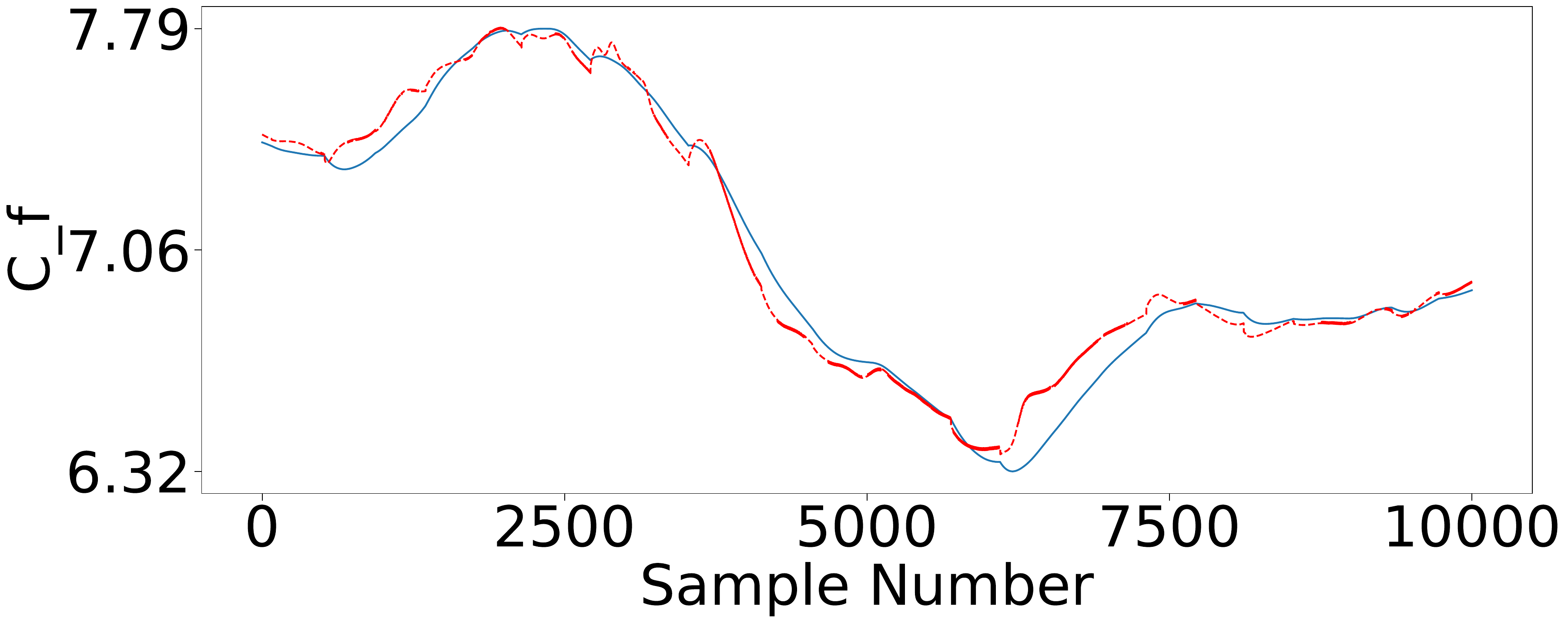}}
    \hfill
    \subfloat[Flotation: FFNN\label{fig:mlp_flotation}]{
        \includegraphics[width=0.48\linewidth]{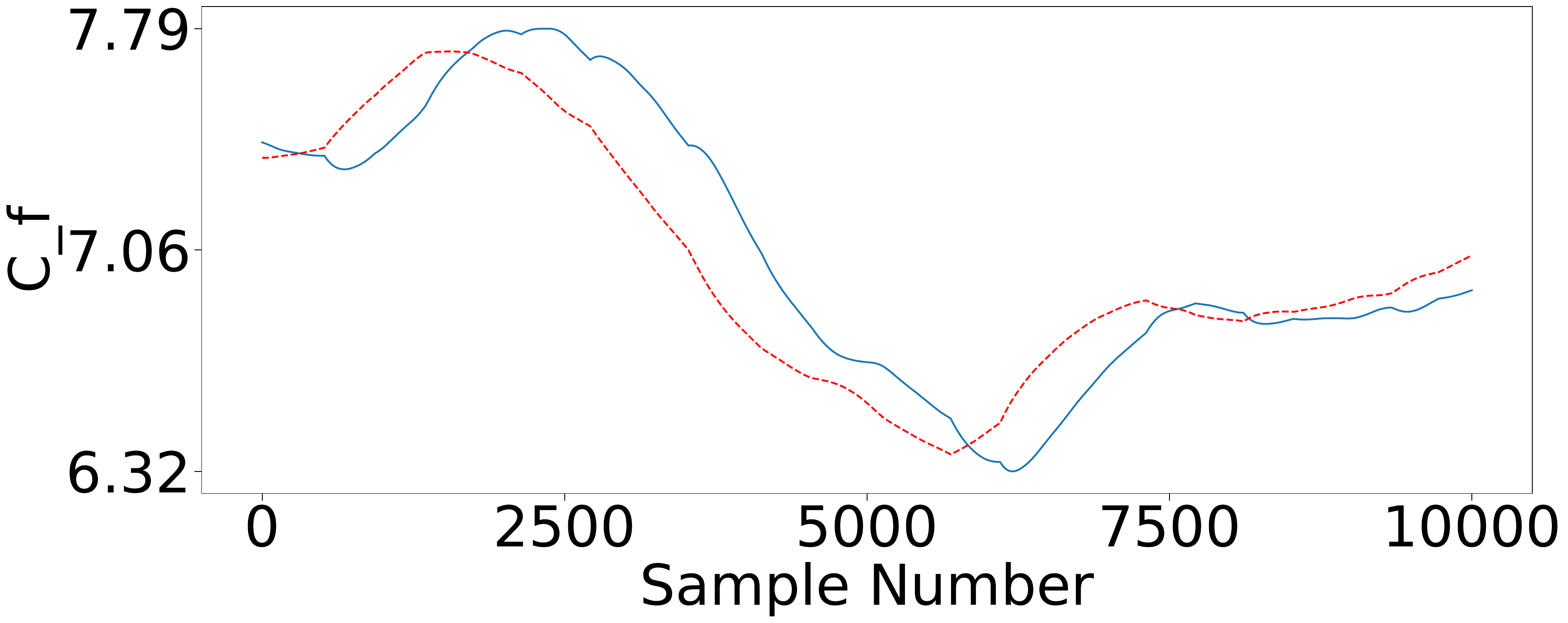}}

    \vspace{0.3cm}
    
    \centering
    \includegraphics[width=\linewidth]{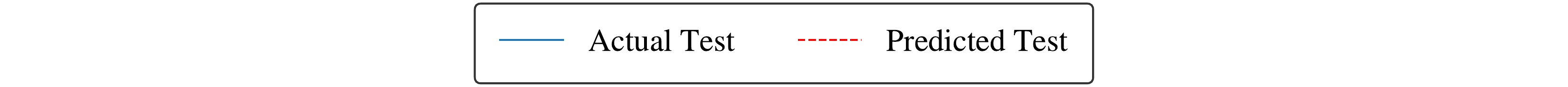}
    
    \caption{Predictive performance of PINNs and their purely data-driven counterparts on test datasets across three dynamical systems: (a-b) CSTR concentration, (c-d) ADPFR concentration profiles at twelve randomly selected temporal snapshots, and (e-f) froth flotation concentrate grade. The left column shows PINN predictions, and the right column shows FFNN predictions. Across all systems, PINNs demonstrate superior accuracy in capturing system dynamics.}
    \label{fig:all_output_predictions}
\end{figure*}

The predictive performance of physics-informed neural networks and data-driven feed-forward neural networks is illustrated in Figure~\ref{fig:all_output_predictions}. The figure compares the models' predictions against the ground truth for (a-b) concentration in the full CSTR test set, (c-d) concentration profiles at twelve randomly selected temporal snapshots of the ADPFR test dataset, and (e-f) concentrate grade on the full flotation test dataset. Across all three systems, both models capture the main dynamic trends, with predictions following the ground truth profiles. However, the physics-informed neural networks consistently demonstrate superior modeling capability, yielding predictions that align more precisely with the actual data. 

The performance of PINNs and data-driven FFNNs in terms of MSE and MRE metrics on both validation and test sets across all three dynamical systems is detailed in Table~\ref{tab:output_prediction_metrics}. PINNs achieve lower error on both validation and test sets compared to their data-driven counterparts for all systems. The lower error obtained by the PINNs on test sets shows their strong generalization ability and capacity to yield more accurate predictions on unseen data.
\begin{table*}[htbp]
\caption{Quantitative evaluation of PINNs and their purely data-driven counterparts across three dynamical systems in terms of MSE and MRE metrics on validation and test datasets.}
\label{tab:output_prediction_metrics}
\centering
\renewcommand{\arraystretch}{1.3}
\begin{tabular}{l l c c c c}
\hline
& & \multicolumn{2}{c}{Validation} & \multicolumn{2}{c}{Test} \\
System & ML Model & MSE & MRE & MSE & MRE \\
\hline
\multirow{2}{*}{CSTR} & PINN & $1.72 \cdot 10^{-8}$ & $1.18 \cdot 10^{-4}$ & $5.65 \cdot 10^{-7}$ & $4.84 \cdot 10^{-4}$ \\
& FFNN & $9.14 \cdot 10^{-7}$ & $2.03 \cdot 10^{-4}$ & $2.19 \cdot 10^{-5}$ & $2.93 \cdot 10^{-3}$ \\
\hline
\multirow{2}{*}{ADPFR} & PINN & $9.09 \cdot 10^{-7}$ & $7.54 \cdot 10^{-4}$ & $4.27 \cdot 10^{-7}$ & $5.08 \cdot 10^{-4}$ \\
& FFNN & $7.08 \cdot 10^{-5}$ & $7.21 \cdot 10^{-3}$ & $7.96 \cdot 10^{-5}$ & $7.73 \cdot 10^{-3}$ \\
\hline
\multirow{2}{*}{Flotation} & PINN & $3.04 \cdot 10^{-3}$ & $6.01 \cdot 10^{-3}$ & $4.22 \cdot 10^{-3}$ & $6.77 \cdot 10^{-3}$ \\
& FFNN & $1.10 \cdot 10^{-2}$ & $1.07 \cdot 10^{-2}$ & $3.59 \cdot 10^{-2}$ & $2.15 \cdot 10^{-2}$ \\
\hline
\end{tabular}
\end{table*}

\subsection{Input Forecasting Performance}
\label{subsec:input_forecast}
\begin{figure*}[htbp]
    \centering

    \subfloat[Exponential\label{fig:cstr_input_ou_noise}]{
    \includegraphics[width=0.48\linewidth]{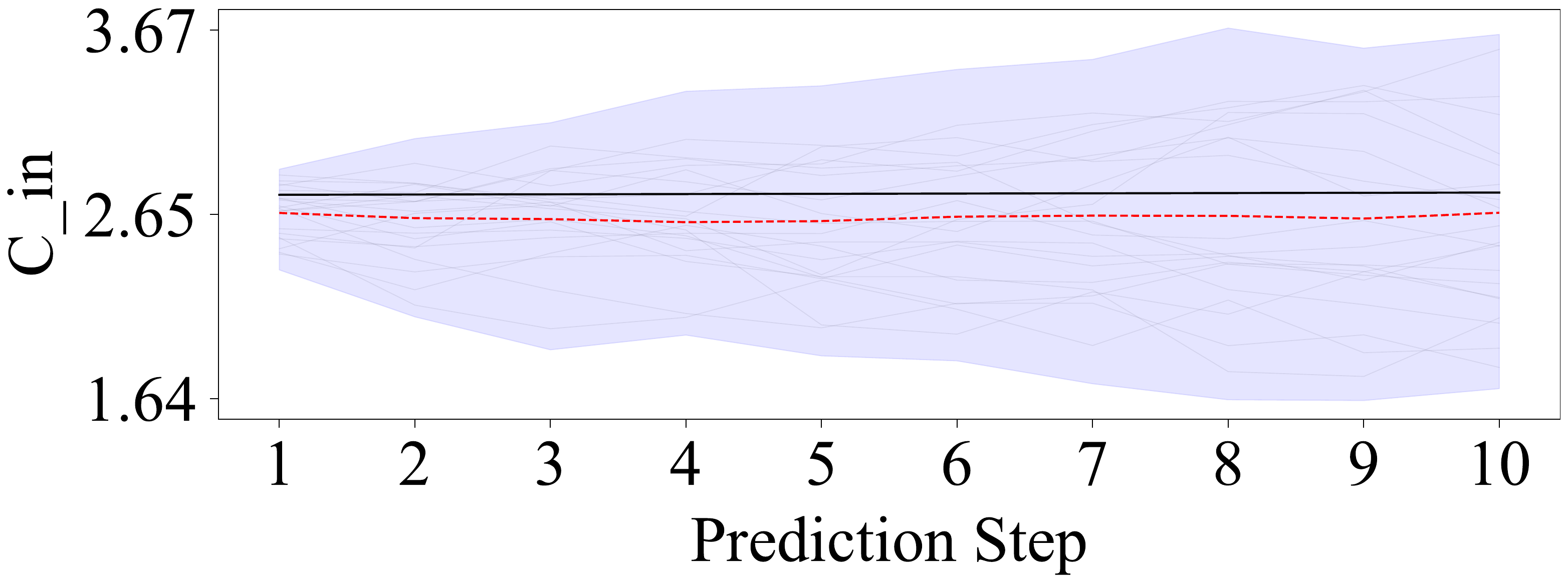}}
    \hfill
    \subfloat[Matérn\label{fig:cstr_input_matern_noise}]{
    \includegraphics[width=0.48\linewidth]{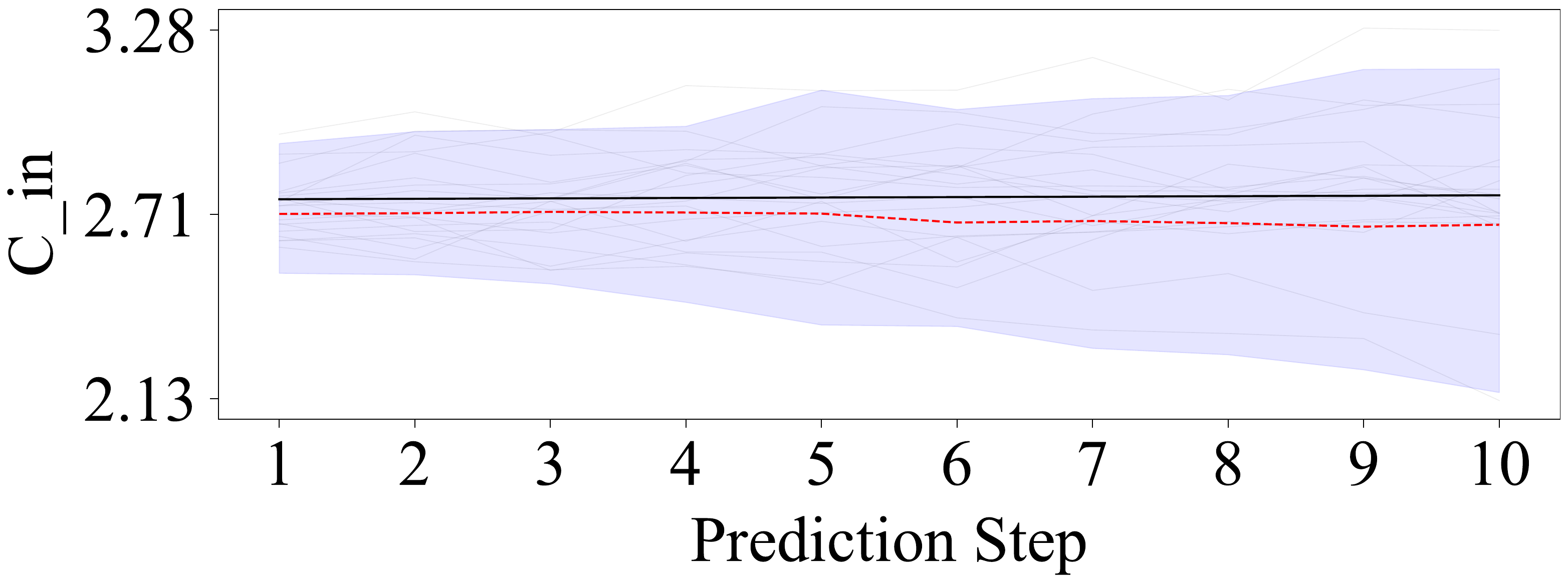}}
        
    \subfloat[Hybrid Exponential\label{fig:cstr_input_hybrid_ou}]{
    \includegraphics[width=0.48\linewidth]{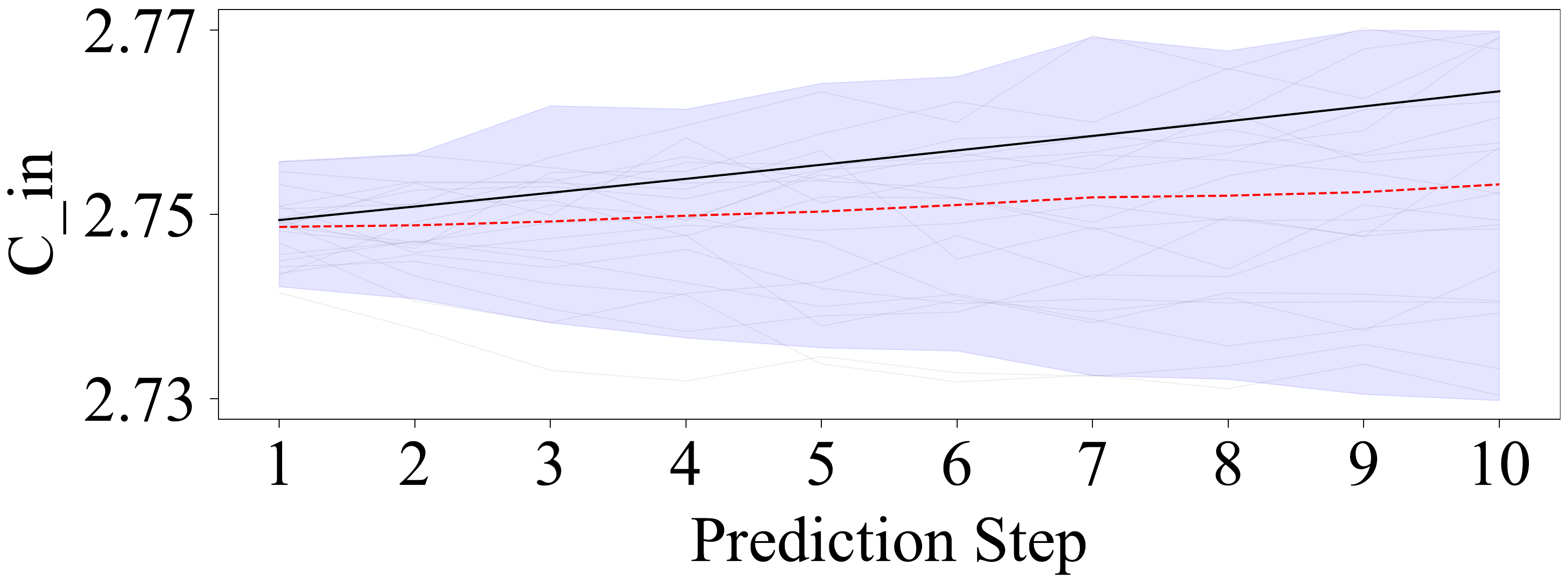}}
    \hfill
    \subfloat[Hybrid Matérn\label{fig:cstr_input_hybrid_matern}]{
    \includegraphics[width=0.48\linewidth]{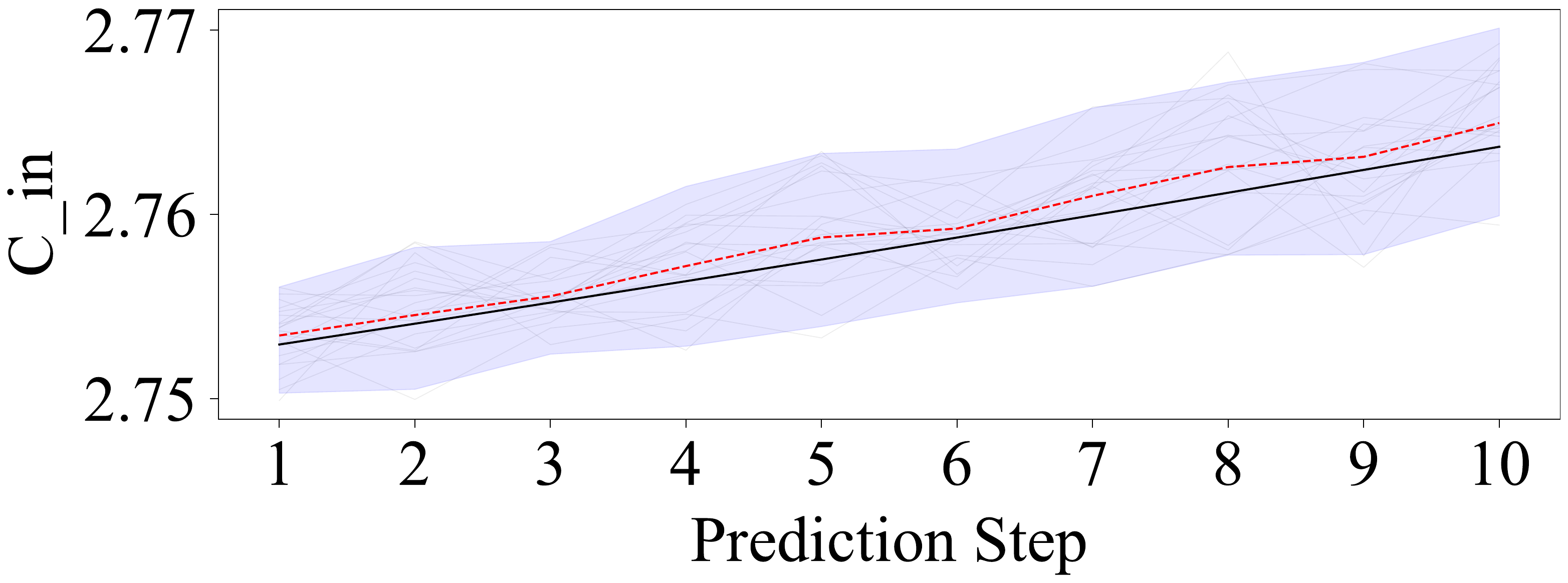}}

    \vspace{0.3cm}
    
    \centering
    \includegraphics[width=\linewidth]{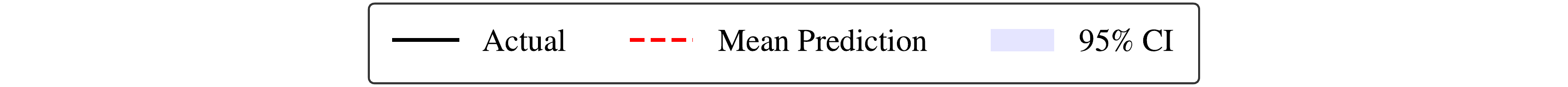}

    \caption{Performance of state transition models in predicting CSTR inlet concentration ($C_{in}$) over a 10-step horizon. Conventional models are compared with hybrid LSTM-integrated counterparts. The shaded region shows the 95\% confidence interval (CI).}
    \label{fig:input_cstr}
\end{figure*}

\begin{figure*}[htbp]
    \centering

    \subfloat[Exponential\label{fig:pfr_input_ou_noise}]{
    \includegraphics[width=0.48\linewidth]{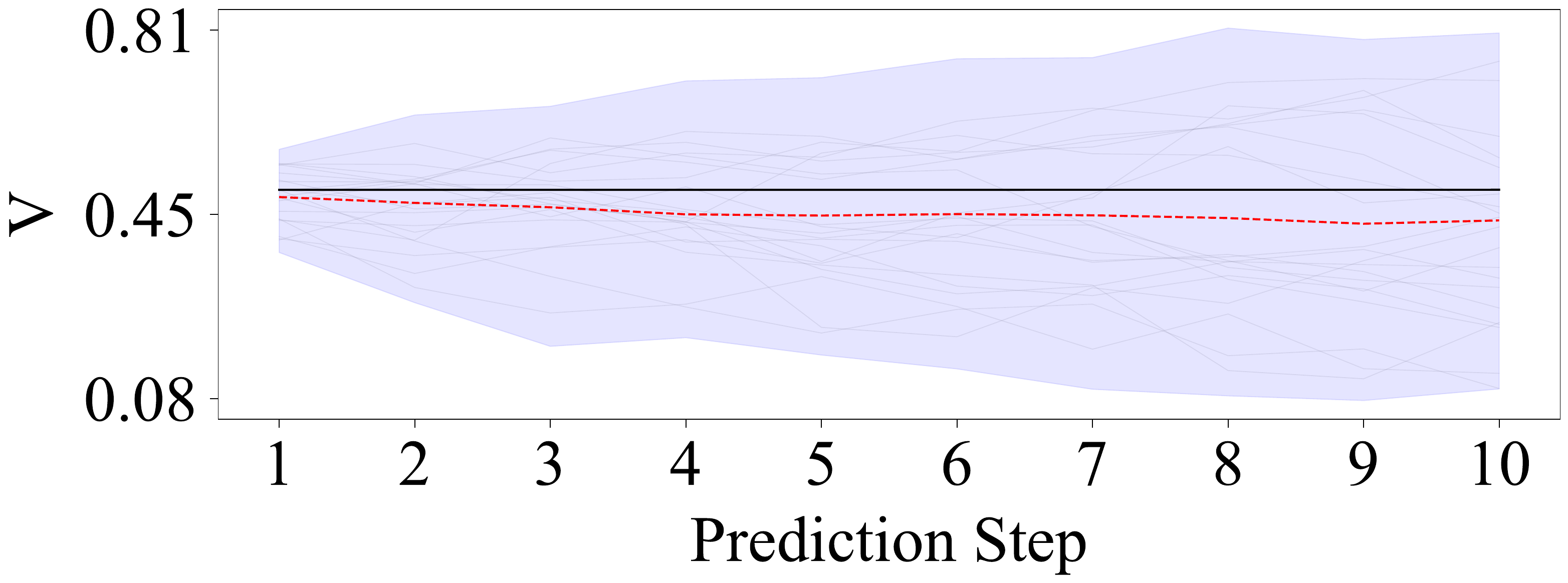}}
    \hfill
    \subfloat[Matérn\label{fig:pfr_input_matern_noise}]{
    \includegraphics[width=0.48\linewidth]{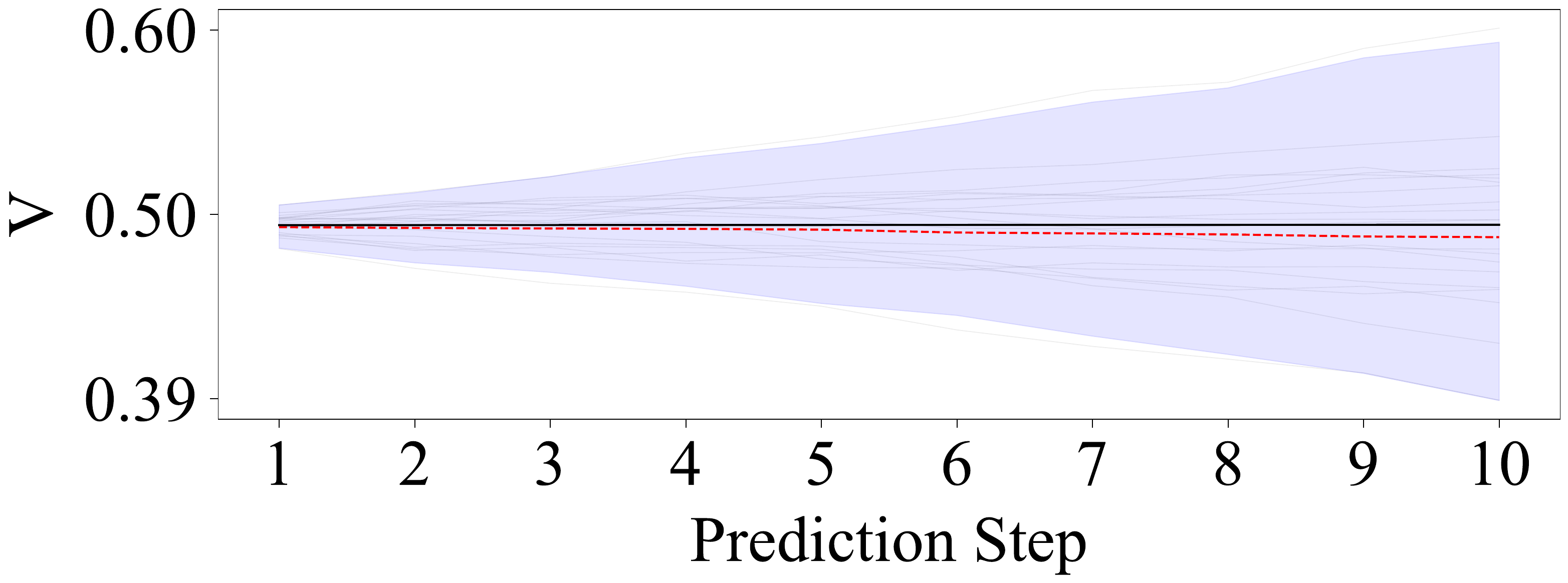}}
        
    \subfloat[Hybrid Exponential\label{fig:pfr_input_hybrid_ou}]{
    \includegraphics[width=0.48\linewidth]{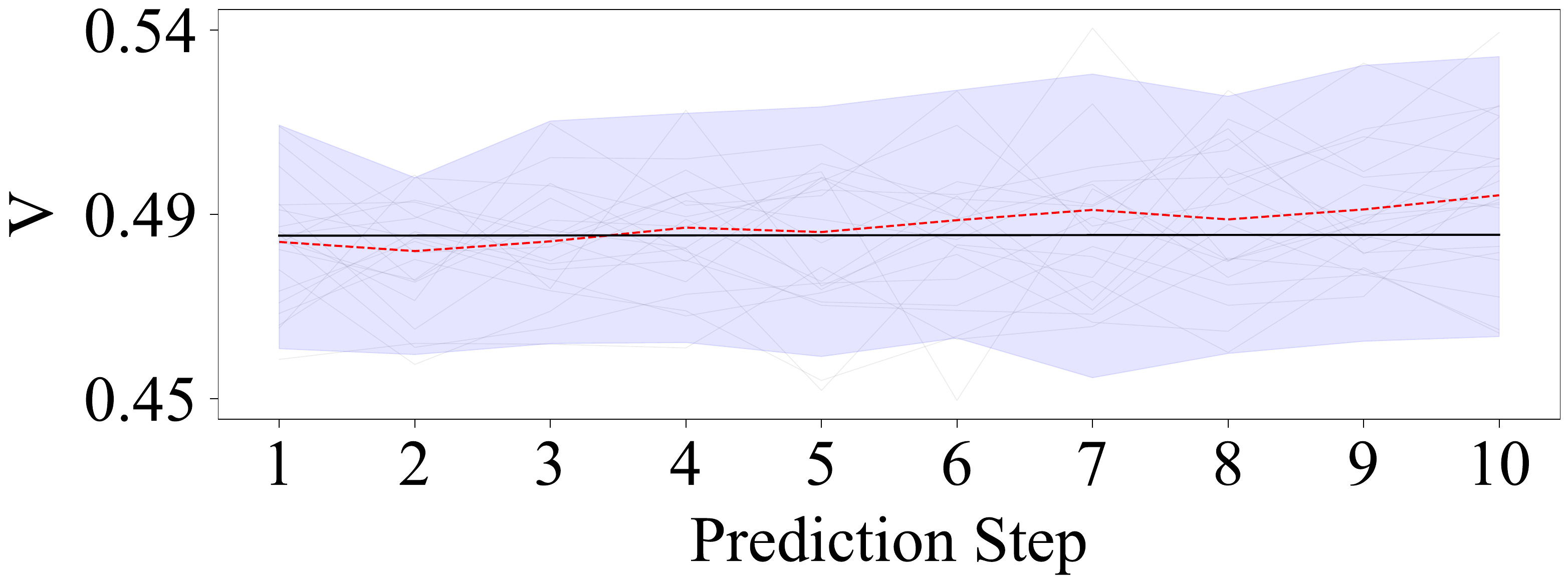}}
    \hfill
    \subfloat[Hybrid Matérn\label{fig:pfr_input_hybrid_matern}]{
    \includegraphics[width=0.48\linewidth]{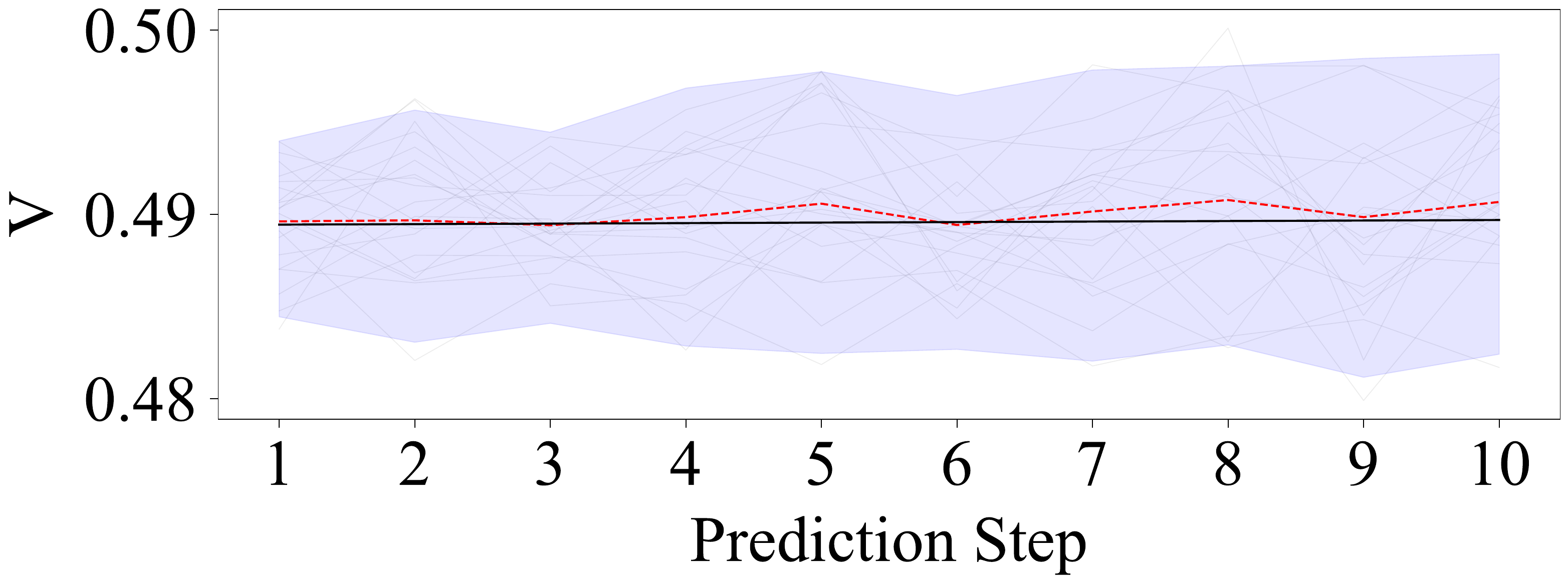}}

    \vspace{0.3cm}
    
    \centering
    \includegraphics[width=\linewidth]{Figures/legend_GP_95.pdf}

    \caption{Comparison of state transition models for 10-step ahead forecasting of ADPFR superficial velocity ($v$). Mean predictions of conventional and LSTM‑integrated models are shown with the 95\% confidence interval.}
    \label{fig:input_pfr_v}
\end{figure*}

\begin{figure*}[htbp]
    \centering

    \subfloat[Exponential\label{fig:flotation_input_ou_noise}]{
    \includegraphics[width=0.48\linewidth]{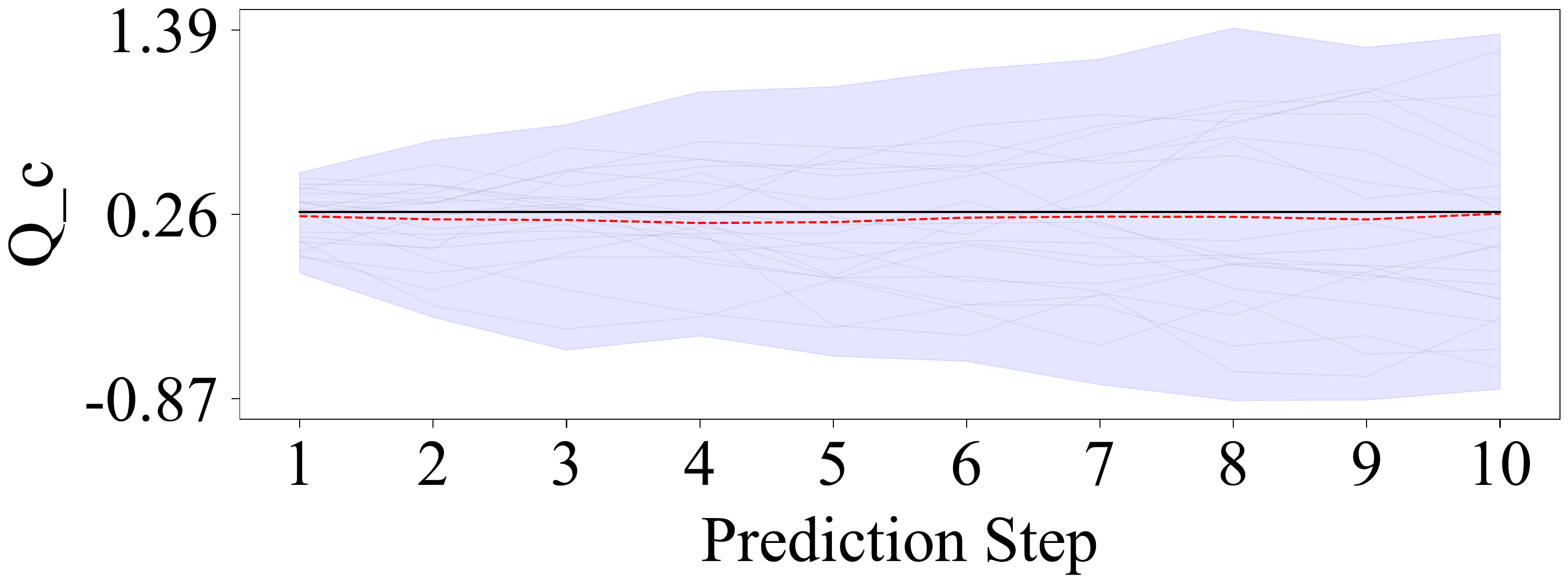}}
    \hfill
    \subfloat[Matérn\label{fig:flotation_input_matern_noise}]{
    \includegraphics[width=0.48\linewidth]{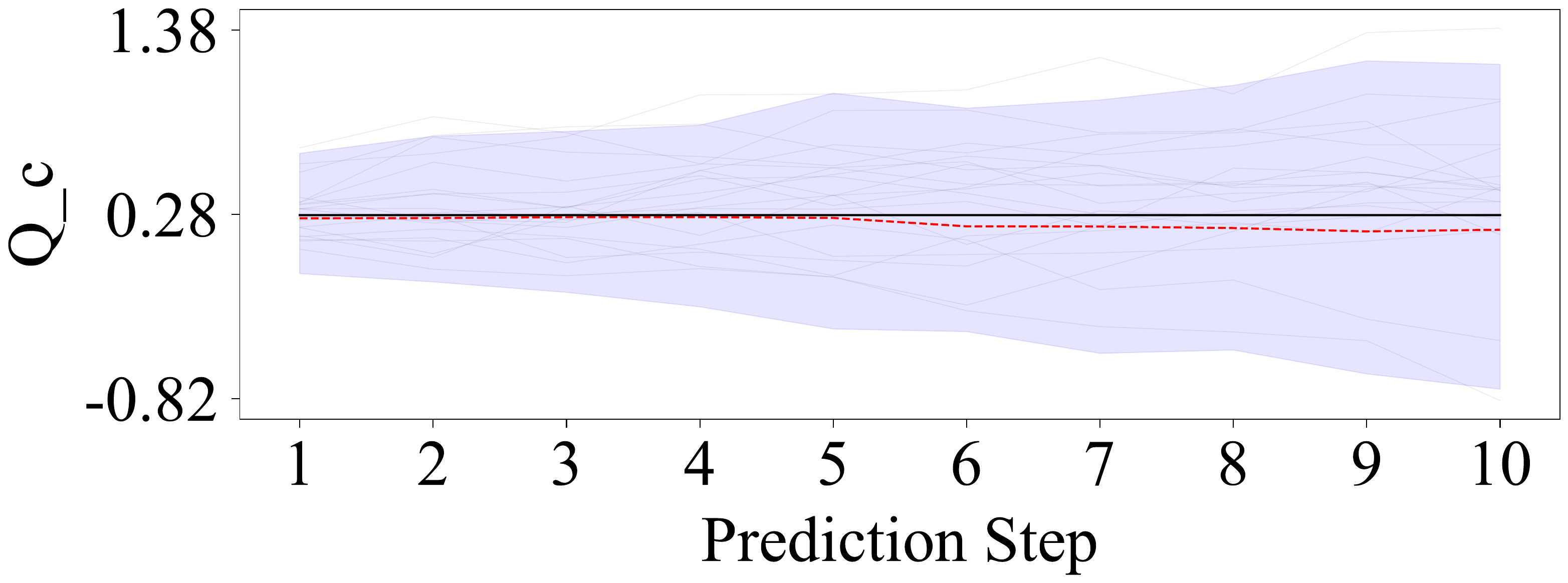}}
        
    \subfloat[Hybrid Exponential\label{fig:flotation_input_hybrid_ou}]{
    \includegraphics[width=0.48\linewidth]{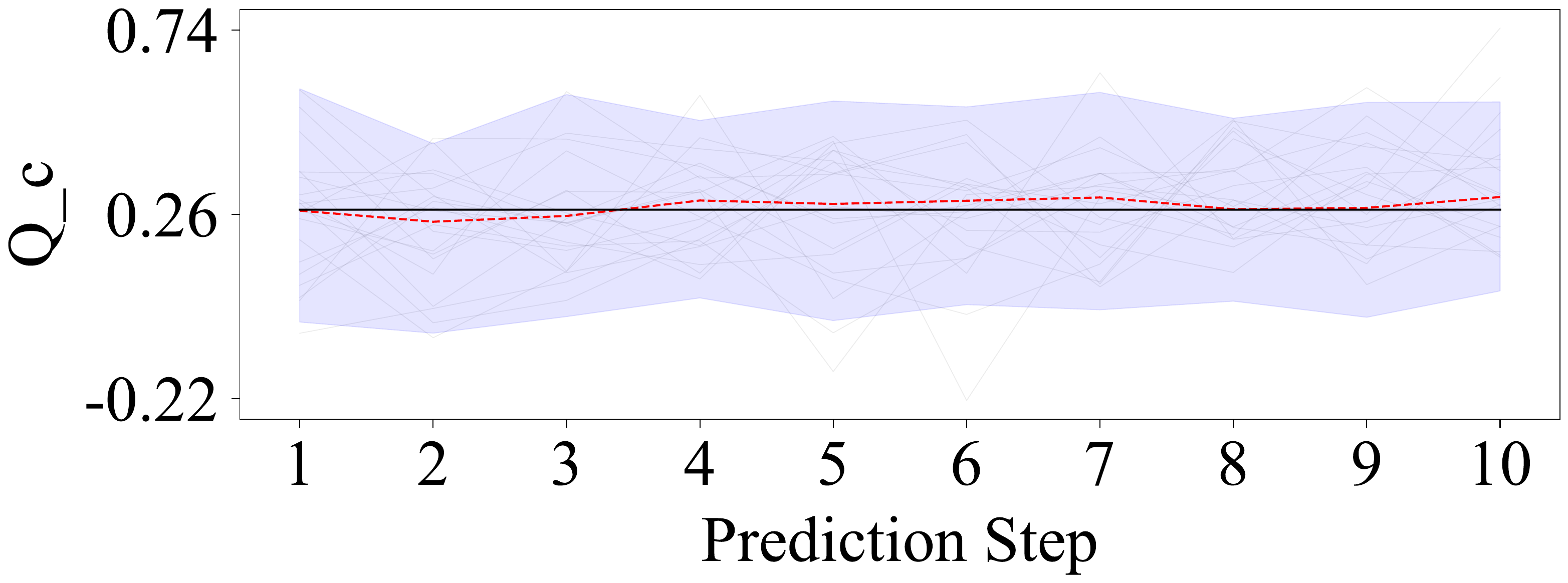}}
    \hfill
    \subfloat[Hybrid Matérn\label{fig:flotation_input_hybrid_matern}]{
    \includegraphics[width=0.48\linewidth]{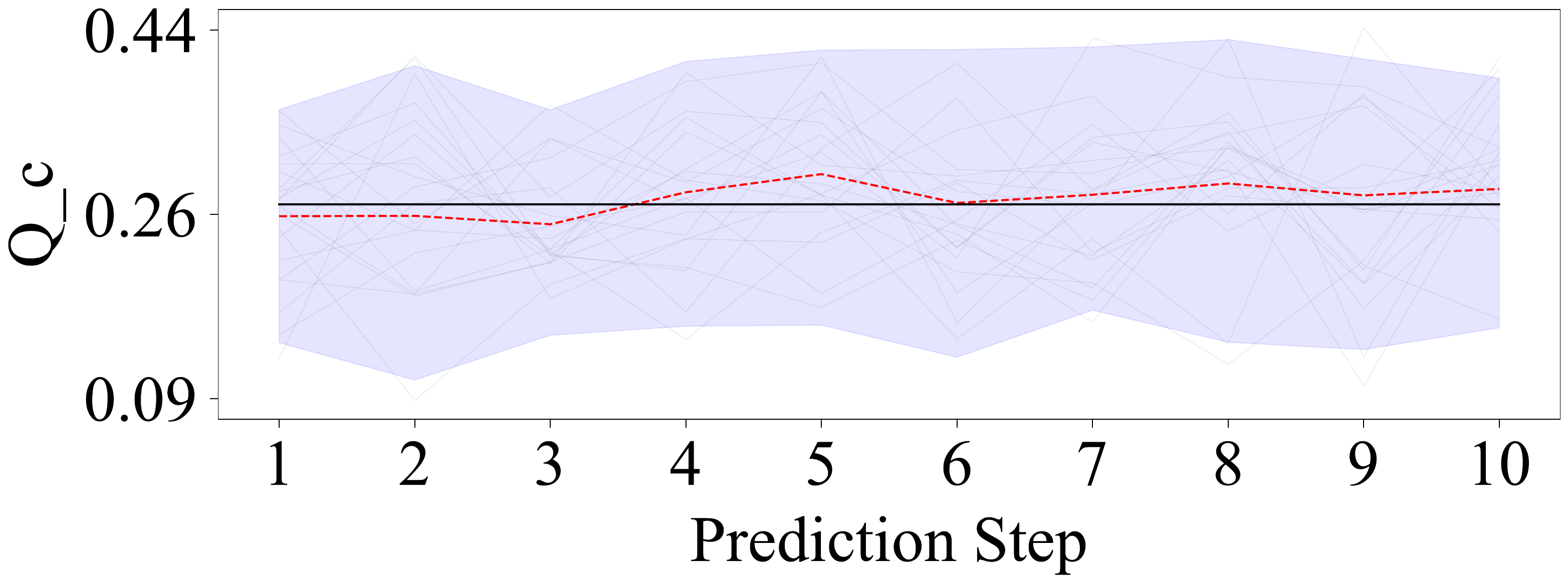}}

    \vspace{0.3cm}
    
    \centering
    \includegraphics[width=\linewidth]{Figures/legend_GP_95.pdf}

    \caption{Forecasting performance for flotation concentrate flow rate ($Q_c$) using state transition models over 10 steps. Both conventional and LSTM‑integrated approaches are shown with the 95\% confidence interval.}
    \label{fig:input_flotation_Q_c}
\end{figure*}

In addition to predicting system outputs, the time-varying inputs are forecast using the four state transition models described in Section~\ref{subsec:complete_forecasting_strategy}. The forecast inputs include inlet concentration ($C_{in}$) for the CSTR system; superficial velocity ($v$) and inlet concentration ($C_{in}$) for the ADPFR system; and feed flow rate ($Q_\textrm{feed}$), tail flow rate ($Q_t$), concentrate flow rate ($Q_c$), and feed grade ($C_\textrm{feed}$) for the flotation system. Figures~\ref{fig:input_cstr}, \ref{fig:input_pfr_v}, and~\ref{fig:input_flotation_Q_c} illustrate representative 10‑step ahead forecasting performance for $C_{in}$ (CSTR), $v$ (ADPFR), and $Q_c$ (flotation), respectively, with probabilistic uncertainty quantification.

The models are first initialized by applying the Kalman filter to three data points of the test dataset to obtain the initial state distribution. The figures show 10‑step ahead forecasts on the subsequent test data points, comparing conventional state transition models with their hybrid LSTM-integrated counterparts. The results demonstrate that across all systems, all models are capable of generating forecasts over the prediction horizon, with mean predictions following the ground truth profiles. However, the LSTM-integrated models produce significantly narrower uncertainty bounds than the conventional models, indicating reduced forecast uncertainty.

\begin{figure*}[htbp]
    \centering

    \subfloat[CSTR: $C_{in}$\label{fig:input_likelihood_cstr}]{
    \includegraphics[width=0.32\linewidth]{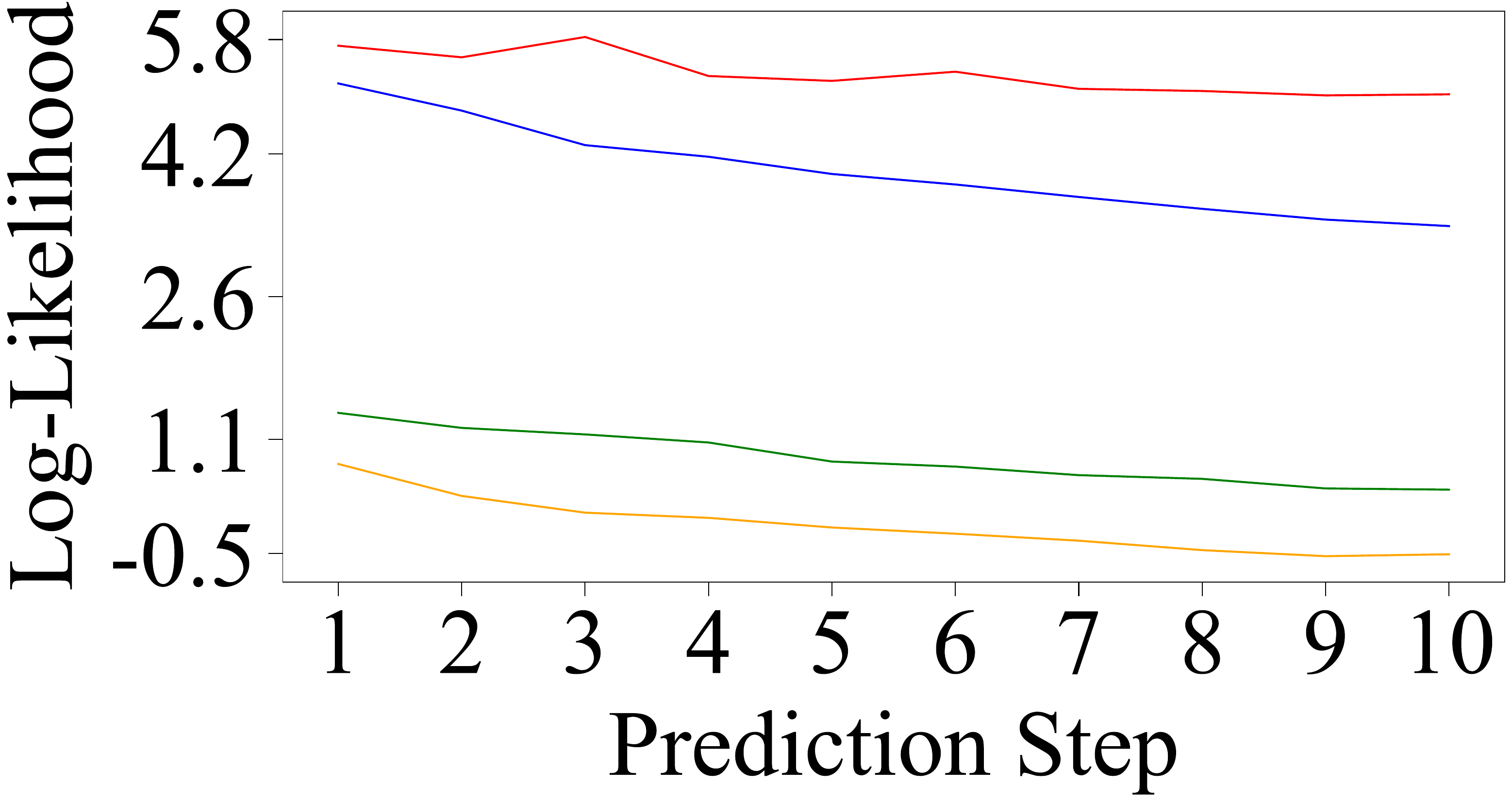}}
    \hfill
    \subfloat[ADPFR: $v$\label{fig:input_likelihood_pfr_v}]{
    \includegraphics[width=0.32\linewidth]{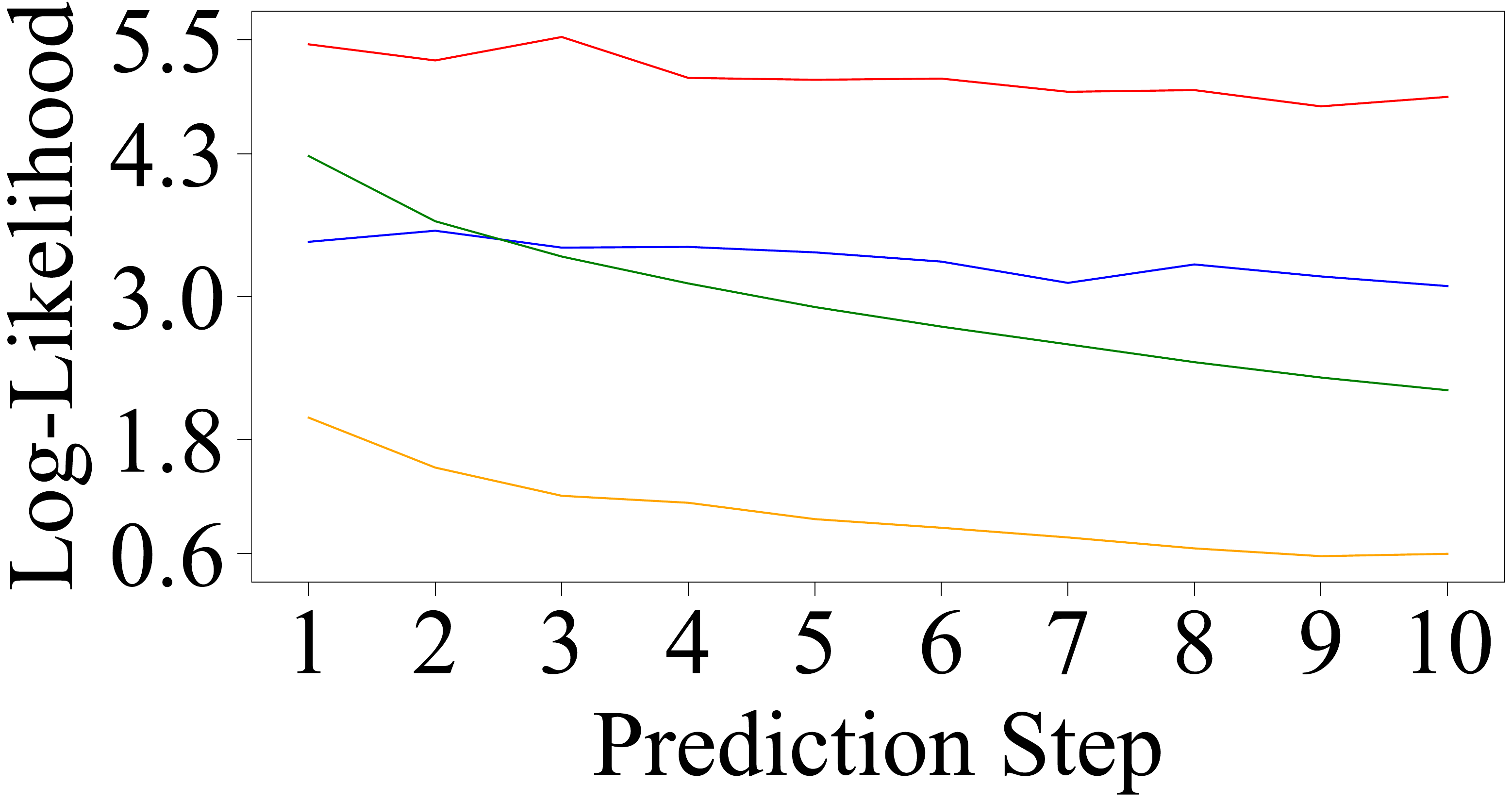}}
    \hfill
    \subfloat[Flotation: $Q_c$\label{fig:input_likelihood_flotation_Q_c}]{
    \includegraphics[width=0.32\linewidth]{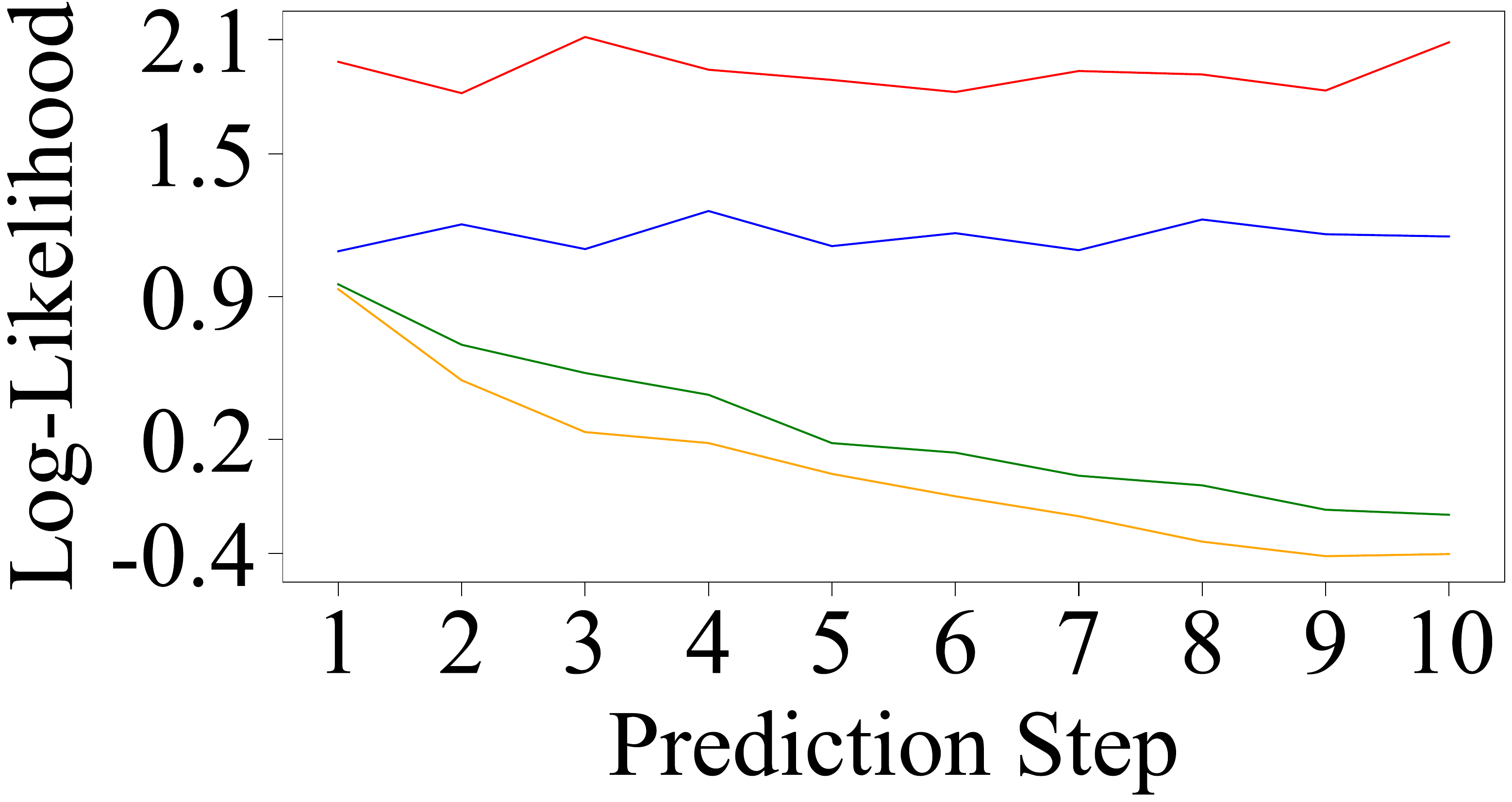}}

    \vspace{0.3cm}
    
    \centering
    \includegraphics[width=\linewidth]{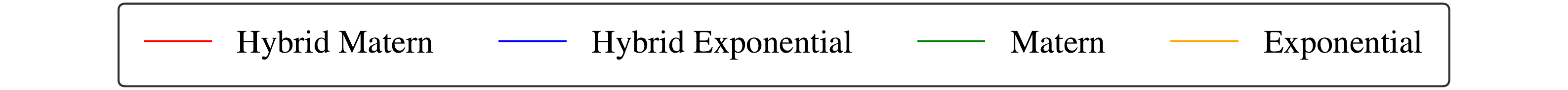}

    \caption{Log-likelihood comparison of input forecasting models across three dynamical systems over 10 prediction steps: (a) CSTR inlet concentration $C_{in}$, (b) ADPFR superficial velocity $v$, and (c) flotation concentrate flow rate $Q_c$. Higher values indicate better probabilistic performance. LSTM-integrated models consistently achieve higher log-likelihood than conventional models, with the hybrid Matérn model performing best across all systems.}
    \label{fig:input_likelihoods}
\end{figure*}

The input forecasting models are compared probabilistically in terms of log-likelihood across all three systems over the 10 forecast steps. Figures~\ref{fig:input_likelihood_cstr},~\ref{fig:input_likelihood_pfr_v}, and~\ref{fig:input_likelihood_flotation_Q_c} show these comparisons for the CSTR, ADPFR, and flotation systems, respectively. Across all systems, the LSTM-integrated state transition models obtain higher log-likelihood values than conventional standalone models throughout the forecast horizon, indicating they offer a more accurate probabilistic representation of the true distribution of the inputs. Among them, the hybrid Matérn model achieves the highest values, demonstrating superior probabilistic performance.

The quantitative evaluation of input forecasting models using MSE and MAE metrics for 10‑step ahead mean predictions on the test subset is presented in Table~\ref{tab:MSE_performance_inputs}. Across all systems, hybrid models outperform their conventional counterparts. The hybrid Matérn model achieves the lowest errors in each case, followed by the hybrid exponential model. The ordering of model performance is in accordance with the log-likelihood results in Figure~\ref{fig:input_likelihoods}, demonstrating that the hybrid approach provides superior forecasting accuracy both in terms of point estimates and predictive distributions.
\begin{table}[htbp]
\caption{Quantitative evaluation of input forecasting models across three dynamical systems. MSE and MAE metrics are reported for 10‑step ahead mean predictions on the test subset.}
\label{tab:MSE_performance_inputs}
\centering
\renewcommand{\arraystretch}{1.3}
\begin{tabular}{l l c c}
\hline
System & Model & MSE & MAE \\
\hline
\multirow{4}{*}{CSTR} 
  & Exponential & $1.40 \cdot 10^{-2}$ & $1.17 \cdot 10^{-1}$ \\
  & Matérn & $3.82 \cdot 10^{-3}$ & $5.89 \cdot 10^{-2}$ \\
  & Hybrid Exponential & $2.39 \cdot 10^{-5}$ & $4.32 \cdot 10^{-3}$ \\
  & Hybrid Matérn & $8.50 \cdot 10^{-7}$ & $8.43 \cdot 10^{-4}$ \\
\hline
\multirow{4}{*}{ADPFR}
  & Exponential & $1.88 \cdot 10^{-3}$ & $4.11 \cdot 10^{-2}$ \\
  & Matérn & $1.55 \cdot 10^{-5}$ & $3.47 \cdot 10^{-3}$ \\
  & Hybrid Exponential & $1.43 \cdot 10^{-5}$ & $3.39 \cdot 10^{-3}$ \\
  & Hybrid Matérn & $2.30 \cdot 10^{-7}$ & $3.75 \cdot 10^{-4}$ \\
\hline
\multirow{4}{*}{Flotation}
  & Exponential & $1.50 \cdot 10^{-3}$ & $3.58 \cdot 10^{-2}$ \\
  & Matérn & $2.69 \cdot 10^{-3}$ & $4.26 \cdot 10^{-2}$ \\
  & Hybrid Exponential & $3.83 \cdot 10^{-4}$ & $1.65 \cdot 10^{-2}$ \\
  & Hybrid Matérn & $1.94 \cdot 10^{-4}$ & $1.23 \cdot 10^{-2}$ \\
\hline
\end{tabular}
\end{table}

\subsection{Integrated Multi-Step Forecasting}
\label{subsec:complete_forecast}
\begin{figure*}[htbp]
    \centering

    \subfloat[PINN-Exponential\label{fig:cstr_pinn_exp}]{
    \includegraphics[width=0.48\linewidth]{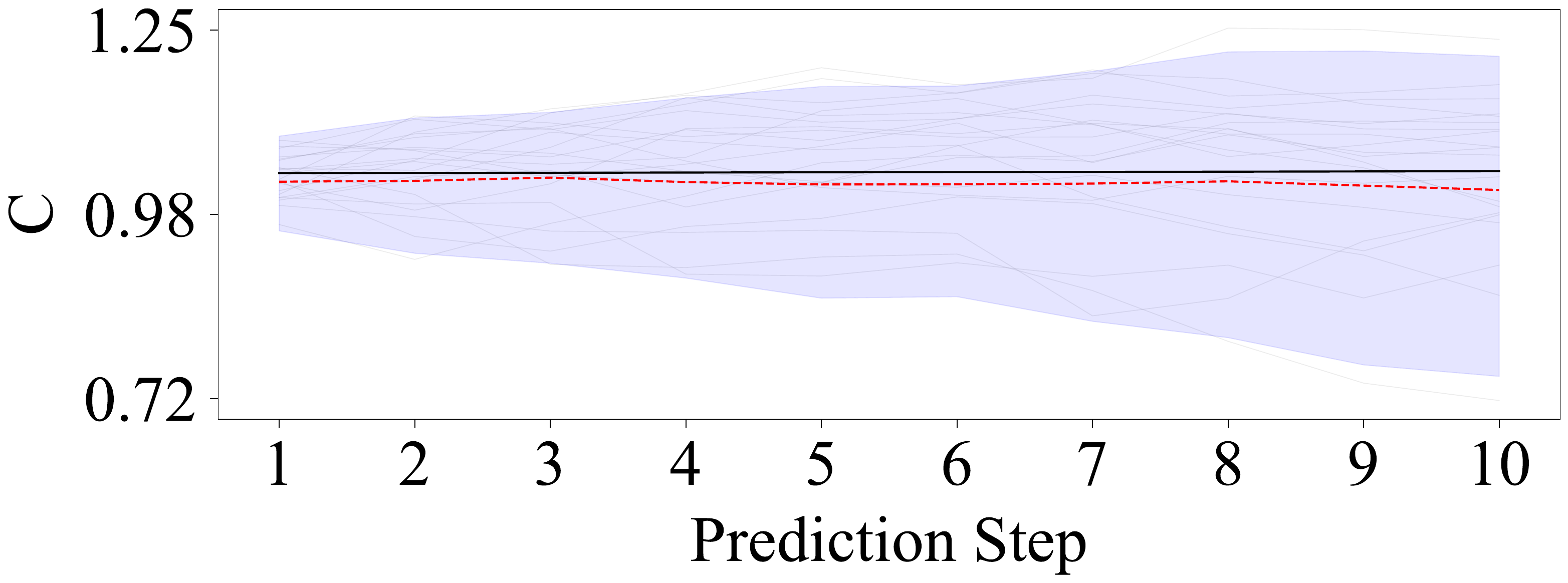}}
    \hfill
    \subfloat[FFNN-Exponential\label{fig:cstr_ffnn_exp}]{
    \includegraphics[width=0.48\linewidth]{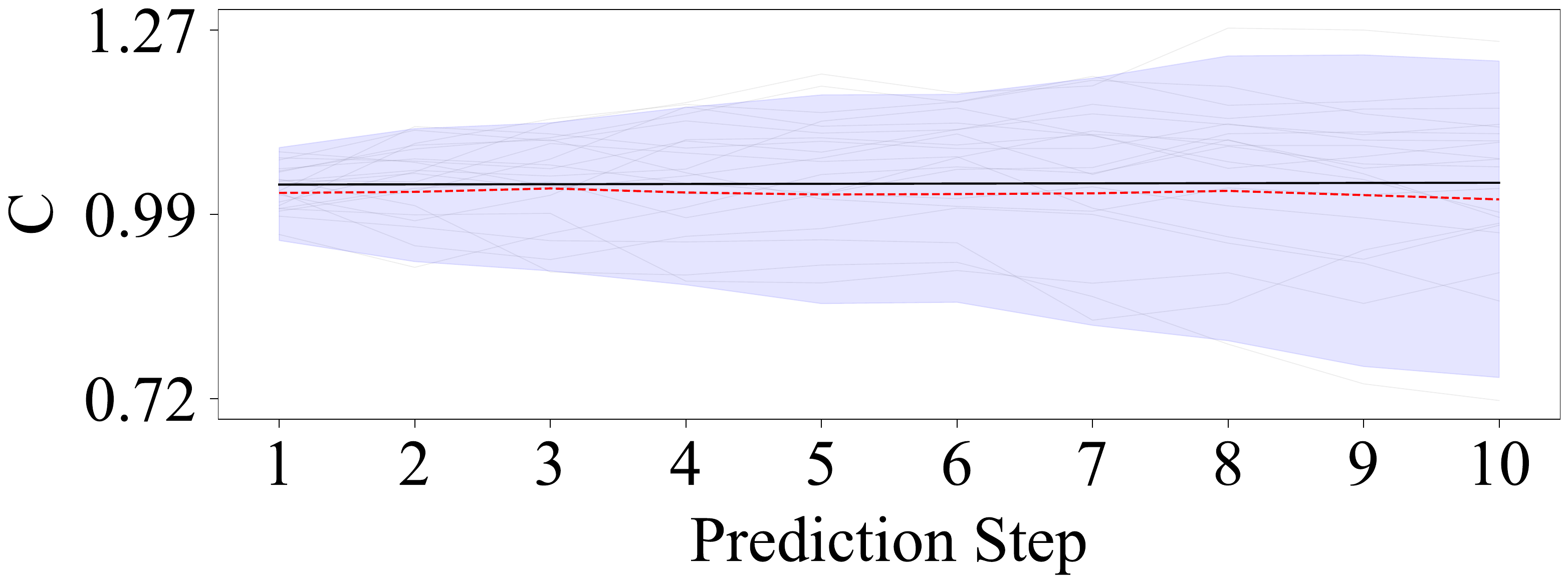}}

    \subfloat[PINN-Matérn\label{fig:cstr_pinn_matern}]{
    \includegraphics[width=0.48\linewidth]{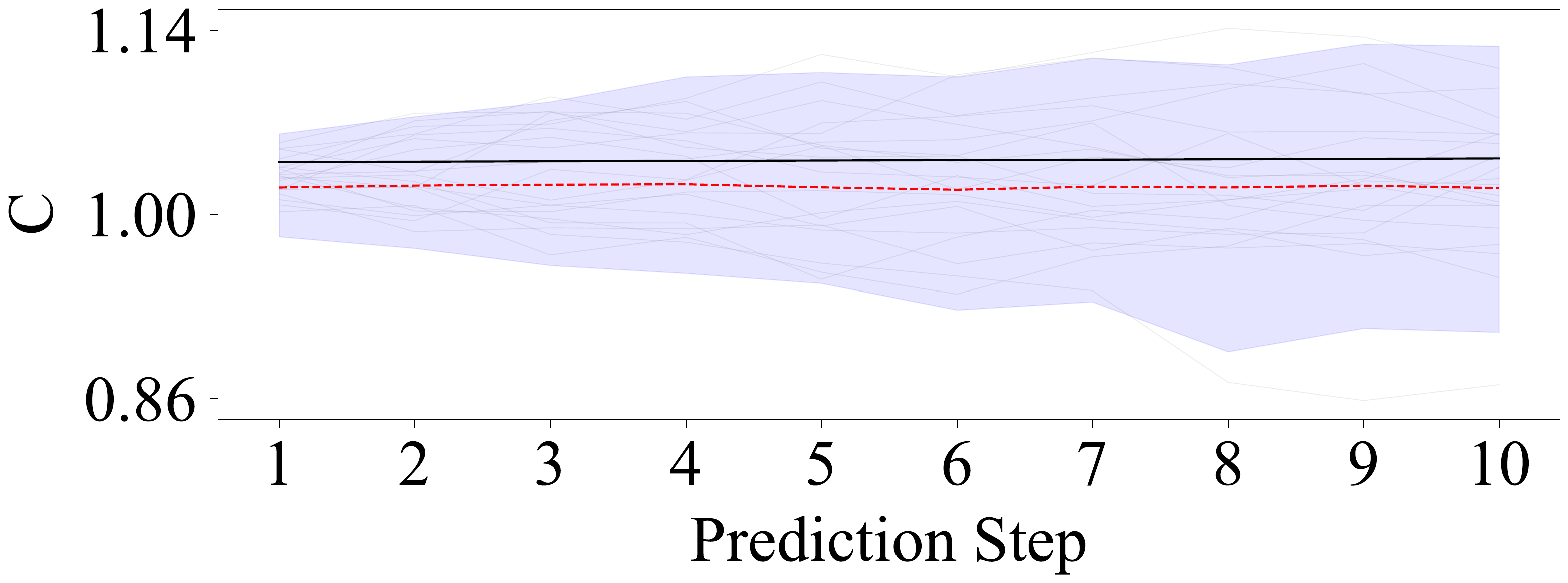}}
    \hfill
    \subfloat[FFNN-Matérn\label{fig:cstr_ffnn_matern}]{
    \includegraphics[width=0.48\linewidth]{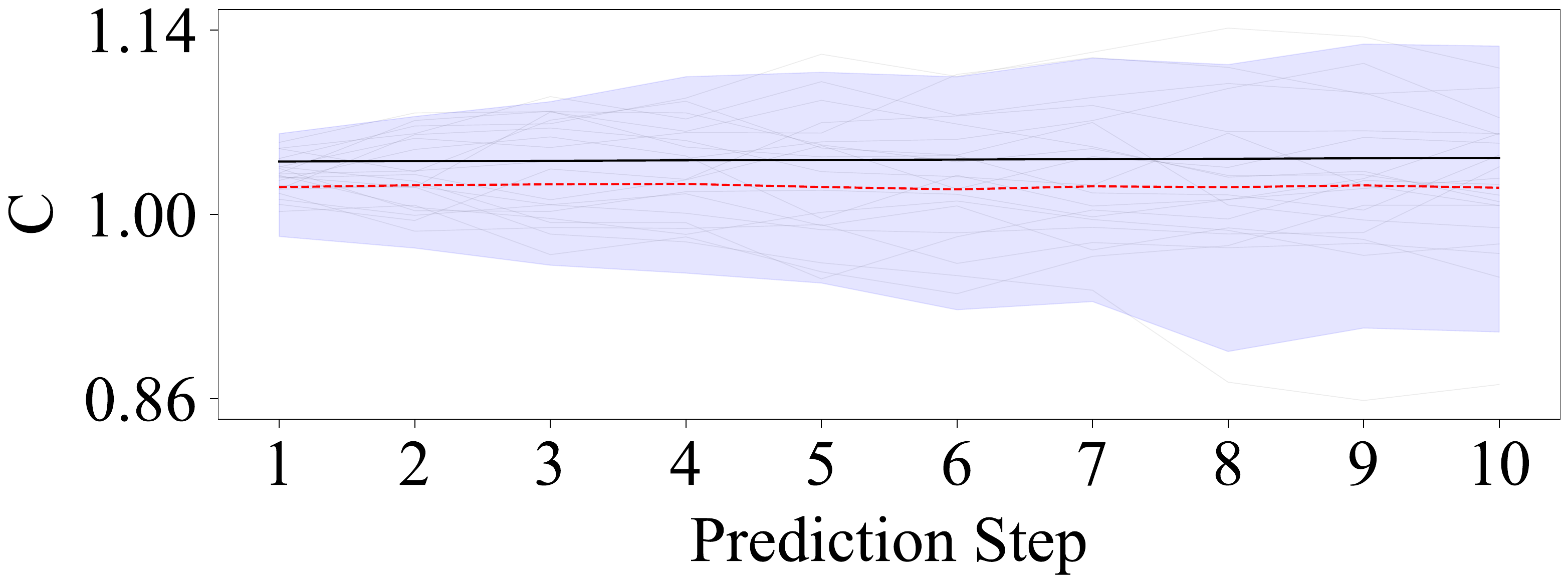}}
    
    \subfloat[PINN-Hybrid Exponential\label{fig:cstr_pinn_hybrid_exp}]{
    \includegraphics[width=0.48\linewidth]{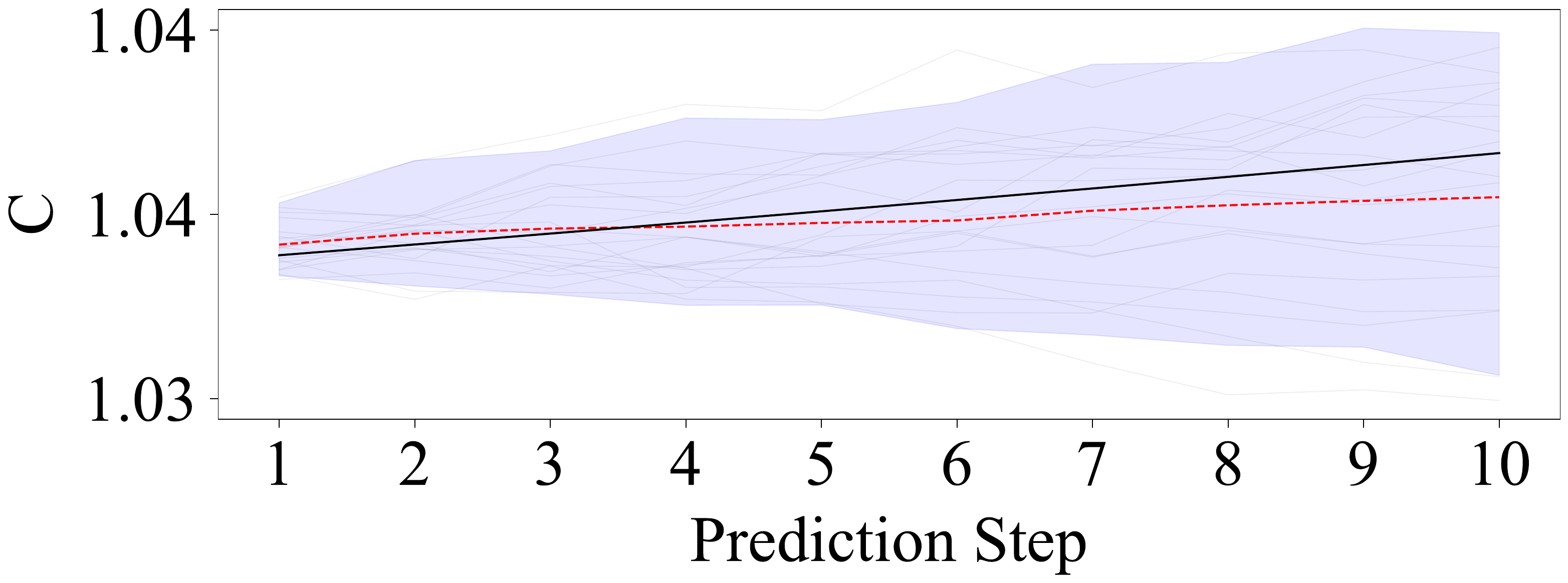}}
    \hfill
    \subfloat[FFNN-Hybrid Exponential\label{fig:cstr_ffnn_hybrid_exp}]{
    \includegraphics[width=0.48\linewidth]{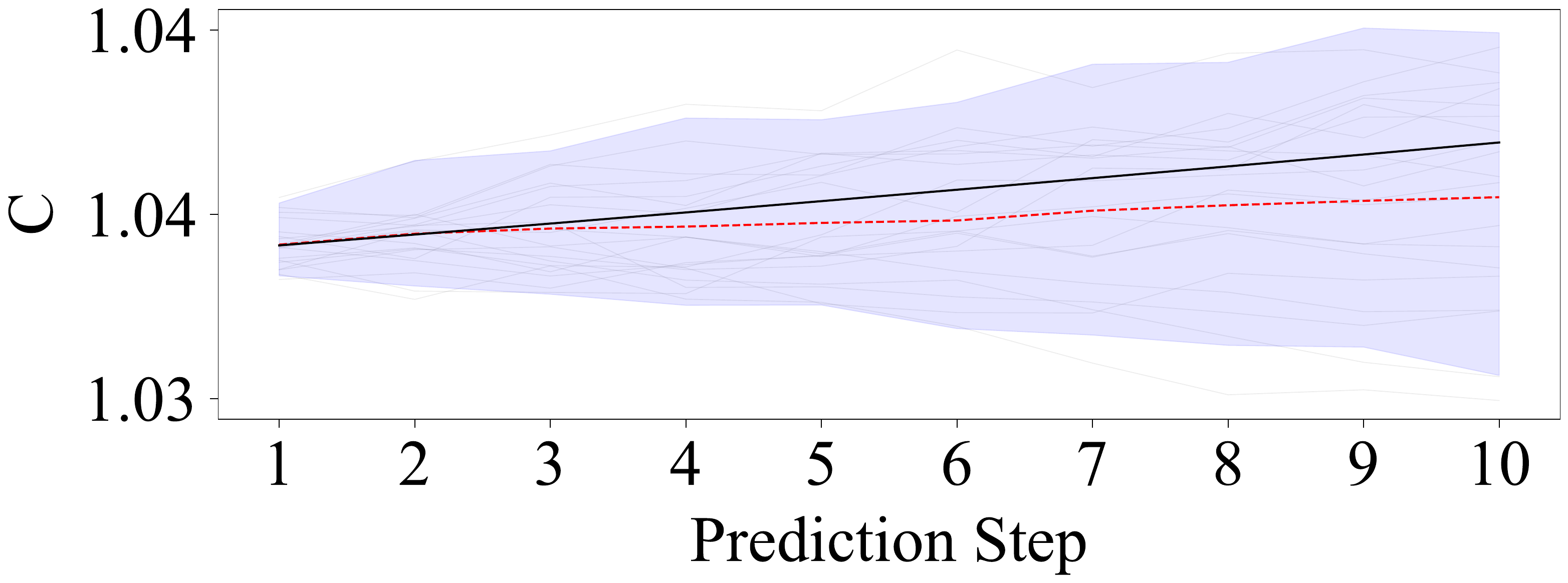}}

    \subfloat[PINN-Hybrid Matérn\label{fig:cstr_pinn_hybrid_matern}]{
    \includegraphics[width=0.48\linewidth]{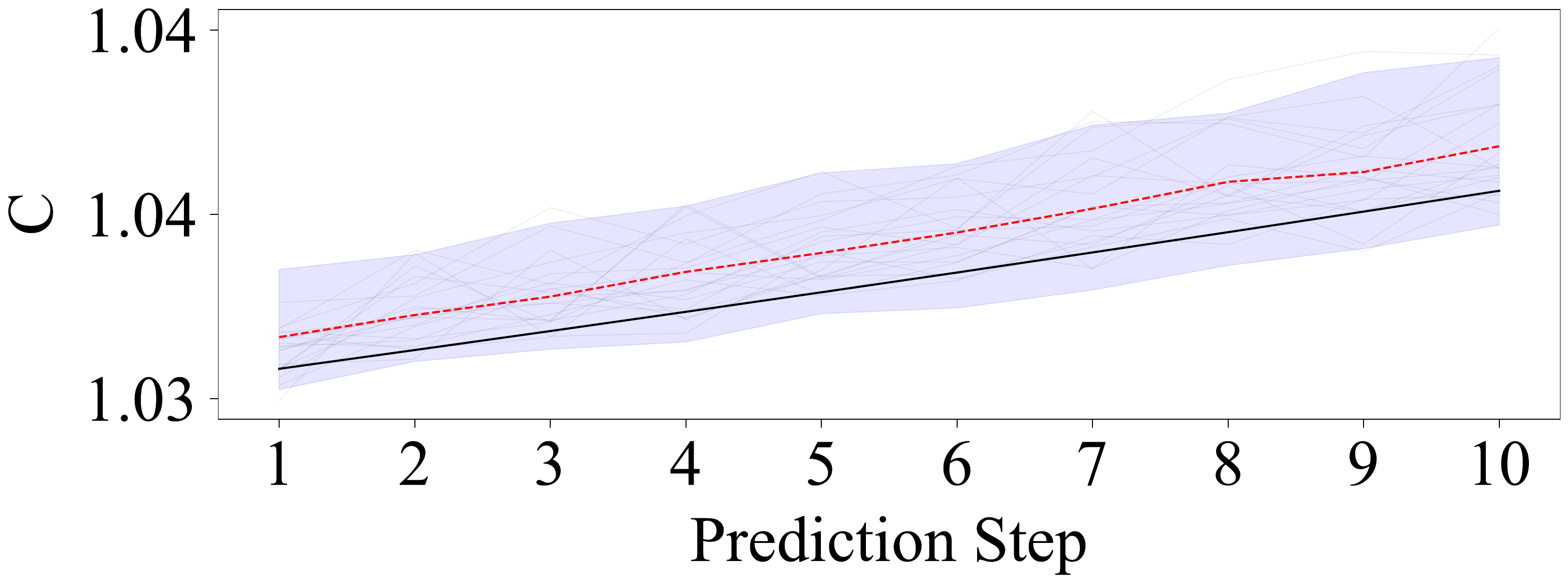}}
    \hfill
    \subfloat[FFNN-Hybrid Matérn\label{fig:cstr_ffnn_hybrid_matern}]{
    \includegraphics[width=0.48\linewidth]{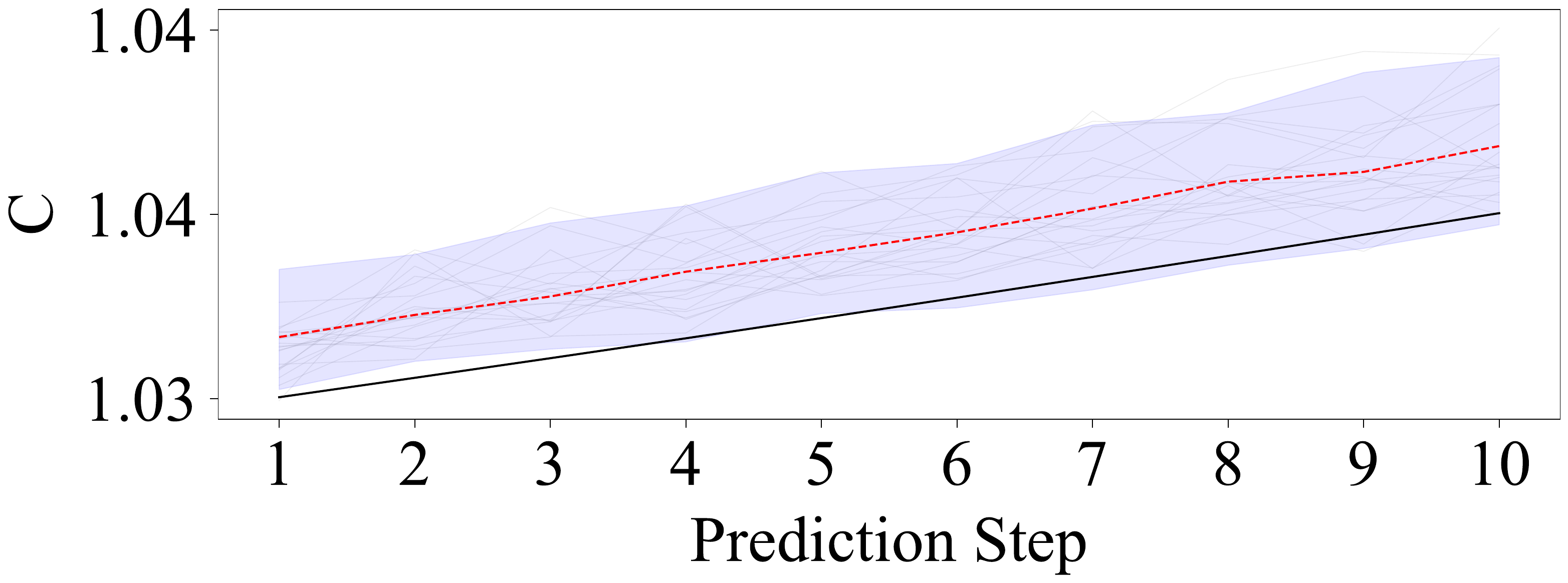}}

    \vspace{0.3cm}
    
    \centering
    \includegraphics[width=\linewidth]{Figures/legend_GP_95.pdf}

    \caption{10-step ahead predictions of CSTR output concentration ($C$) given forecasted input ($C_{in}$) from the four state transition models. PINN (left) and FFNN (right) predictions are shown with the 95\% confidence interval for each input forecasting model type.}
    \label{fig:pinn_ffnn_complete_cstr}
\end{figure*}

\begin{figure*}[htbp]
    \centering

    \subfloat[PINN-Exponential\label{fig:pfr_pinn_exp}]{
    \includegraphics[width=0.48\linewidth]{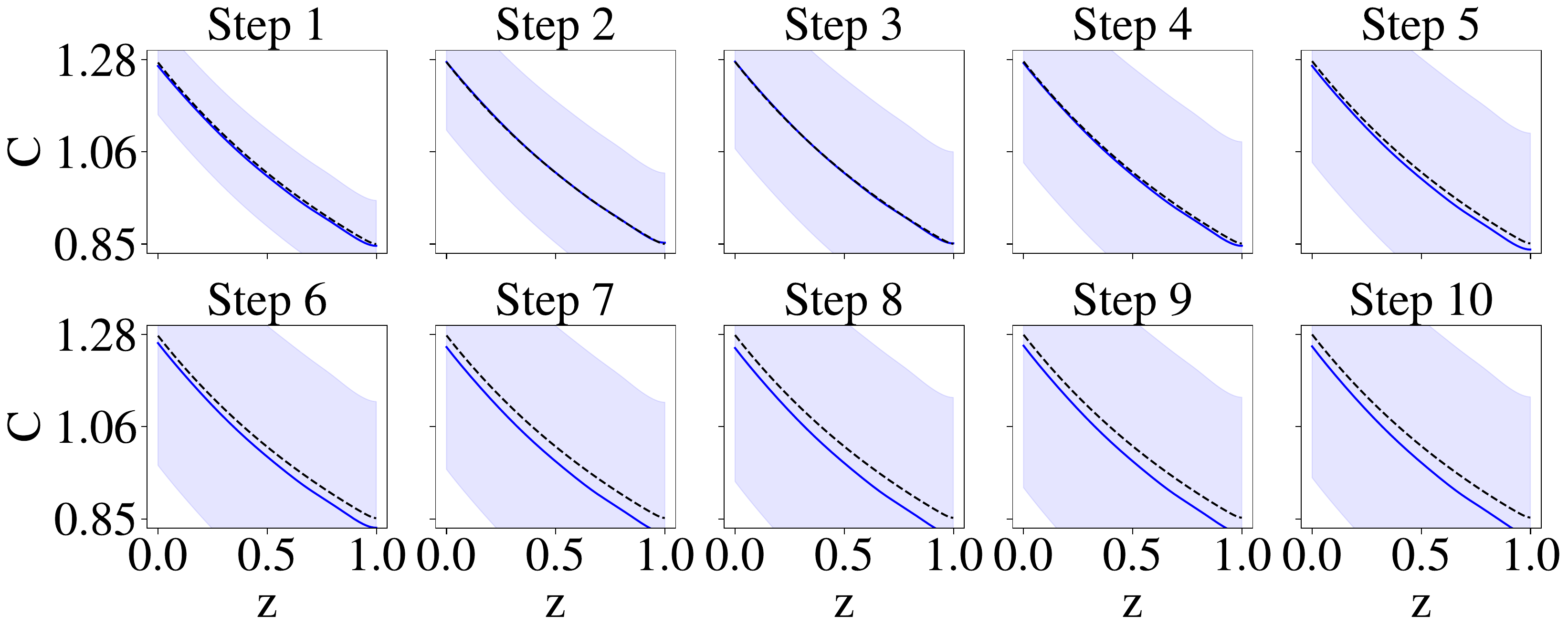}}
    \hfill
    \subfloat[FFNN-Exponential\label{fig:pfr_ffnn_exp}]{
    \includegraphics[width=0.48\linewidth]{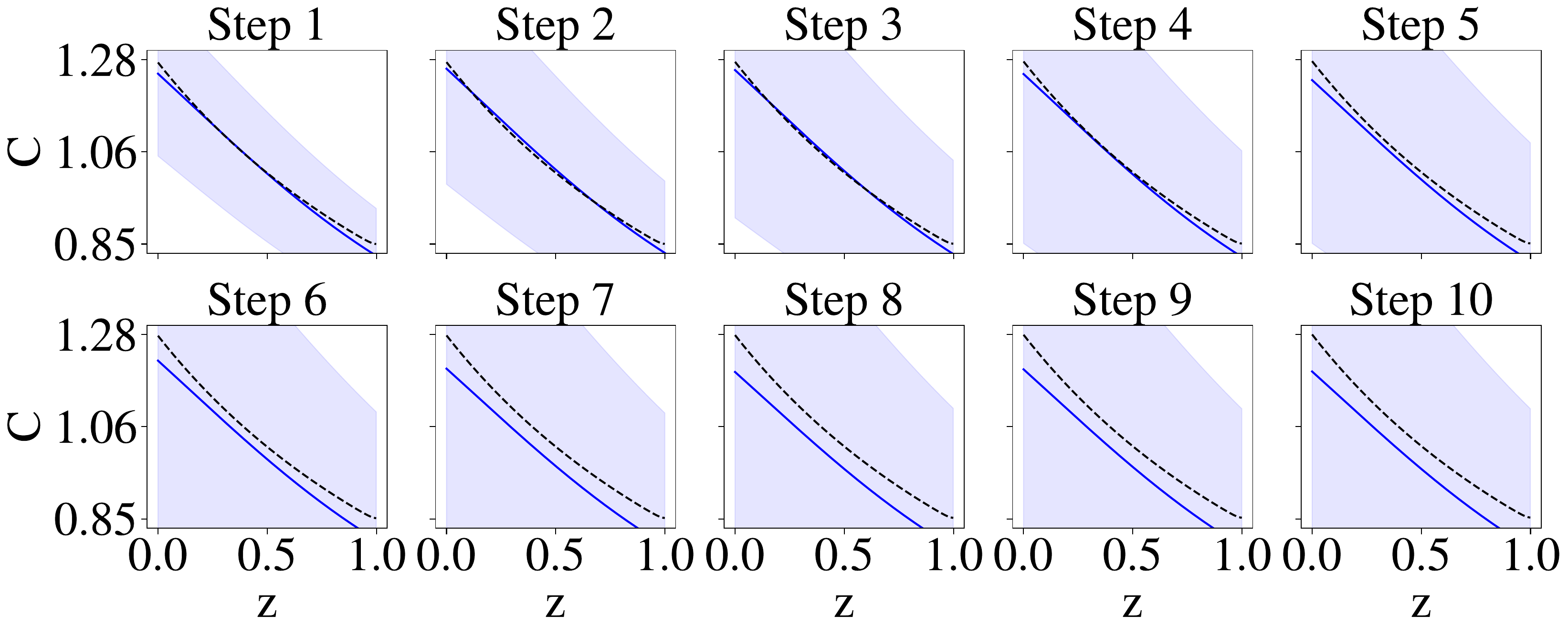}}

    \subfloat[PINN-Matérn\label{fig:pfr_pinn_matern}]{
    \includegraphics[width=0.48\linewidth]{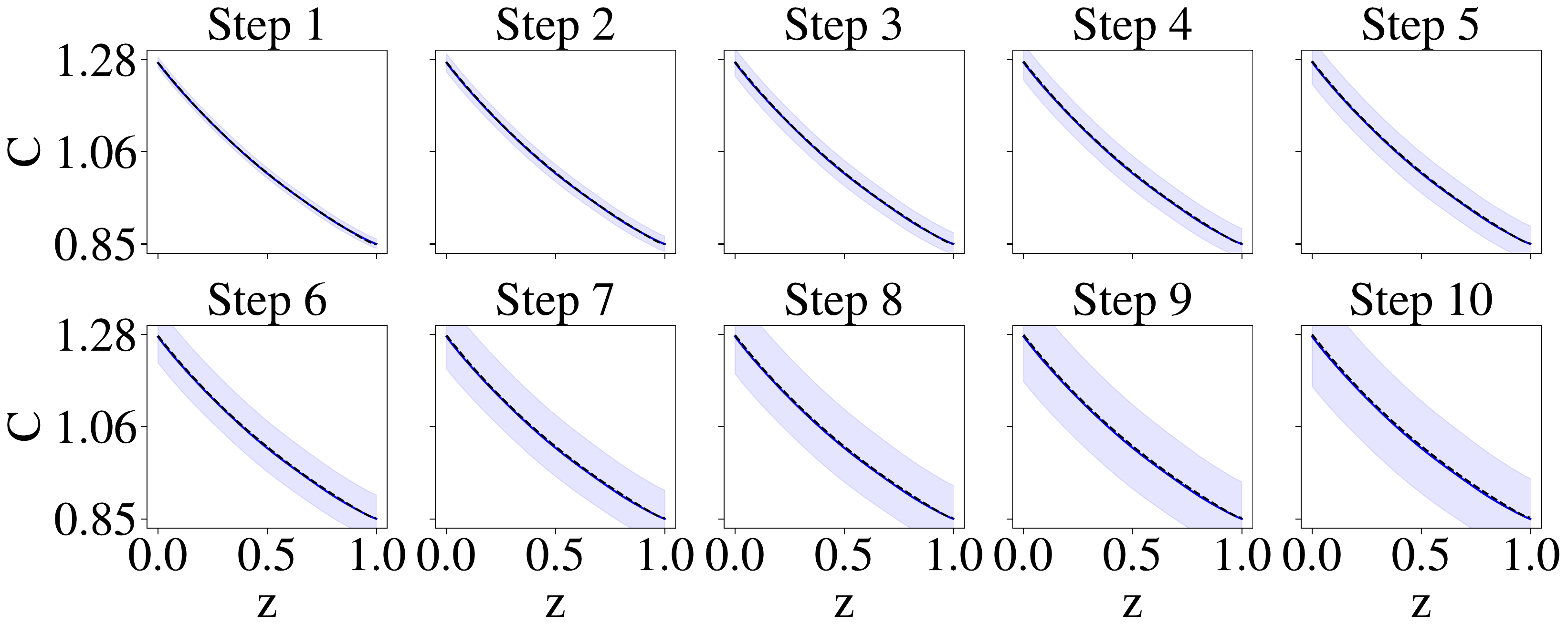}}
    \hfill
    \subfloat[FFNN-Matérn\label{fig:pfr_ffnn_matern}]{
    \includegraphics[width=0.48\linewidth]{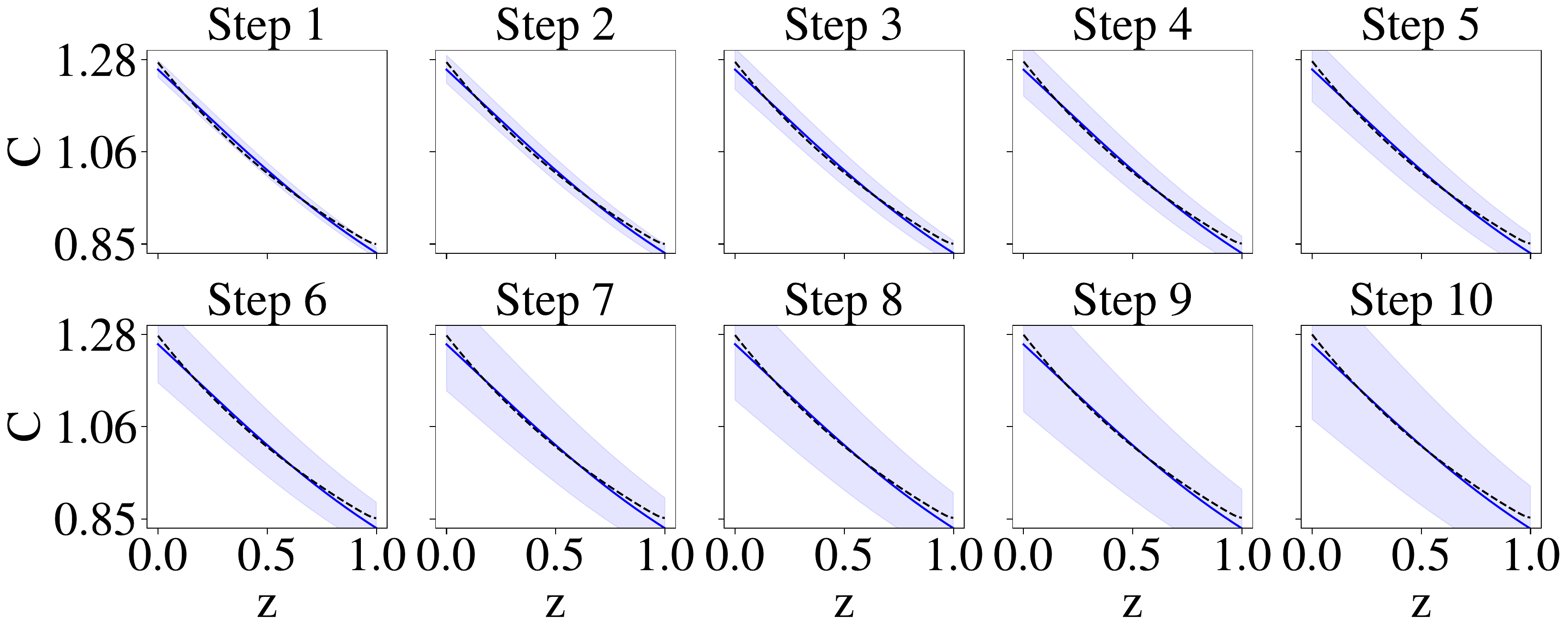}}
    
    \subfloat[PINN-Hybrid Exponential\label{fig:pfr_pinn_hybrid_exp}]{
    \includegraphics[width=0.48\linewidth]{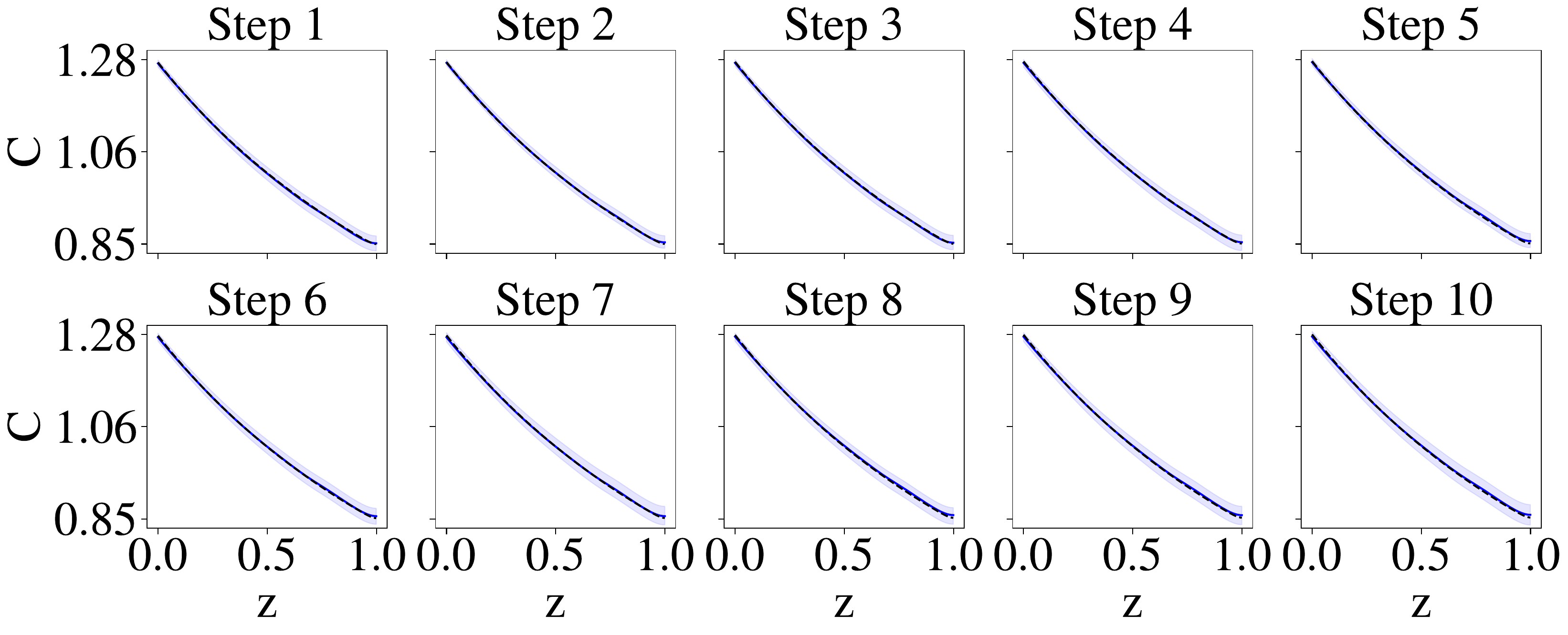}}
    \hfill
    \subfloat[FFNN-Hybrid Exponential\label{fig:pfr_ffnn_hybrid_exp}]{
    \includegraphics[width=0.48\linewidth]{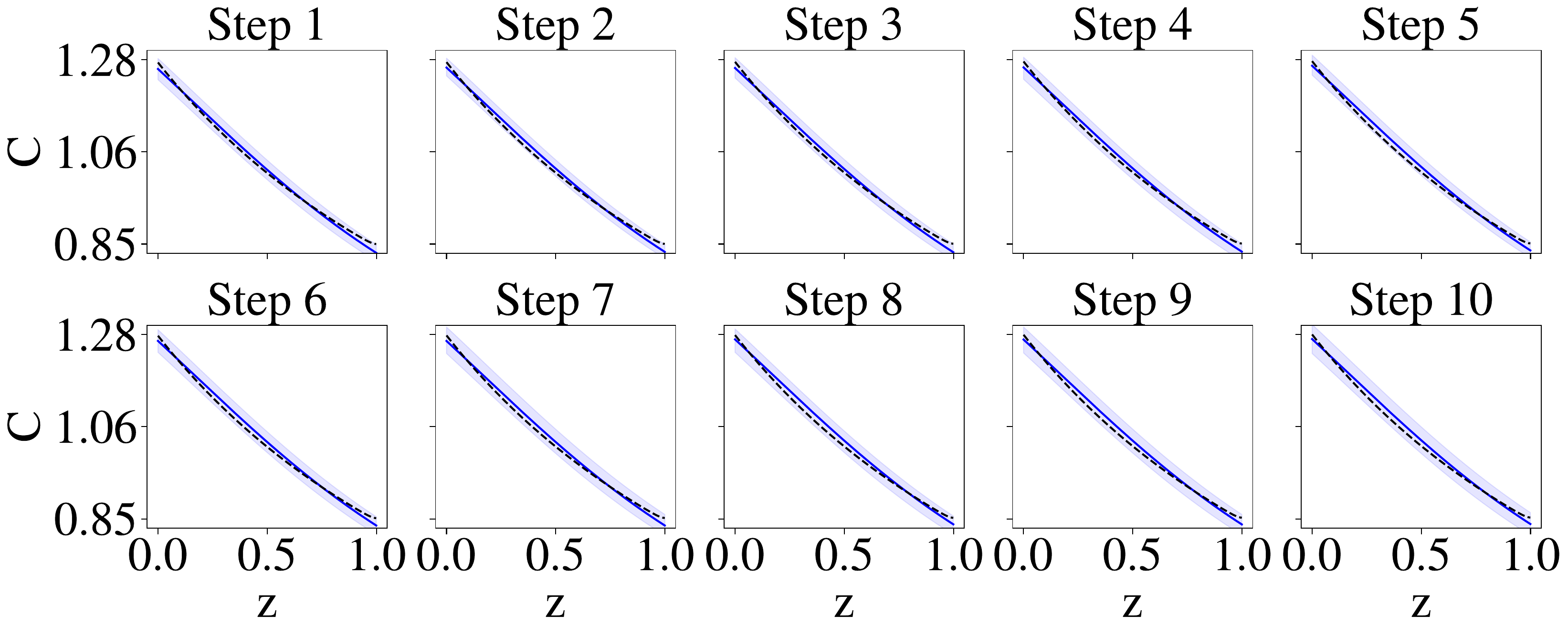}}

    \subfloat[PINN-Hybrid Matérn\label{fig:pfr_pinn_hybrid_matern}]{
    \includegraphics[width=0.48\linewidth]{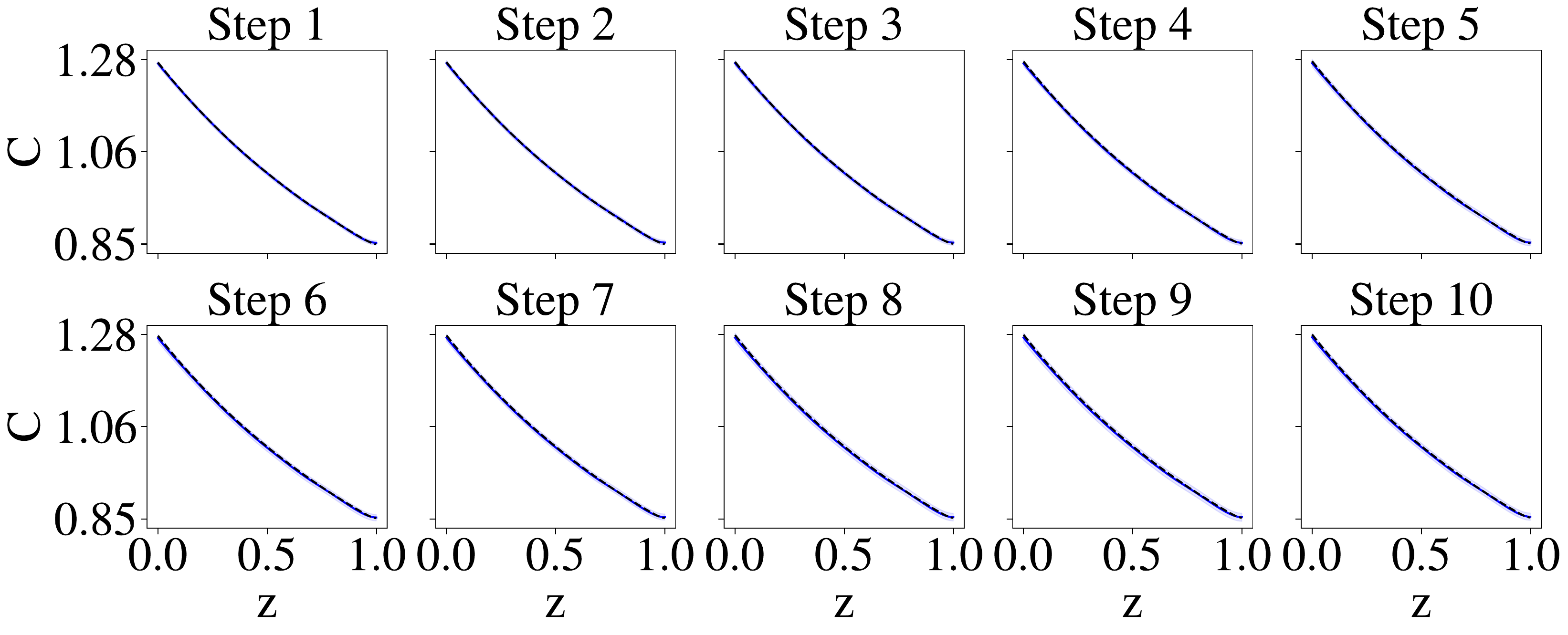}}
    \hfill
    \subfloat[FFNN-Hybrid Matérn\label{fig:pfr_ffnn_hybrid_matern}]{
    \includegraphics[width=0.48\linewidth]{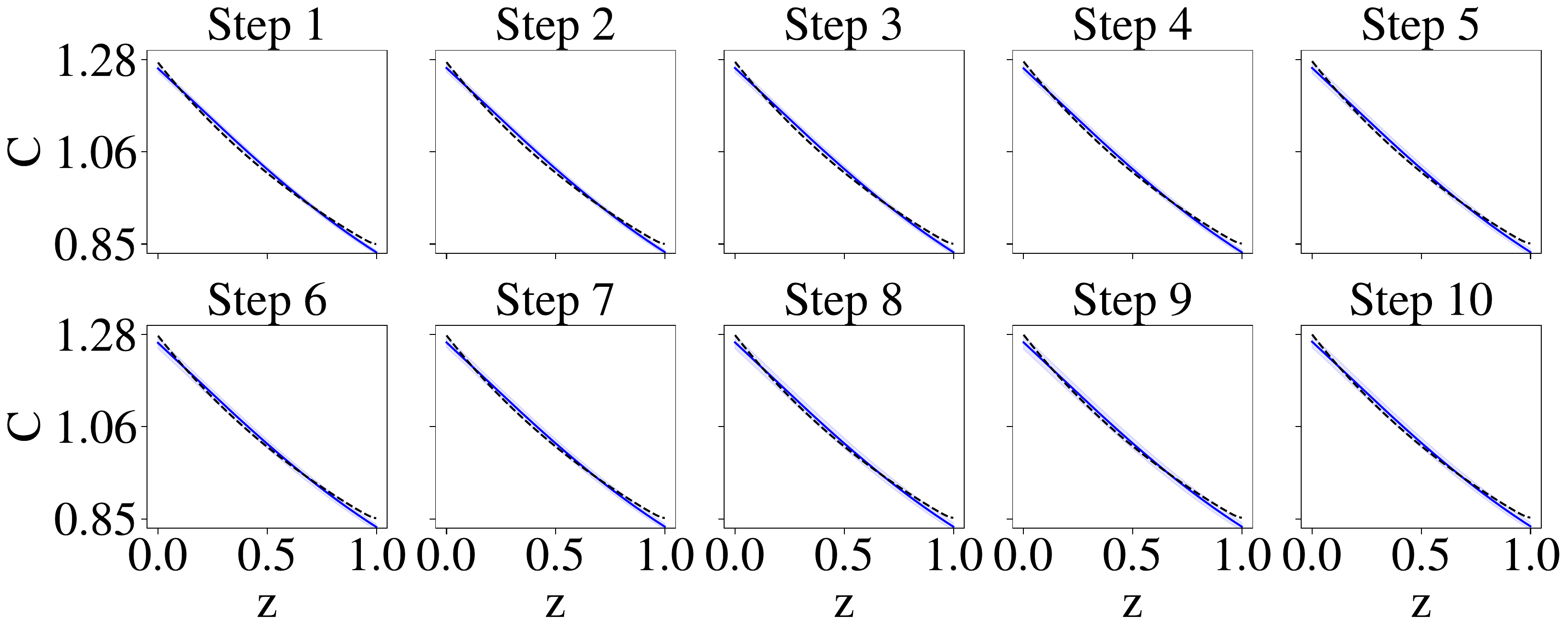}}

    \vspace{0.3cm}
    
    \centering
    \includegraphics[width=\linewidth]{Figures/legend_GP_95.pdf}

    \caption{Multi-step forecasting of ADPFR concentration profiles using predicted velocity ($v$) and inlet concentration ($C_{in}$) inputs. The four state transition models drive PINN (left) and FFNN (right) predictions over 10 steps, with shaded regions indicating the 95\% confidence interval.}
    \label{fig:pinn_ffnn_complete_pfr}
\end{figure*}

\begin{figure*}[htbp]
    \centering

    \subfloat[PINN-Exponential\label{fig:flotation_pinn_exp}]{
    \includegraphics[width=0.48\linewidth]{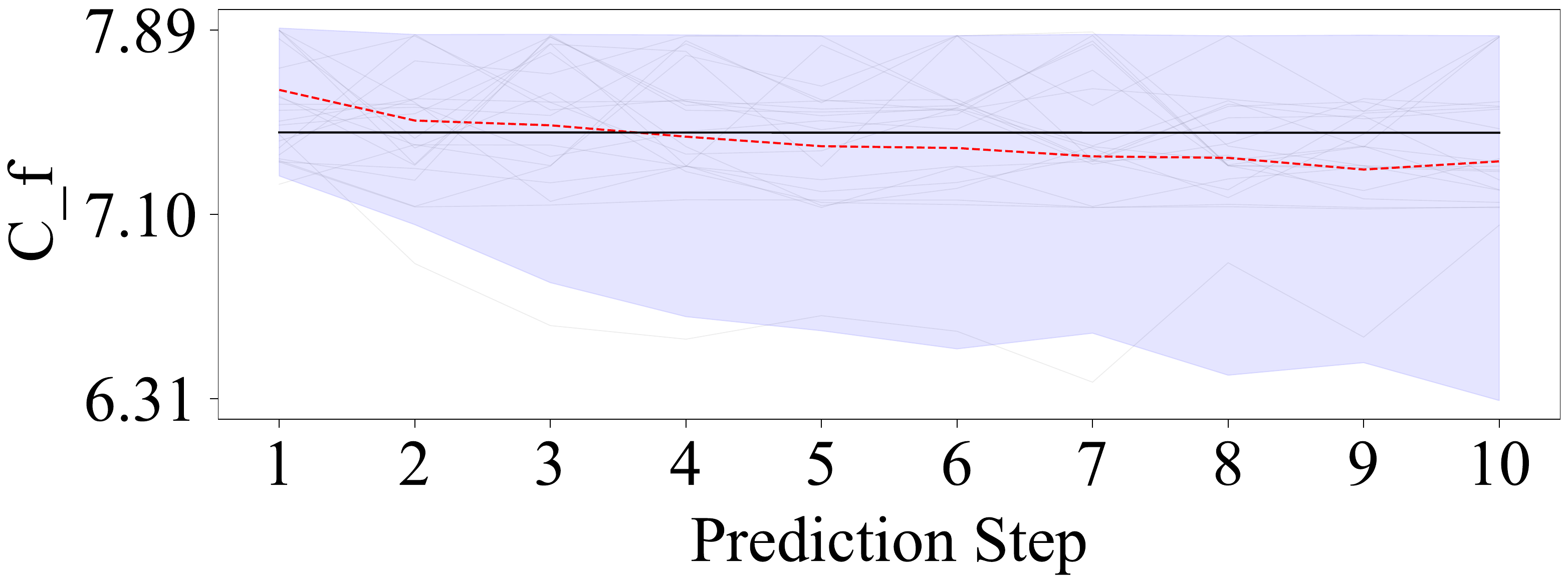}}
    \hfill
    \subfloat[FFNN-Exponential\label{fig:flotation_ffnn_exp}]{
    \includegraphics[width=0.48\linewidth]{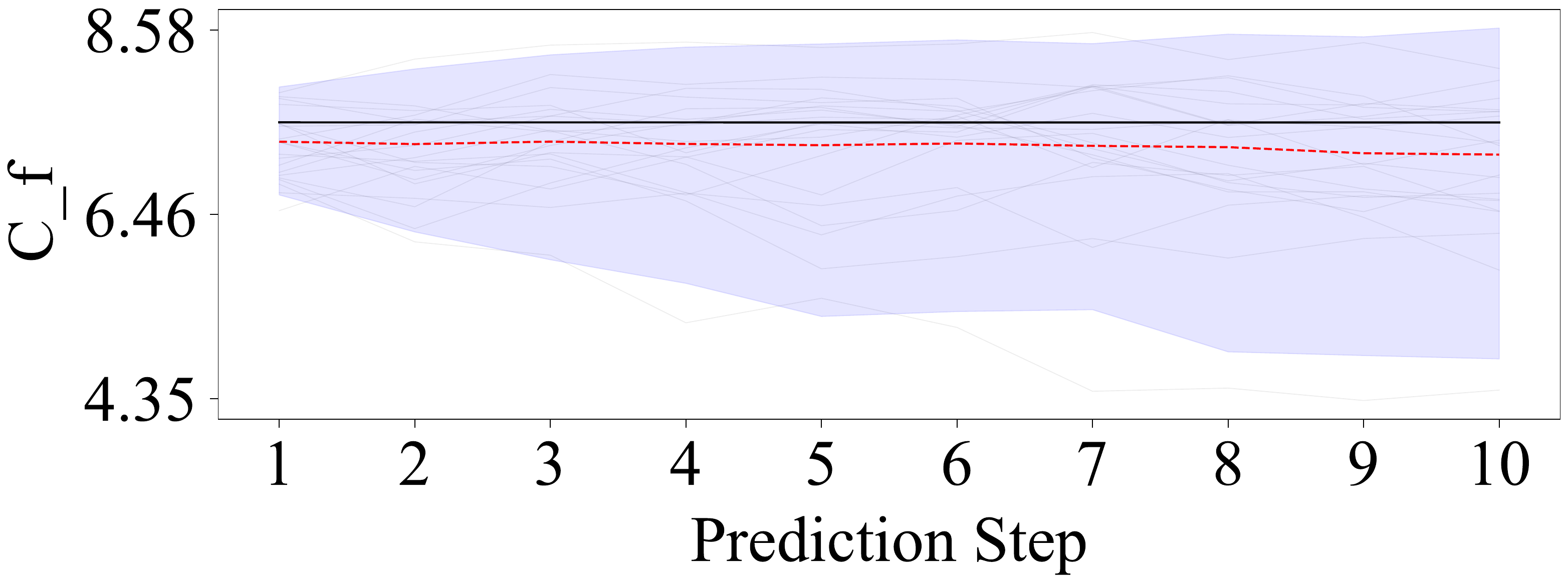}}

    \subfloat[PINN-Matérn\label{fig:flotation_pinn_matern}]{
    \includegraphics[width=0.48\linewidth]{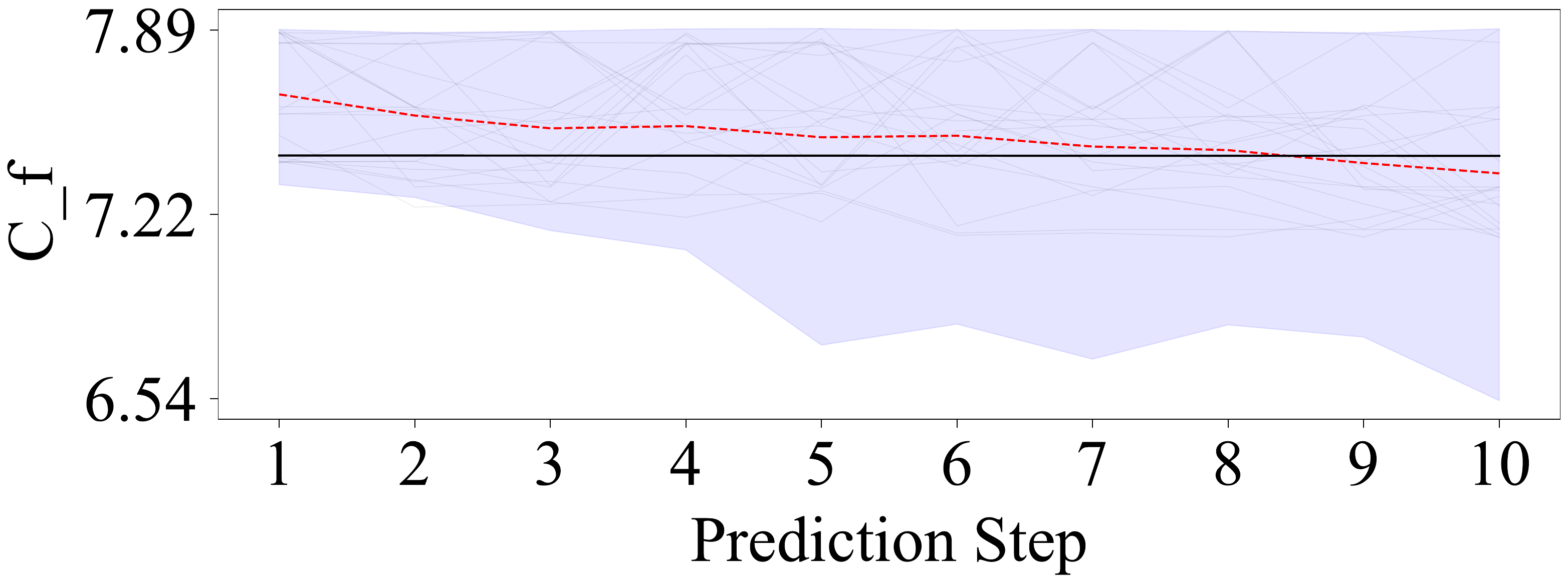}}
    \hfill
    \subfloat[FFNN-Matérn\label{fig:flotation_ffnn_matern}]{
    \includegraphics[width=0.48\linewidth]{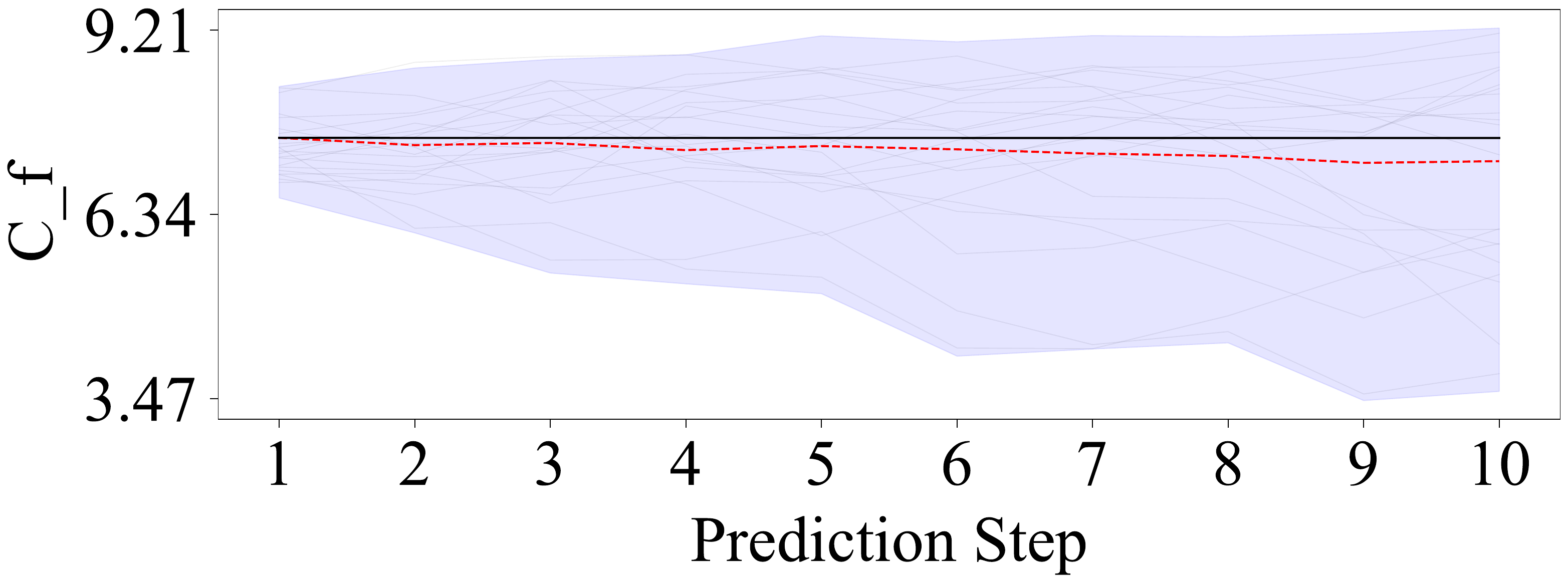}}
    
    \subfloat[PINN-Hybrid Exponential\label{fig:flotation_pinn_hybrid_exp}]{
    \includegraphics[width=0.48\linewidth]{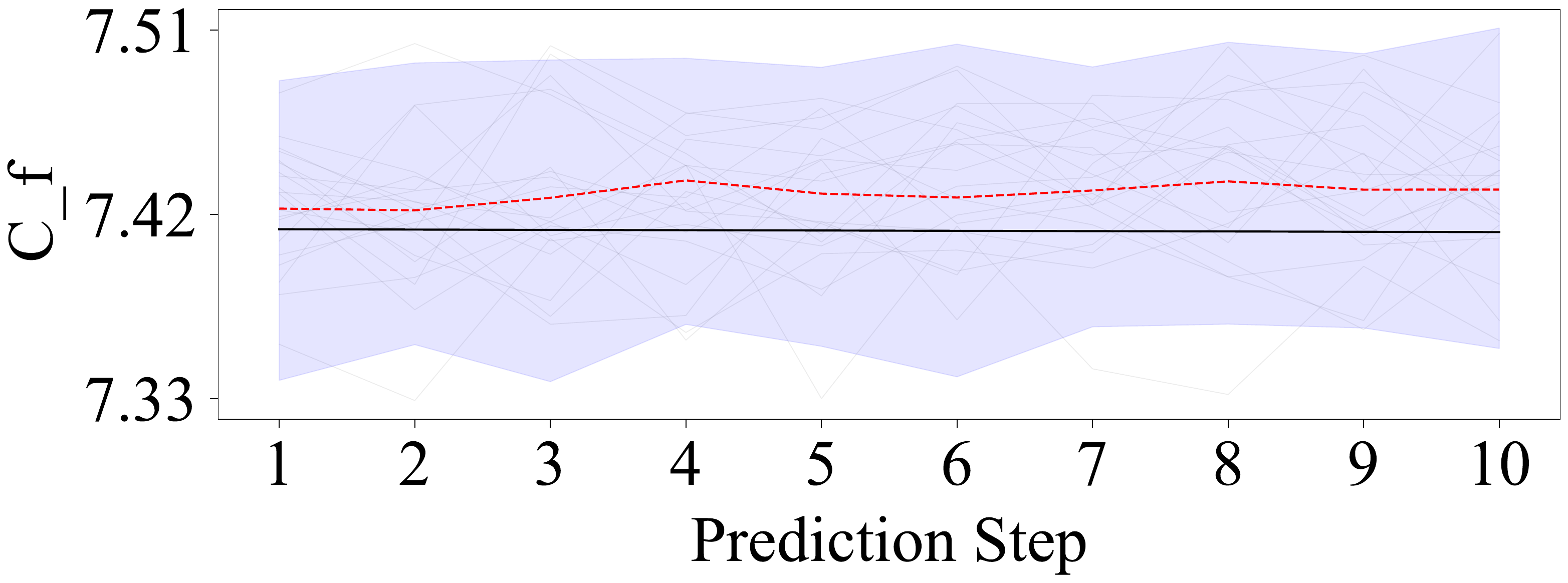}}
    \hfill
    \subfloat[FFNN-Hybrid Exponential\label{fig:flotation_ffnn_hybrid_exp}]{
    \includegraphics[width=0.48\linewidth]{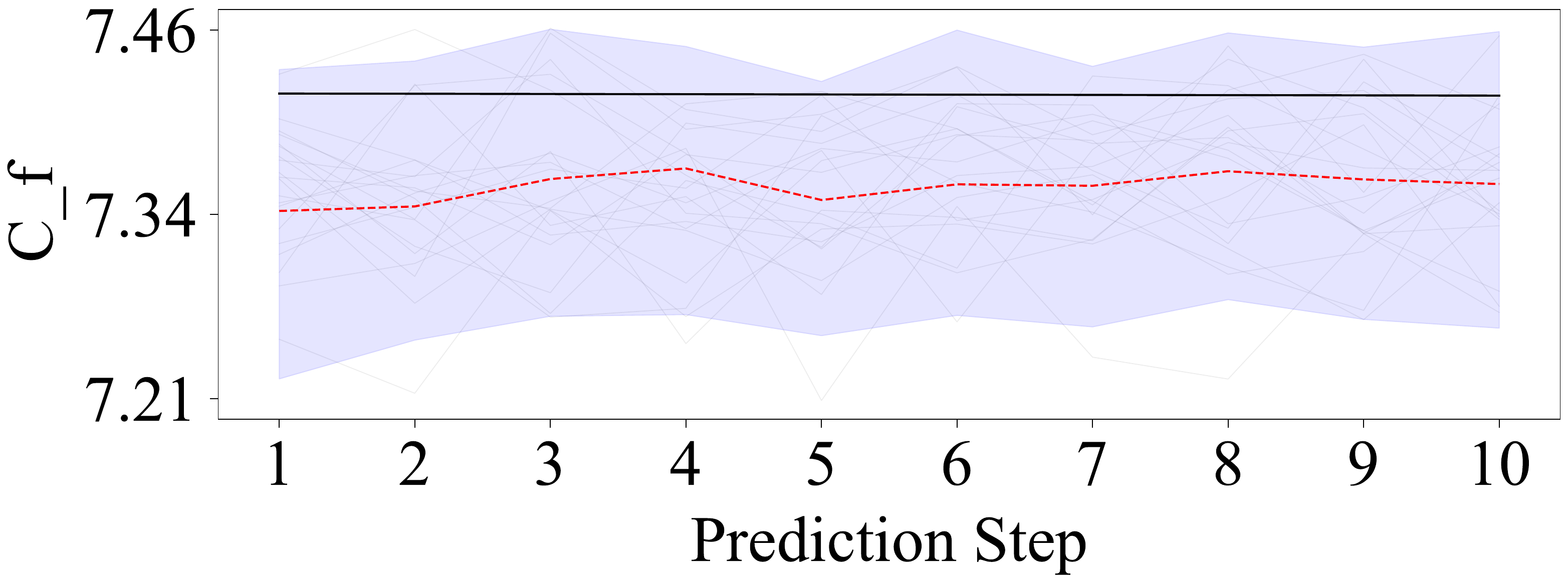}}

    \subfloat[PINN-Hybrid Matérn\label{fig:flotation_pinn_hybrid_matern}]{
    \includegraphics[width=0.48\linewidth]{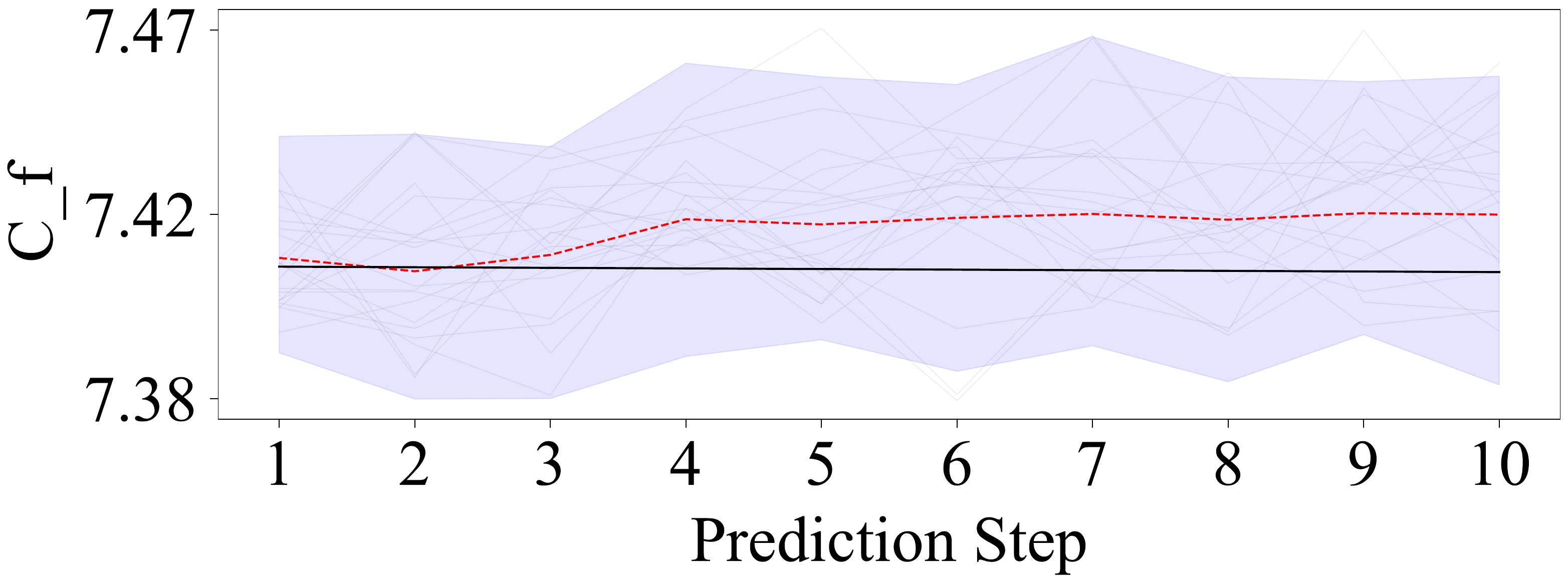}}
    \hfill
    \subfloat[FFNN-Hybrid Matérn\label{fig:flotation_ffnn_hybrid_matern}]{
    \includegraphics[width=0.48\linewidth]{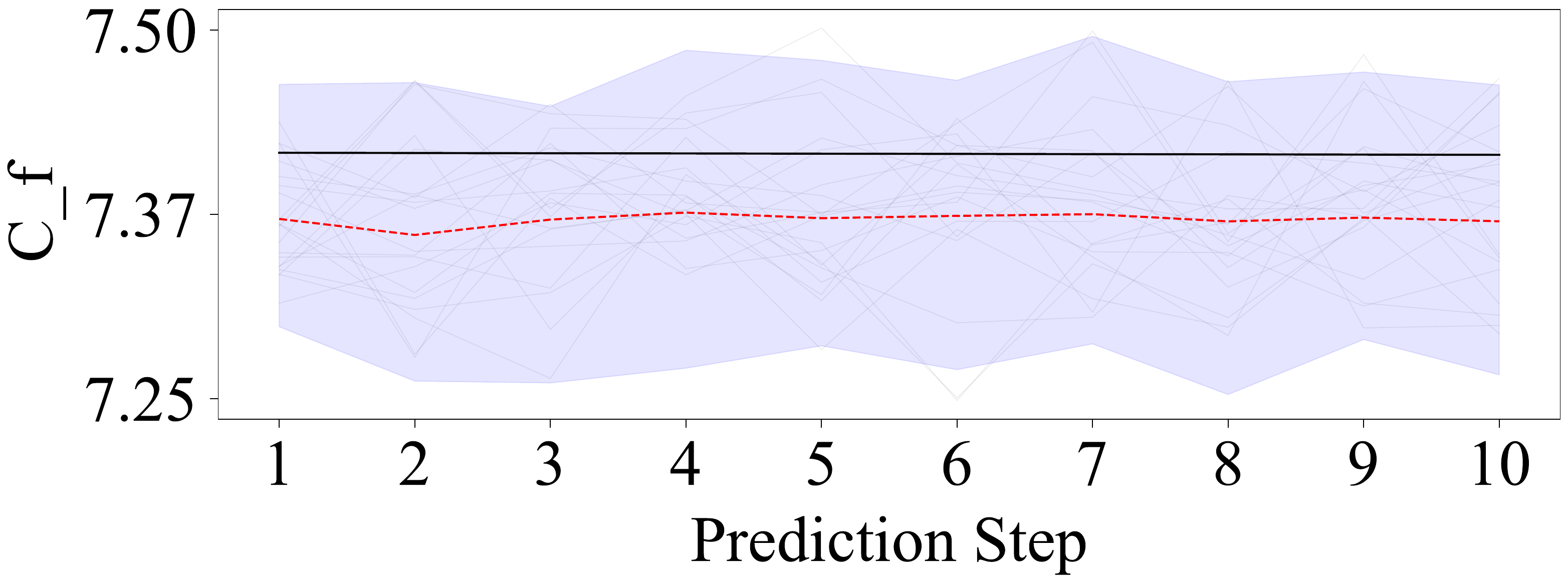}}

    \vspace{0.3cm}
    
    \centering
    \includegraphics[width=\linewidth]{Figures/legend_GP_95.pdf}

    \caption{Integrated forecasting of flotation concentrate grade ($C_f$) over 10 steps using predicted inputs. Each panel compares PINN (left) and FFNN (right) performance when driven by different state transition models, with uncertainty bounds shown at the 95\% confidence level.}
    \label{fig:pinn_ffnn_complete_flotation}
\end{figure*}

In the complete forecasting approach, predicted input trajectories from the state transition models are sequentially fed to the trained neural networks for multi-step output prediction. For each input trajectory realization, the PINNs and their data-driven counterparts recursively compute system outputs using the forecasted inputs, enabling 10‑step ahead forecasting with uncertainty quantification. Figures~\ref{fig:pinn_ffnn_complete_cstr},~\ref{fig:pinn_ffnn_complete_pfr}, and~\ref{fig:pinn_ffnn_complete_flotation} illustrate the multi‑step forecasting capability of PINNs and purely data-driven FFNNs for CSTR, ADPFR, and flotation systems, respectively. 

The results show that in all three systems, all integrated models are capable of multi‑step forecasting, with their mean predictions following the system dynamics. However, the uncertainty bounds show two main observations. First, forecasts driven by hybrid state transition models exhibit tighter confidence intervals than those using conventional state transition models. Second, within each input forecasting model category, PINNs generally produce narrower confidence intervals than their FFNN counterparts, with the hybrid Matérn model yielding the narrowest confidence intervals across all systems.

\begin{figure*}[htbp]
    \centering

    \subfloat[CSTR: $C$\label{fig:pinn_ffnn_likelihood_cstr}]{
    \includegraphics[width=0.32\linewidth]{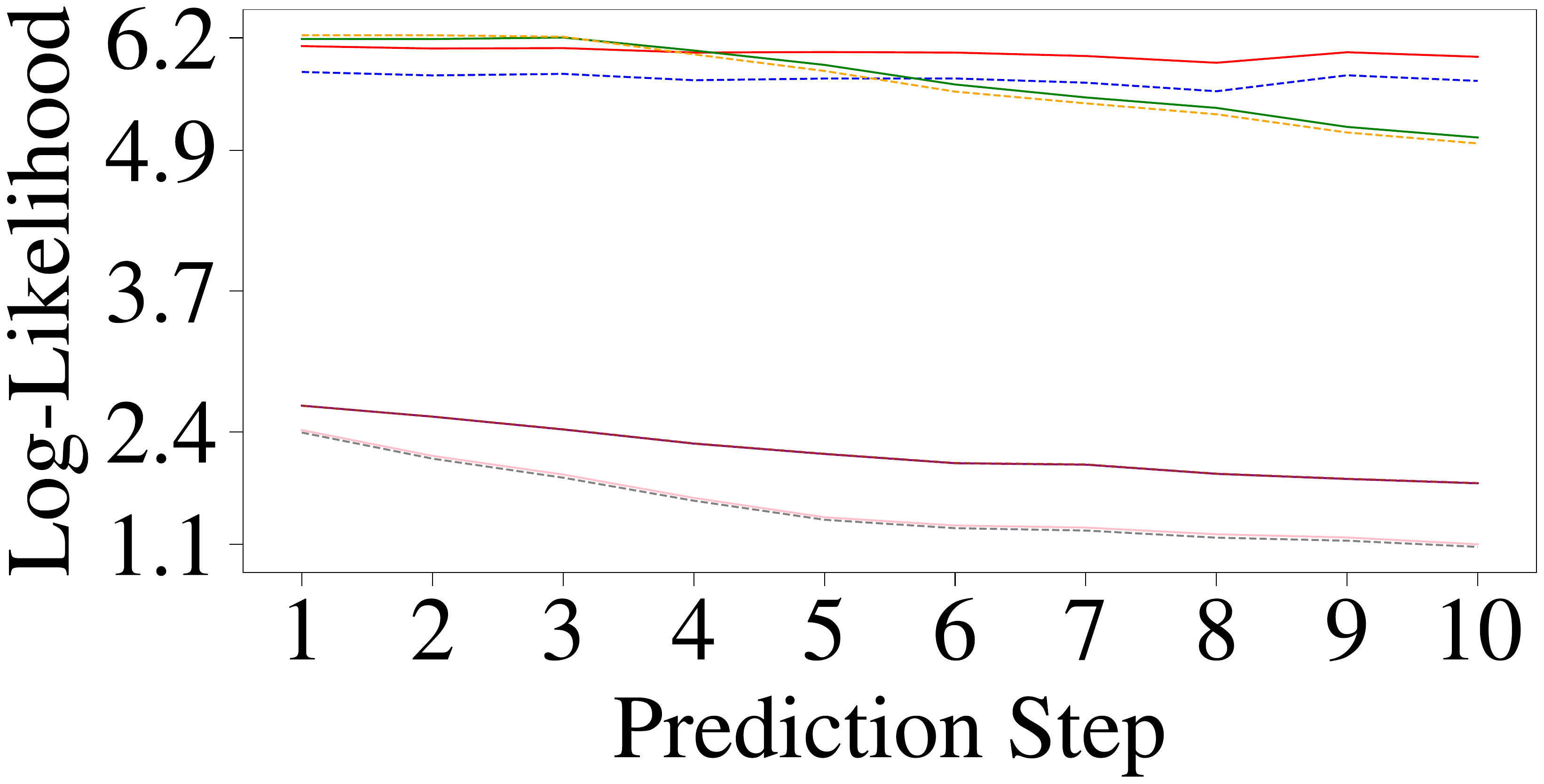}}
    \hfill
    \subfloat[ADPFR: $C$\label{fig:pinn_ffnn_likelihood_pfr}]{
    \includegraphics[width=0.32\linewidth]{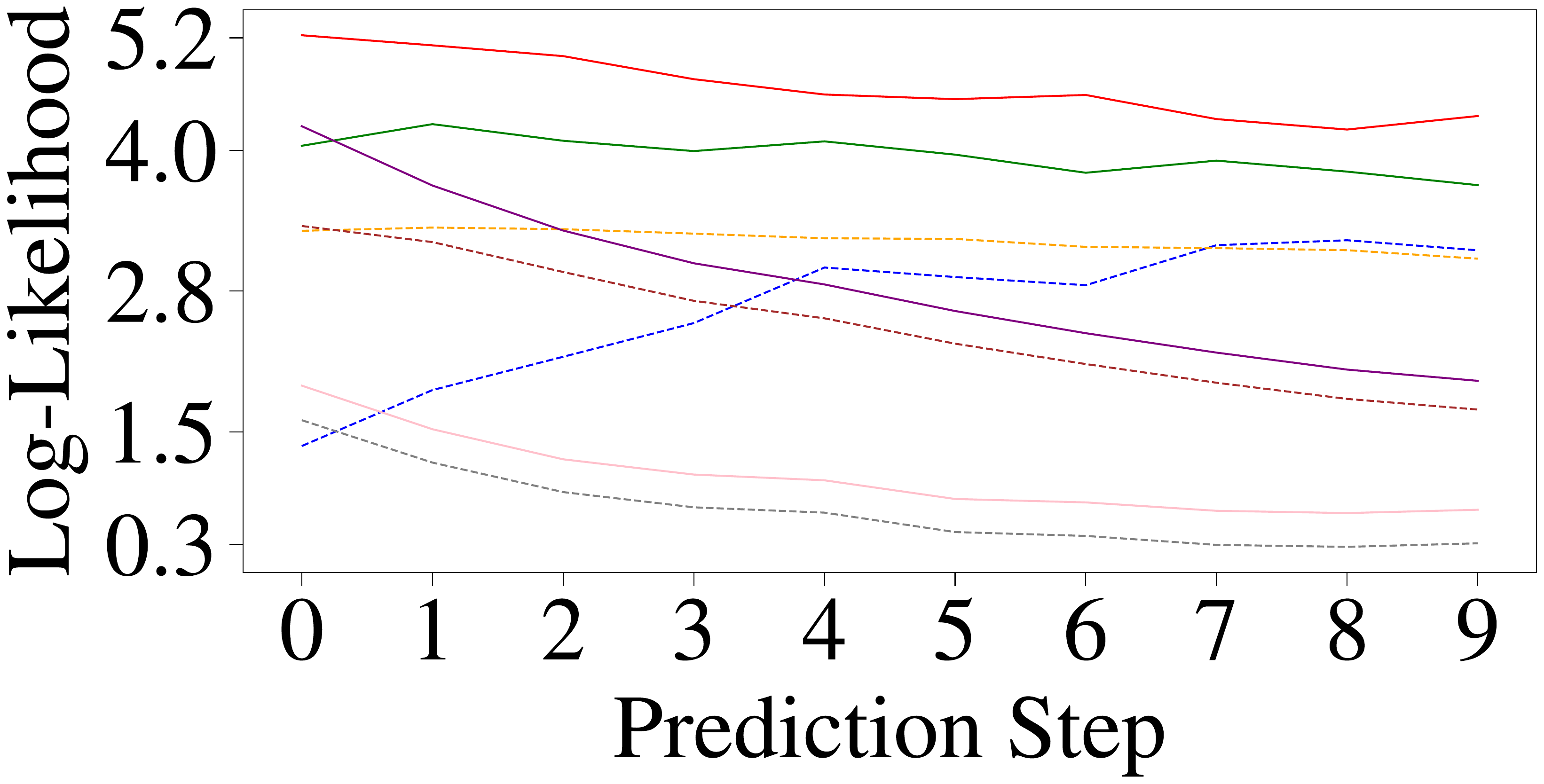}}
    \hfill
    \subfloat[Flotation: $C_f$\label{fig:pinn_ffnn_likelihood_flotation}]{
    \includegraphics[width=0.32\linewidth]{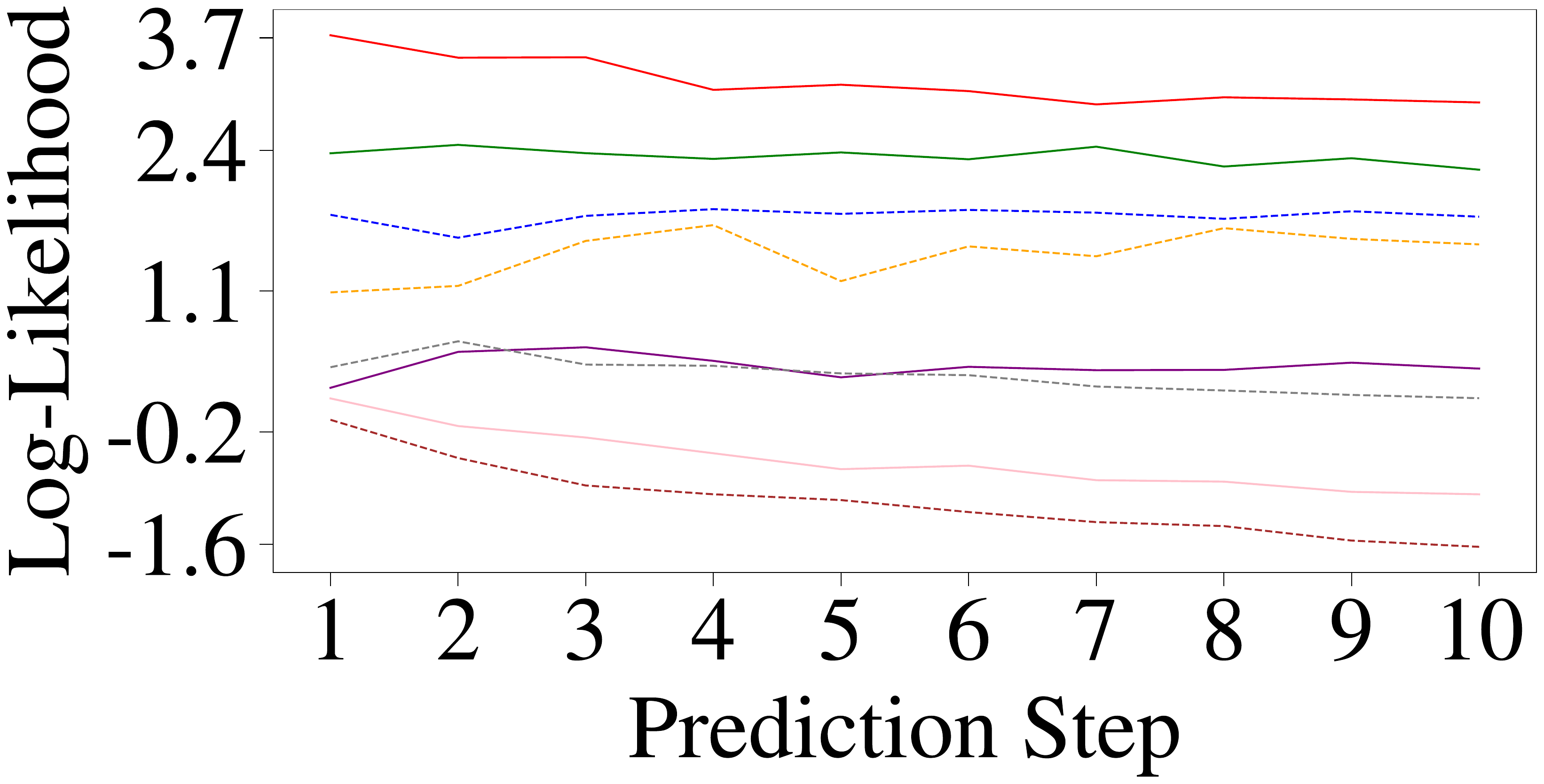}}

    \vspace{0.3cm}
    
    \centering
    \includegraphics[width=\linewidth]{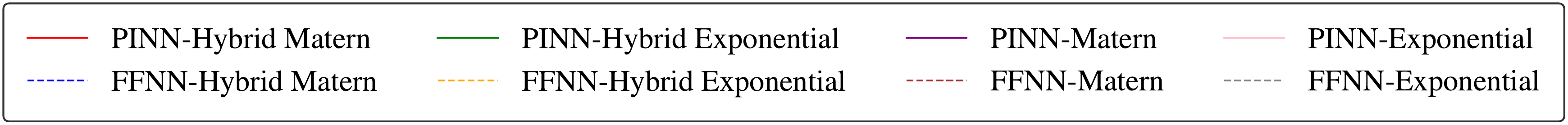}

    \caption{Log-likelihood comparison of integrated input and output models for 10‑step ahead forecasting across (a) CSTR, (b) ADPFR, and (c) flotation systems. For each system, PINNs and FFNNs are compared when driven by four state transition models. Higher values indicate better probabilistic calibration. PINNs with hybrid inputs consistently achieve the highest log-likelihood.}
    \label{fig:all_pinn_ffnn_likelihood}
\end{figure*}

A probabilistic evaluation through log-likelihood comparison for all three dynamical systems is shown in Figures~\ref{fig:pinn_ffnn_likelihood_cstr}, \ref{fig:pinn_ffnn_likelihood_pfr}, and \ref{fig:pinn_ffnn_likelihood_flotation}. The results demonstrate two trends. First, for any given input forecasting model, PINNs achieve higher log-likelihood values than their FFNN counterparts. Second, models driven by hybrid input forecasting models obtain significantly higher log-likelihood values across the prediction horizon than those using conventional input forecasting models, regardless of whether a PINN or data-driven FFNN is used for output prediction. 

Table~\ref{table:pinn_ffnn_MSE_performance_complete} quantifies the performance of integrated forecasting models using MSE and MAE metrics for 10‑step ahead mean predictions across all the systems. The results demonstrate that for every input forecasting model type, PINNs achieve lower errors than their FFNN counterparts. Furthermore, neural networks driven by inputs produced by LSTM-integrated state transition models lead to lower errors than those using conventional input forecasting models. The PINN with hybrid input forecasting models achieves the best overall performance, with the hybrid Matérn model yielding the lowest errors. From Figures~\ref{fig:all_pinn_ffnn_likelihood} and this table, it can be concluded that PINNs driven by hybrid state transition inputs provide more accurate forecasting results both in terms of predictive distributions and point predictions.
\begin{table}[htbp]
\caption{Performance evaluation of integrated forecasting models using MSE and MAE metrics for 10‑step ahead mean predictions across three dynamical systems.}
\label{table:pinn_ffnn_MSE_performance_complete}
\centering
\renewcommand{\arraystretch}{1.3}
\begin{tabular}{l l c c}
\hline
System & Model & MSE & MAE \\
\hline
\multirow{8}{*}{CSTR} 
  & PINN-Exponential & $1.88 \cdot 10^{-4}$ & $1.31 \cdot 10^{-2}$ \\
  & FFNN-Exponential & $2.24 \cdot 10^{-4}$ & $1.42 \cdot 10^{-2}$ \\
  & PINN-Matérn & $3.33 \cdot 10^{-4}$ & $1.82 \cdot 10^{-2}$ \\
  & FFNN-Matérn & $3.36 \cdot 10^{-4}$ & $1.83 \cdot 10^{-2}$ \\
  & PINN-Hybrid Exponential & $3.42 \cdot 10^{-7}$ & $4.73 \cdot 10^{-4}$ \\
  & FFNN-Hybrid Exponential & $5.50 \cdot 10^{-7}$ & $5.97 \cdot 10^{-4}$ \\
  & PINN-Hybrid Matérn & $3.20 \cdot 10^{-7}$ & $5.60 \cdot 10^{-4}$ \\
  & FFNN-Hybrid Matérn & $7.90 \cdot 10^{-7}$ & $8.90 \cdot 10^{-4}$ \\
\hline
\multirow{8}{*}{ADPFR}
  & PINN-Exponential & $5.96 \cdot 10^{-4}$ & $1.94 \cdot 10^{-2}$ \\
  & FFNN-Exponential & $1.58 \cdot 10^{-3}$ & $3.24 \cdot 10^{-2}$ \\
  & PINN-Matérn & $7.54 \cdot 10^{-6}$ & $2.44 \cdot 10^{-3}$ \\
  & FFNN-Matérn & $7.70 \cdot 10^{-5}$ & $7.10 \cdot 10^{-3}$ \\
  & PINN-Hybrid Exponential & $2.76 \cdot 10^{-6}$ & $1.27 \cdot 10^{-3}$ \\
  & FFNN-Hybrid Exponential & $8.68 \cdot 10^{-5}$ & $8.15 \cdot 10^{-3}$ \\
  & PINN-Hybrid Matérn & $5.38 \cdot 10^{-6}$ & $1.96 \cdot 10^{-3}$ \\
  & FFNN-Hybrid Matérn & $6.87 \cdot 10^{-5}$ & $7.23 \cdot 10^{-3}$ \\
\hline
\multirow{8}{*}{Flotation}
  & PINN-Exponential & $8.61 \cdot 10^{-3}$ & $8.05 \cdot 10^{-2}$ \\
  & FFNN-Exponential & $6.13 \cdot 10^{-2}$ & $2.44 \cdot 10^{-1}$ \\
  & PINN-Matérn & $8.92 \cdot 10^{-3}$ & $7.79 \cdot 10^{-2}$ \\
  & FFNN-Matérn & $4.23 \cdot 10^{-2}$ & $1.76 \cdot 10^{-1}$ \\
  & PINN-Hybrid Exponential & $2.88 \cdot 10^{-4}$ & $1.64 \cdot 10^{-2}$ \\
  & FFNN-Hybrid Exponential & $3.25 \cdot 10^{-3}$ & $5.63 \cdot 10^{-2}$ \\
  & PINN-Hybrid Matérn & $1.06 \cdot 10^{-4}$ & $9.10 \cdot 10^{-3}$ \\
  & FFNN-Hybrid Matérn & $1.67 \cdot 10^{-3}$ & $4.06 \cdot 10^{-2}$ \\
\hline
\end{tabular}
\end{table}

\section{Conclusions}
\label{sec:concl}
In this paper, a dual-level, interpretable approach for multi-step forecasting with uncertainty quantification in dynamical systems is proposed. The method first generates probabilistic forecasts of time-varying inputs using LSTM-integrated state-space models. These forecast inputs are then propagated sequentially through discrete-time physics-informed neural networks to produce multi-step output predictions. The methodology is evaluated across three distinct dynamical systems: CSTR, ADPFR, and an industrial froth-flotation process.

The performance of the models is evaluated in three phases. First, the capability of physics-informed neural networks and their purely data-driven counterparts in modeling the dynamical systems is compared using known inputs. The results show that PINNs achieve lower mean squared error on unseen data, demonstrating their stronger generalization capability. Second, the multi-step ahead forecasting ability of the LSTM-integrated state transition models and their conventional counterparts is assessed. The LSTM-integrated STMs achieve higher log-likelihood and lower MSE, demonstrating superior forecasting accuracy in both point estimates and predictive distributions.

Finally, the integrated dual-level approach is evaluated for multi-step ahead forecasting. The results demonstrate that for any given input forecasting model, PINNs achieve higher log-likelihood and lower mean squared error than their purely data-driven counterparts. Furthermore, forecasts driven by LSTM-integrated state transition models outperform those using conventional models, regardless of whether the model is PINN or data-driven FFNN. Consequently, the combined approach of using PINNs with inputs from hybrid state transition models yields the most accurate forecasting results both in terms of predictive distributions and point predictions. 

\section*{Acknowledgements}
We acknowledge the financial support of the Finnish Ministry of Education and Culture through the Intelligent Work Machines Doctoral Education Pilot Program (IWM VN/3137/2024-OKM-4). 
%
%
%
%
%
%
%
%
%
%

\bibliography{references}



\end{document}